\documentclass[10 pt, journal]{IEEEtran} 
\usepackage{times}
\usepackage[pdftex]{graphicx}
\usepackage{subfigure}
\usepackage{amsmath,amssymb,amsopn,amstext,amsfonts}
\usepackage{cancel}
\usepackage[space]{cite}
\usepackage{pdfsync}
\usepackage{balance}
\usepackage{color}
\usepackage{mathtools}
\usepackage{algpseudocode}
\usepackage{algorithm} %\usepackage[ruled,vlined,linesnumbered]{algorithm2e}

\algnewcommand\algorithmicforeach{\textbf{for each}}
\algdef{S}[FOR]{ForEach}[1]{\algorithmicforeach\ #1\ \algorithmicdo}

\usepackage{bm}

\usepackage{diagbox}
\usepackage{float}
\usepackage{epstopdf}
\usepackage{pifont}
\usepackage{fixltx2e}
\usepackage{mathrsfs}
\usepackage{multirow}
\usepackage{booktabs}
\usepackage{tabularx}
\usepackage{url}
\usepackage{verbatim}
\usepackage[linkcolor=black,citecolor=black,urlcolor=black,colorlinks=true]{hyperref}
\usepackage{wrapfig}
\graphicspath{{./figures/}}
\DeclareGraphicsExtensions{.png,.jpg,.eps,.pdf}
\IEEEoverridecommandlockouts
\begin{document}

\title{Autonomous Flights inside Narrow Tunnels}
\author{Luqi Wang, Yan Ning, Hongming Chen, Peize Liu, Yang Xu, Hao Xu, Ximin Lyu and Shaojie Shen
\thanks{Manuscript received: June, 1, 2024; Revised November, 20, 2024; Accepted January, 31, 2025.}
\thanks{This paper was recommended for publication by Editor Mac Schwager upon evaluation of the Associate Editor and Reviewers' comments.
This work was supported by the HKUST-DJI Joint Innovation Laboratory, HKUST Postgraduate studentship, the Hong Kong Center for Construction Robotics (InnoHK center supported by Hong Kong ITC), and Guangdong-Hongkong-Macao Joint Research of Science and Technology Planning Funding from Guangdong Province under Grant: 2023A0505010019.}
\thanks{Luqi Wang, Yan Ning, Peize Liu, Yang Xu, and Shaojie Shen are with the Department of Electronic and Computer Engineering, the Hong Kong University of Science and Technology, Hong Kong, China (e-mail: lwangax@connect.ust.hk, yningaa@connect.ust.hk, peize.liu@connect.ust.hk, yxuew@connect.ust.hk, eeshaojie@ust.hk). Hongming Chen and Ximin Lyu are with the School of Intelligent Systems Engineering, SunYat-sen University, Guangzhou, China (e-mail: chenhm223@mail2.sysu.edu.cn, lvxm6@mail.sysu.edu.cn). Hao Xu is with the Institute of Unmanned Systems, Beihang University, Beijing, China (e-mail: xuhao3e8@buaa.edu.cn). (\textit{Corresponding author: Luqi Wang, Hao Xu, and Ximin Lyu.})}
\thanks{This article has supplementary material provided by the authors and color
versions of one or more figures available at: see top of this page.}
\thanks{Digital Object Identifier (DOI): see top of this page.}
}
\markboth{IEEE Transactions on Robotics. Preprint Version. Accepted January, 2025}
{Wang \MakeLowercase{\textit{et al.}}: Autonomous Flights inside Narrow Tunnels}

\maketitle

\begin{abstract}

Multirotors are usually desired to enter confined narrow tunnels that are barely accessible to humans in various applications including inspection, search and rescue, and so on. This task is extremely challenging since the lack of geometric features and illuminations, together with the limited field of view, cause problems in perception; the restricted space and significant ego airflow disturbances induce control issues. This paper introduces an autonomous aerial system designed for navigation through tunnels as narrow as 0.5 m in diameter. The real-time and online system includes a virtual omni-directional perception module tailored for the mission and a novel motion planner that incorporates perception and ego airflow disturbance factors modeled using camera projections and computational fluid dynamics analyses, respectively. Extensive flight experiments on a custom-designed quadrotor are conducted in multiple realistic narrow tunnels to validate the superior performance of the system, even over human pilots, proving its potential for real applications. Additionally, a deployment pipeline on other multirotor platforms is outlined and open-source packages are provided for future developments\footnote{\url{https://github.com/HKUST-Aerial-Robotics/FINT}}.

%Multirotors are usually desired to enter confined narrow tunnels that are barely accessible to humans in various applications including inspection, search and rescue and so on. This task is extremely challenging since the lack of geometry features and illuminations, together with the limited field of view, cause problems in perception; the restricted space and significant ego airflow disturbances induce control issues. This paper introduces an autonomous aerial system designed for navigation through tunnels as narrow as 0.5 m in diameter. The real-time and online system includes an virtual omni-directional perception module tailored for the mission and a novel motion planner that incorporates perception and ego airflow disturbance factors modeled using camera projections and computational fluid dynamics analyses, respectively. Extensive flight experiments on a custom-designed quadrotor are conducted in multiple realistic narrow tunnels to validate the superior performance of the system, even over human pilots, proving its potential for real applications. Additionally, a deployment pipeline on other multirotor platforms is outlined and open-source packages are provided for future developments.

\end{abstract}


\section{Introduction}
\label{sec:introduction}

\begin{figure}[t]
\begin{center}
%\vspace{-0.1cm}
%\subfigure[\label{fig:tunnel_0} The straight tunnel.]
%{\includegraphics[height=0.4\columnwidth]{tunnel_straight}} 
%\vspace{-0.1cm}
\subfigure[\label{fig:tunnel_ra} 2-D tunnel case 1.]
{\includegraphics[height=0.34\columnwidth]{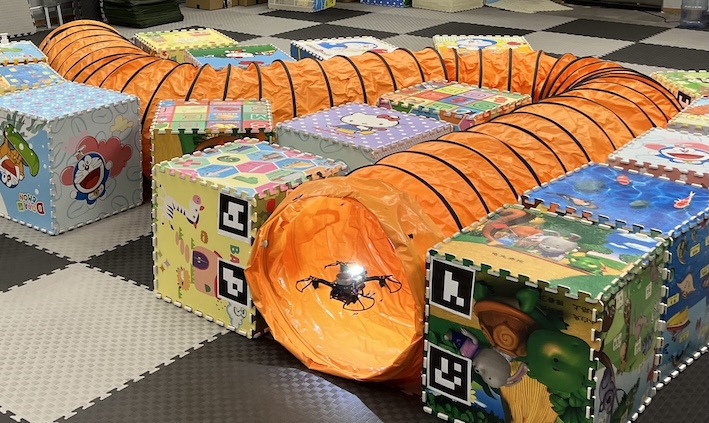}} 
\subfigure[\label{fig:tunnel_n} 2-D tunnel case 2.]
{\includegraphics[height=0.34\columnwidth]{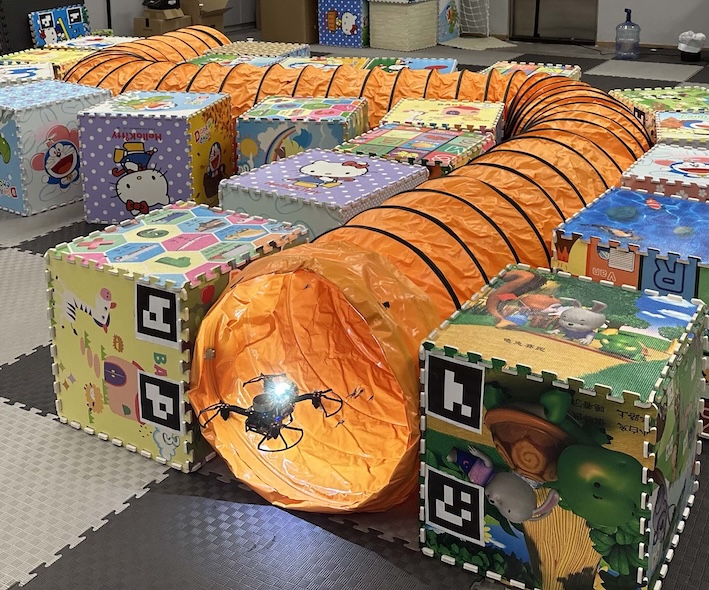}} 
%\vspace{-0.2cm}
\subfigure[\label{fig:tunnel_2d} 2-D tunnel case 3 with different cross-section shapes.]
{\includegraphics[height=0.36\columnwidth]{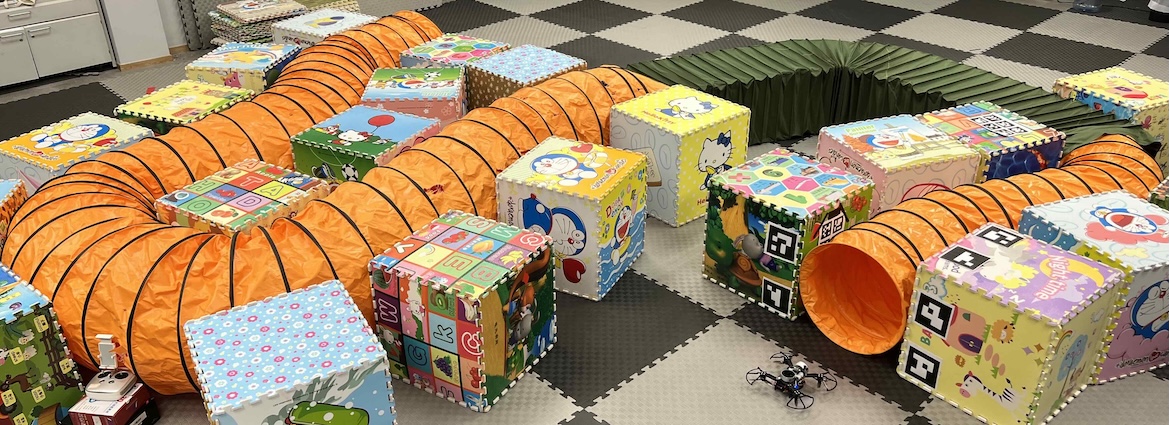}} 
\subfigure[\label{fig:tunnel_3d_circle_7} 3-D tunnel case 1.]
{\includegraphics[height=0.34\columnwidth]{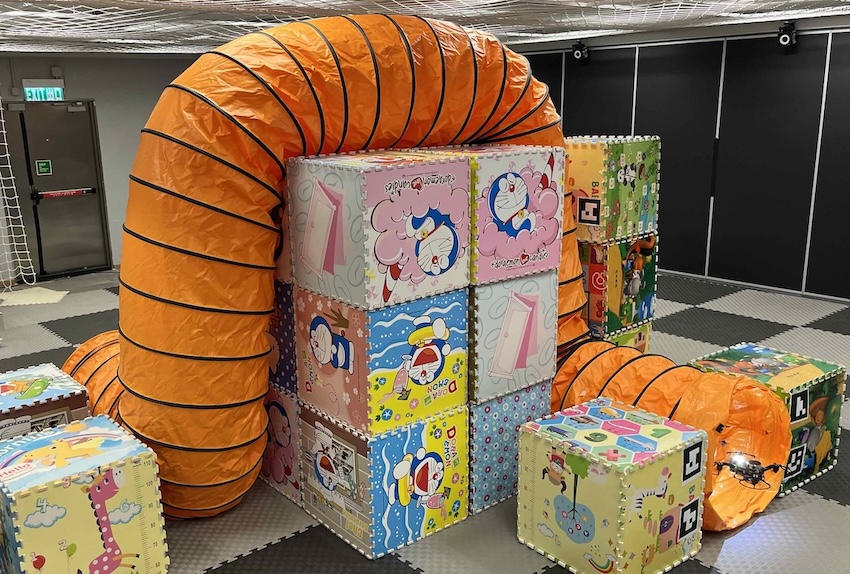}}
\subfigure[\label{fig:tunnel_3d} 3-D tunnel case 2 with different cross-section shapes.]
{\includegraphics[height=0.34\columnwidth]{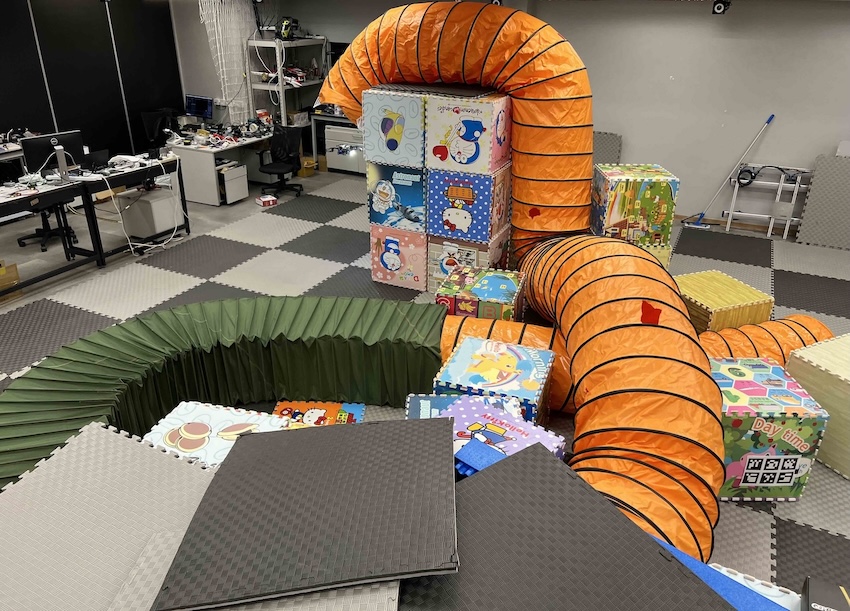}} 
\subfigure[\label{fig:vent} Rigid vent pipe with various cross-section sizes and sharp turns.]
{\includegraphics[height=0.37\columnwidth]{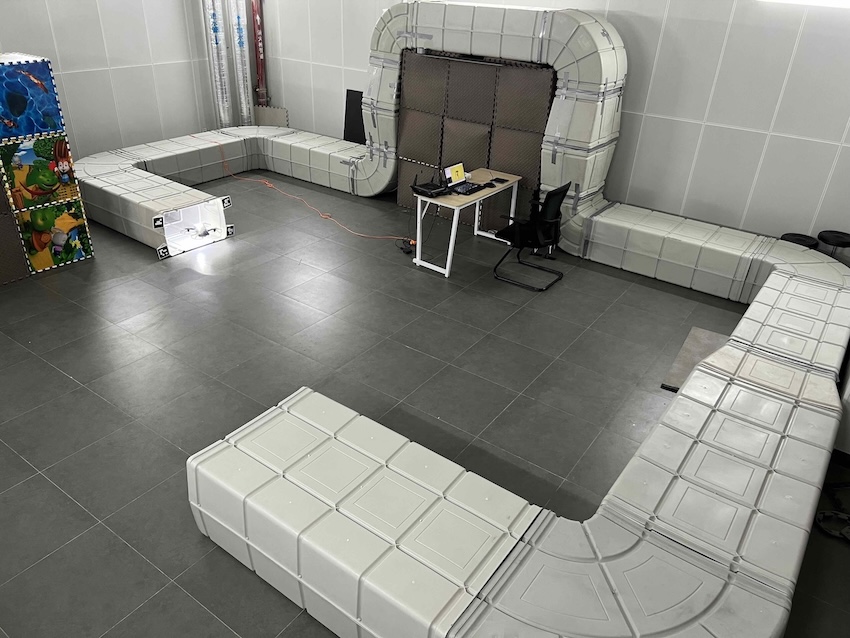}} 
\subfigure[\label{fig:tunnel_ssl} A vent pipe on a construction site.]
{\includegraphics[height=0.37\columnwidth]{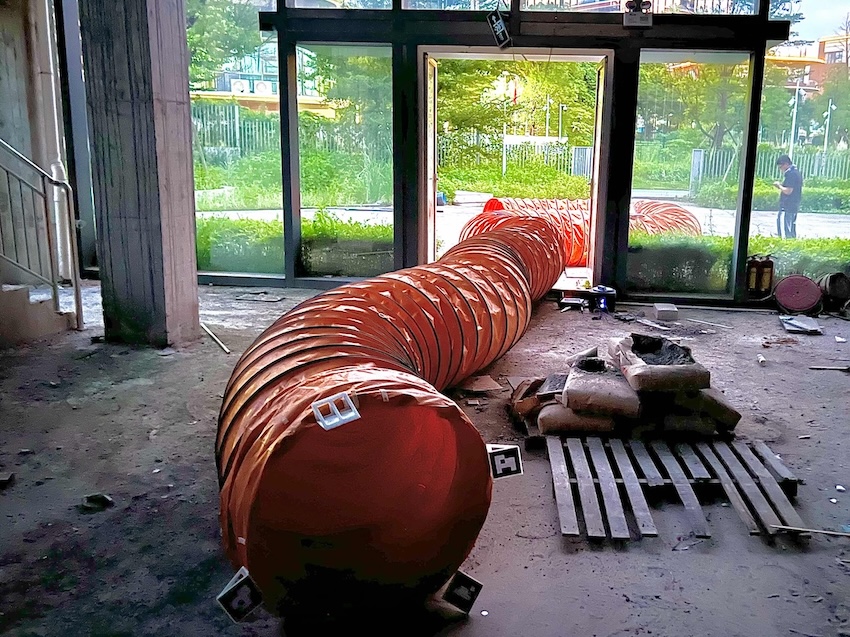}} 
\end{center}
\vspace{-0.4cm}
\caption{\label{fig:tunnel} The narrow tunnels to test the proposed autonomous aerial system.}
\vspace{-0.5cm}
\end{figure}

\IEEEPARstart{I}{n} recent years, micro aerial vehicles (MAVs), particularly multirotors, have been widely adopted in various types of applications, including inspection\cite{mathe2016vision}, search \& rescue\cite{luqi2018collaborative}, and surveillance\cite{manyam2017surveillance}, on account of their agility and compact size. These attributes enable multirotors to access confined and narrow spaces that are barely accessible to humans and ground vehicles during those missions, especially in unstructured or indoor environments.

Despite the theoretical capability to navigate in narrow spaces, multirotors face significant perception and control challenges within such environments, which have barely been addressed in previous research. In this paper, we focus on one of the most common and desirable yet extremely challenging scenarios: multirotor flights in narrow tunnel-like structures, such as drainage and ventilation conduits or other types of pipelines. As highlighted in previous studies\cite{ozaslan2017autonomous,vong2019integral,wang2022neither}, navigating these narrow tunnels presents substantial difficulties in terms of perception and control:

\begin{itemize}

\item In addition to the absence of a global positioning system, the lack of geometric features and external illuminations, as well as the narrow space at corners challenges the perception system with a restricted field of view (FoV).

\item The strong and complicated ego airflow disturbances intensively challenge the control system of multirotors in these confined and restricted spaces, where the disturbances from the complex and chaotic aerodynamic proximity effects can hardly be modeled with simple equations.

\end{itemize}

Generalized aerial systems using state-of-the-art motion planning methods, for instance \cite{zhou2019robust}, are designed for navigating wider indoor or outdoor areas, rather than the confined narrow tunnels. They are tested to be impracticable in such narrow areas, as indicated in \cite{wang2022neither}. Therefore, a customized aerial platform with proper navigation method is required to conquer the perception and control difficulties and navigate the narrow tunnels. Although the system proposed in \cite{wang2022neither} offers a potential strategy for navigating certain narrow tunnels, it is primarily designed for near-horizontal tunnels with minor curvatures and unchanged cross-sections that are often overly idealized. Additionally, the system operates at a constant speed without considering variations in tunnel shapes, and the complicated speed selection procedure through numerous flights takes a large amount of labor effort, making deployment difficult.

%Although maintaining a proper speed along the tunnel centerline and utilizing visual-inertial odometry with LEDs, as well as adopting an RGB-D camera can partially address these issues in limited scenarios, the system remains impractical for broader applications\cite{wang2022neither}. It primarily considers near-horizontal tunnels with minor curvatures and unchanged cross-sections which are usually too idealistic. In typical applications, the centerline of general tunnels may not be extracted properly, the selected ”proper” flight speed may not be valid, and the perception system may fail to comprehend the surrounding environment due to the limited FoV of the camera. Additionally, the system operates at a constant speed without considering variations in tunnel shapes, and the complicated speed selection procedure through numerous flights takes a large amount of labor effort, making deployment difficult.

To conquer the aforementioned challenges, we develop an autonomous aerial system distinct from \cite{wang2022neither} to handle more complex and general scenarios. The newly developed system features a virtual omni-directional perception module and a novel perception-and-disturbance-aware planning algorithm to navigate through narrow tunnels of varying cross-sectional shapes and sizes. It can also accommodate different directions and corner geometries, which covers most real-world scenarios. The virtual omni-directional perception module is capable of sensing the surrounding environment forward, upward, and downward, and can achieve omni-directional perception through yaw movements of the quadrotor. It can be functional in tunnels with any slope while maintaining a compact and lightweight design to ensure traversability and sufficient power margin. Additionally, a novel perception-and-disturbance-aware tunnel planning framework, which utilizes a novel perception and ego airflow disturbance model is proposed. The planner, along with the virtual omni-directional perception module, is integrated into a quadrotor platform and tested in various narrow tunnels to validate the practicability of the entire system. All the algorithms and modules operate online in real-time to ensure practicability. Finally, the systematic framework is distilled into a comprehensive pipeline for migration to other multirotor platforms for further deployment.

The contributions of this paper are summarized as follows:

\begin{enumerate}
	\item A virtual omni-directional perception module comprising state estimation, mapping, and tunnel center waypoint extraction, tailored for narrow tunnel flights.
	\item A novel perception-and-disturbance-aware motion planning framework that enables a quadrotor to navigate safely and smoothly through narrow tunnels of various cross-sectional shapes and in arbitrary directions. 
	\item A complete real-time and online narrow tunnel autonomous flight system, which includes the integration of the tailored perception, planning, as well as control and other necessary modules, on a customized quadrotor platform with compact design.
	\item Extensive flight experiments and comparison to validate the entire system in real and challenging narrow tunnel scenarios, as shown in Fig.~\ref{fig:tunnel}.
	
%	Extensive flight tests and comparisons in realistic and challenging narrow tunnels shown in Fig.~\ref{fig:tunnel} to validate the entire system.
	
%	Comprehensive integration of the narrow tunnel autonomous flight system, which includes the proposed motion planning framework, along with virtual omni-directional RGBD-inertial state estimation and mapping, as well as control modules. The real-time and online system is implemented on a customized quadrotor platform equipped with illuminations and with compact design. Extensive flight tests in realistic and challenging narrow tunnels shown in Fig.~\ref{fig:tunnel} are conducted to validate the planning method and the robustness of the entire system. \textbf{To the best of the authors' knowledge, this is the first known autonomous quadrotor system capable of flying through realistic tunnels in arbitrary directions and as narrow as 0.5 m, outperforming human pilots}.
	\item A complete pipeline to extend the developed framework to other multirotor platforms, accommodating different multirotor sizes and more tunnel cross-section shapes. Additionally, all components in the system are released as open-source packages for future developments.
\end{enumerate}

\textbf{To the best of the authors’ knowledge, this is the first known autonomous quadrotor system capable of flying through real tunnels in arbitrary directions and as narrow as 0.5 m in cross-section diameter, which outperforms human pilots.}

%The rest of the paper is organized as follows: Sec.~\ref{sec:related_work} covers related work. In Sec.~\ref{sec:system}, we provide an overview of our autonomous system. Sec.~\ref{sec:tunnel_perception} details the perception module for narrow tunnel flights. Sec.~\ref{sec:disturbance_perception_cost} formulates the perception and ego airflow disturbance models, which facilitates our perception-and-disturbance-aware planning approach described in Sec.~\ref{sec:tunnel_planning}. In Sec.~\ref{sec:result}, we provide the experimental results. In Sec.~\ref{sec:extendability}, we discuss extendability and possible improvements of the system. Finally, Sec.~\ref{sec:conclusion} concludes the paper.

The rest of this paper is organized as follows: In Sec~\ref{sec:related_work}, we introduces related work about multirotor narrow tunnel flights. In Sec.~\ref{sec:system}, we provide an overview of our autonomous system and the tunnel flight workflow. Sec.~\ref{sec:tunnel_perception} gives details of the virtual omni-directional perception module, including state estimation, mapping, and tunnel center waypoint extraction. In Sec.~\ref{sec:disturbance_perception_cost}, we formulate the perception and ego airflow disturbance models, which facilitate our perception-and-disturbance-aware planning approach described in Sec.~\ref{sec:tunnel_planning}. Sec.~\ref{sec:extendability} describes the experiment setup in multiple scenarios and provides results. In Sec.~\ref{sec:extendability}, we discuss the extendability and possible future improvement of the system, and finally Sec.~\ref{sec:conclusion} concludes the paper.

\section{Related Work}
\label{sec:related_work}
\subsection{Ego Airflow Disturbances in Narrow Tunnels}
Ego airflow disturbances strongly impact rotorcraft flight dynamics in narrow tunnels. These disturbances are primarily due to proximity effects, which are more significant and complex within narrow tunnels compared to open areas. These effects have been extensively studied by researchers for several decades. In general, proximity effects can be categorized into ground, ceiling, and wall effects. The ground effect model, initially proposed by Betz\cite{betz1937ground} and Chesseman\cite{cheeseman1955effect}, reveals that when a rotorcraft flies near ground, additional lift force will be induced. The model is widely adopted by researchers and is analyzed on small scale rotorcrafts in recent years \cite{eberhart2017modeling, conyers2019empirical, kan2019analysis, britcher2021use}. This well-documented model also provides operational guidelines for real applications \cite{powers2013influence, danjun2015autonomous}. Similarly, the ceiling effect model has also been developed and analyzed for years \cite{johnson2012helicopter, britcher2021use}, which reveals that addition lift force will also be generated to pull the rotorcraft up when it flies under a ceiling. The model has also been integrated into flights\cite{nishio2020stable} to enhance safety and control performance. Another proximity effect, the wall effect, characterized by induced forces and torque pushing the rotorcraft towards nearby walls, has also been analyzed extensively\cite{robinson2014computational, conyers2019empirical, britcher2021use}. However, a concise wall effect model as simple as the ground effect model has yet to be established due to its complexity. The disturbances generated from the proximity effects disrupt the control of flights, and can pose significant safety risks. Although in a broader area, the large disturbances near the obstacles can be modeled or avoided\cite{wang2021estimation}, in a confined narrow tunnel, the intensified disturbances become more complicated, invalidating the proximity effect models, and there is no space to avoid the high-disturbance zones. Consequently, navigating narrow tunnels remains a formidable challenge.

\subsection{Impact of Planning on Perception and Control in Narrow Tunnel Flights}
Planning acts as a bridge between perception and control, impacting both during autonomous flights. In narrow tunnels, maintaining a safety margin by aligning the planned trajectory with the centerline is essential due to significant airflow disturbances and limited maneuvering space. In this context, speed profile planning becomes critical. High-speed flights is usually considered to pose significant challenges for both perception and control systems. High velocities induce issues like severe motion blur and substantial parallax, complicating stable feature extraction and tracking\cite{mueggler2014event}. Moreover, a low-latency computation required in high-speed flights is also hard to achieve\cite{shen2013vision}. Additionally, control performance generally degrades with increasing flight speed due to modeling inaccuracies and the physical limitations of multirotors. As indicated in \cite{morrell2018comparison} and \cite{fridovich2018planning}, higher flight speeds lead to larger control errors, increasing the risk of collisions near obstacles. As a result, strategies such as adjusting safety margins for flights at different speeds\cite{fridovich2018planning} or adjusting speed based on available space\cite{quan2021eva} are proposed to mitigate the risk. However, previous studies have shown that slow flight speeds near obstacles can intensify ego airflow disturbances caused by proximity effects, potentially causing larger control tracking errors than those at higher speeds\cite{wang2021estimation}. Thus, flying at slow speeds and neglecting proximity effects can also be risky during flights. In narrow tunnels, where perception is difficult and ego airflow disturbances are extremely severe, it is crucial to maintain appropriate flight speeds by considering both factors\cite{wang2022neither}.

\subsection{Multirotor Systems in Tunnel-like Confined Areas}
Although multirotor flights in cluttered environments have been extensively studied\cite{chen2016online,loianno2016estimation,zhou2019robust,gao2020teach}, the exploration of multirotor capabilities in tunnel-like confined areas remains limited due to intense challenges. As highlighted in prior studies \cite{ozaslan2017autonomous,petrlik2020robust}, difficulties in state estimation within tunnels arise due to the lack of light and geometry features. Additionally, control issues are exacerbated by significant ego airflow disturbances resulting from proximity effects\cite{vong2019integral}. Previous attempts to develop tunnel navigation systems have been inadequate. For instance, a minimum-time tunnel-following navigation was proposed in \cite{arrizabalaga2022towards} to follow a tunnel, while the authors overlooked the critical perception and ego airflow disturbance issues and relied solely on simulation results, making it impractical for real-world application. Another tunnel navigation method with an integrated system was proposed in \cite{elmokadem2021method}. However, the proposed system only follows horizontal large tunnels without addressing key perception and control issues, making it ineffective in narrow tunnels as small as 0.5 m in diameter. Although a quadrotor system in \cite{wang2022neither} was proposed with the capability of navigating narrow tunnels, as mentioned in Sec.~\ref{sec:introduction}, the system is not complete for applications in generalized narrow tunnels. It is limited to nearly horizontal tunnels with minimal curvature and consistent cross-sections, and it utilizes a constant speed planning approach that is not suitable in more general cases.

\section{System Overview}
\label{sec:system}

%\vspace{-0.7cm}

\subsection{Quadrotor Platform Design}
\label{subsec:quadrotor}

%\begin{figure}[t]
%\begin{center}
%\includegraphics[width=0.98\columnwidth]{drone}
%\subfigure[\label{fig:drone}The customized quadrotor platform.]
%{\includegraphics[width=0.98\columnwidth]{drone}}             
%\subfigure[\label{fig:system}The system architecture and the workflow.]
%{\includegraphics[width=0.98\columnwidth]{system}} 
%\end{center}
%\vspace{-0.3cm}
%\caption{\label{fig:hard_soft}The hardware and software system architecture, together with the workflow for the quadrotor to fly through narrow tunnels.}
%\vspace{-0.3cm}
%\end{figure}

Given the impracticality of installing external localization systems in all potential tunnel environments during real-world deployment, an onboard perception system is essential for autonomous tunnel navigation. As mentioned in \cite{wang2022neither}, the tunnel’s internal environment typically lacks geometry features and external illumination, making visual-inertial odometry (VIO) together with supplementary LED lighting the most viable state estimation strategy at present. Additionally, the significant variations in illumination conditions at tunnel entrances and exits can severely disrupt vision-based depth estimation, making a relatively light-insensitive LiDAR a practical solution for the perception module. Consequently, we employ LiDAR-based RGB-D cameras as our primary perception sensor.

\begin{figure}[t]
\begin{center}
{\includegraphics[width=1.0\columnwidth]{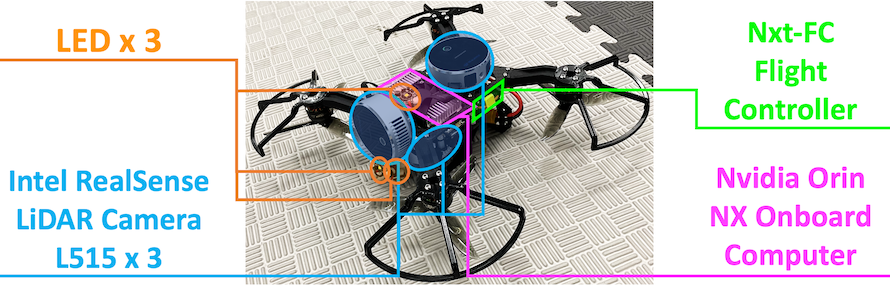}} 
\vspace{-0.4cm}
\caption{\label{fig:drone}The customized quadrotor platform for narrow tunnel flights.}
\end{center}
\vspace{-0.9cm}
\end{figure}

To ensure safe navigation through narrow tunnels with limited maneuvering space, it is crucial to maintain a flight path along the tunnel's centerline to gain maximum possible safety margins. Furthermore, tunnels may extend in various directions in practice, including vertically, the quadrotor must possess omni-directional perception capabilities to safely navigate such confined spaces.
Given that only yaw motion can be utilized for turning, the perception module still requires sensing in three directions: front, upward, and downward, as the minimal combination to achieve virtual omnidirectional perception.
%Considering the only feasible yaw direction for maneuvering for a quadrotor during narrow tunnel flights, the perception system must cover at least the front, upward, and downward directions to achieve virtual omni-directional sensibility. 

Therefore, we design a customized quadrotor platform equipped with three RGB-D camera units and three LEDs, oriented in the aforementioned directions. Additionally, the platform is designed with a compact and light structure, as shown in Fig. \ref{fig:drone}. It utilizes 5-inch propellers, features a diameter of 40 cm, a wheelbase of 25 cm, and a total weight of just 1.085 kg, despite carrying more equipment.

\begin{figure*}[t]
\begin{center}
%\captionsetup{justification=centering}
{\includegraphics[width=1.8\columnwidth]{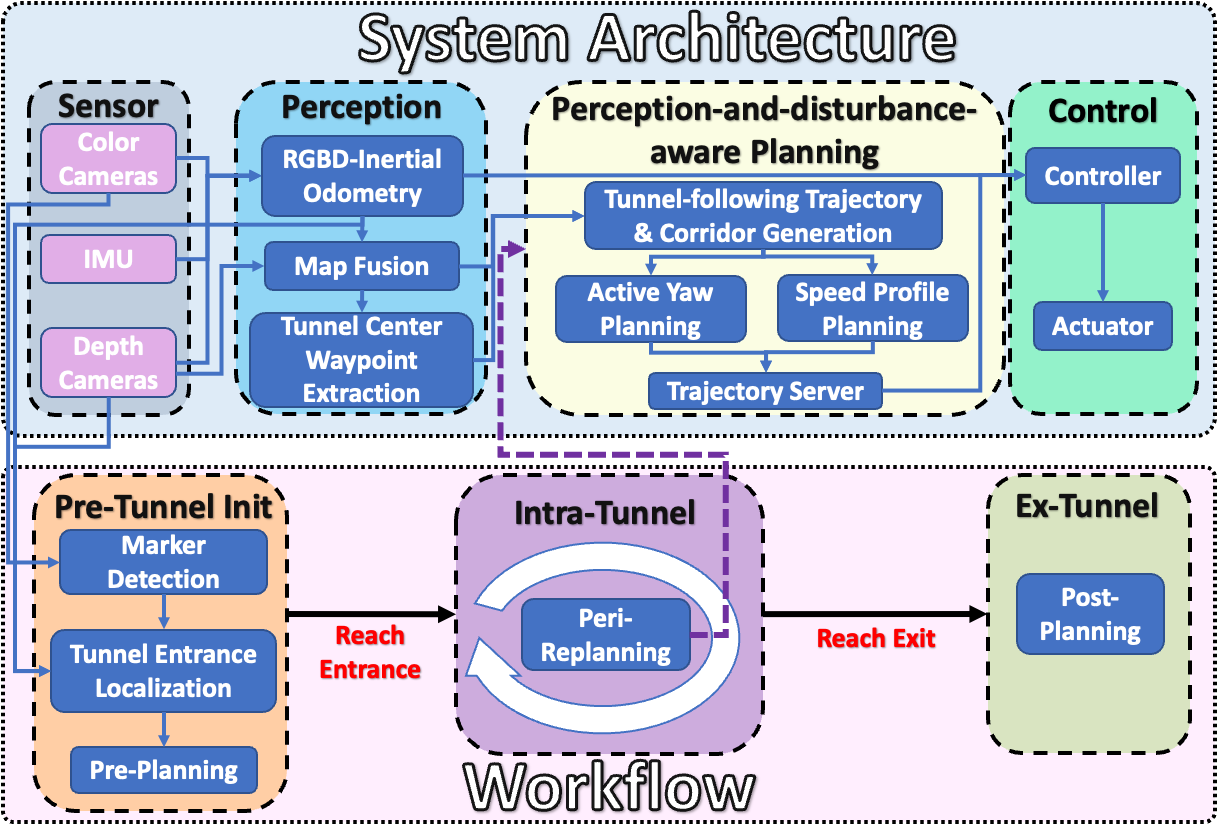}} 
%\vspace{-0.4cm}
\caption{\label{fig:system}The system architecture and the workflow for narrow tunnel flights. The perception module adopts the color and depth images, as well as IMU data to produce RGBD-inertial odometry for state estimation and then performs map fusion. The tunnel center waypoints extracted from the mapping result facilitate the perception-and-disturbance-aware planning, which generates flight trajectories, yaw trajectories and speed profiles. The results are then utilized to generate control commands and executed by the motors. The workflow consists of pre-tunnel initialization, which perform pre-planning to the localized tunnel entrance according to the detected marker, intra-tunnel peri-replanning and ex-tunnel post-planning.}
\vspace{-0.4cm}
\end{center}
\end{figure*}

\subsection{System Architecture}
\label{subsec:sys_arch}

To facilitate narrow tunnel flights using the customized quadrotor platform, a comprehensive software framework shown in Fig. \ref{fig:system} is integrated on the onboard computer. The framework consists of perception, planning, and high-level control. To fully utilize the color and depth images from multiple camera modules, we developed an RGBD-inertial state estimation system that fuses these images with IMU data. (Sec. \ref{subsec:state_estimation}). The fused RGBD-inertial odometries and depth images are then fed to the map fusion module to update occupancy map and the corresponding Euclidean distance field (EDF) (Sec. \ref{subsec:mapping}). The updated EDF then facilitates the tunnel center waypoint extraction (Sec. \ref{subsec:center_point_extraction}). The perception-and-disturbance-aware planning module leverages a perception (Sec. \ref{subsec:perception_factor}) and an ego airflow disturbance model
(Sec. \ref{subsec:ego_disturbance}) to generate trajectories from the extracted waypoints. The planning module performs flight trajectory and corridor generation (Sec. \ref{subsec:tunnel_traj}), active yaw planning (Sec. \ref{subsec:yaw_planning}) and speed profile planning (Sec. \ref{subsec:speed_planning}). Subsequently, the position, velocity, and acceleration commands are generated from the planned results and sent to the high-level controller. Then, the high-level controller generates corresponding attitude and thrust commands, which are fed to the flight controller to produce the necessary low-level commands executed by the motors.

\subsection{Tunnel Flight Workflow}
\label{subsec:workflow}

\begin{figure}[t]
\begin{center}
\subfigure[\label{fig:marker}The detected ArUco markers and the corresponding marker IDs.]
{\includegraphics[width=0.48\columnwidth]{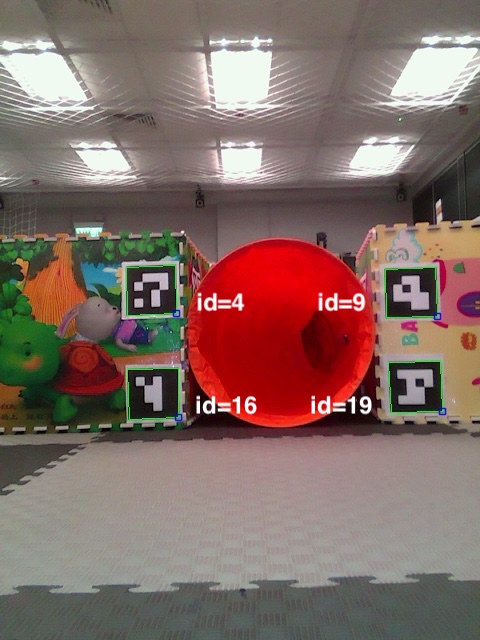}}             
\subfigure[\label{fig:entrance_viz}The visualization of the detected entrance and the voxel map.]
{\includegraphics[width=0.48\columnwidth]{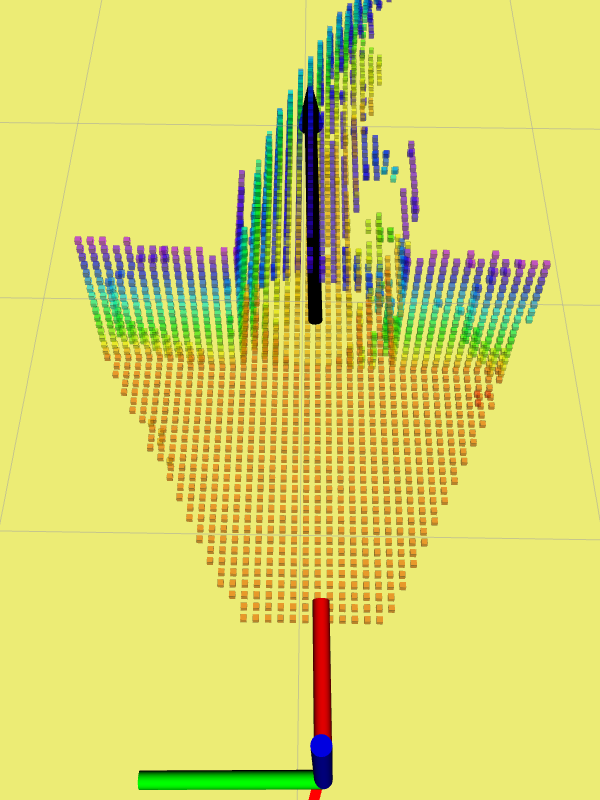}} 
\end{center}
\vspace{-0.3cm}
\caption{\label{fig:entrance}The illustration of entrance localization of the tunnel shown in Fig. \ref{fig:tunnel_2d}. The color code indicates the height; the axis indicates the pose of the quadrotor and the black arrow indicates the estimated tunnel entrance pose.}
\vspace{-0.3cm}
\end{figure}

As the quadrotor needs to enter and exit a narrow tunnel safely and smoothly, a workflow containing pre-tunnel initialization, intra-tunnel replanning and ex-tunnel post-planning shown in Fig. \ref{fig:system} is developed. 
\begin{itemize}

\item \textbf{Pre-tunnel initialization:}
Before the flight, four ArUco markers are symmetrically placed at the tunnel entrance to aid in entrance localization, as shown in Fig. \ref{fig:entrance}. The quadrotor utilizes the captured color and depth images, along with the estimated camera poses to detect the markers and calculate their global 3-D positions. Then, a Random Sample Consensus (RANSAC) algorithm is applied to the marker positions to reject outliers and ensure accuracy. The position and direction of the tunnel entrance are determined by averaging the four filtered marker positions and calculating the normal of the least-squares error plane constructed from these positions, respectively. This estimated tunnel entrance pose allows the quadrotor to generate a smooth B-spline trajectory towards it, completing the pre-tunnel initialization phase.

%Before the flight, four ArUco markers are symmetrically placed at the tunnel entrance to aid in entrance localization, as shown in Fig. \ref{fig:entrance}. The quadrotor utilizes its cameras to detect the 2-D positions of these markers in the captured color images. Subsequently, it calculates their global 3-D positions based on the detected 2-D positions and the corresponding depths provided by depth cameras, as well as the estimated camera poses. To ensure the accuracy of these detections, a Random Sample Consensus (RANSAC) algorithm is applied to reject outliers in the marker positions across the incoming frames. Once sufficient high-quality detections are confirmed, the entrance position of the tunnel is determined by averaging the four filtered marker positions. Additionally, the entrance direction is determined by calculating the normal of the least-squares error plane constructed from the four marker positions, using the singular value decomposition (SVD). This calculated tunnel entrance pose allows the quadrotor to generate a smooth B-spline trajectory towards it, completing the pre-tunnel initialization phase.

\item \textbf{Intra-tunnel replanning:}
Upon reaching the tunnel entrance, the quadrotor switches to the intra-tunnel state. It starts the mapping module and extract the tunnel shape to perform perception, while initiates an ego-airflow-disturbance-and-perception-aware replanning process that continues throughout the tunnel traversal. Details on perception and planning will be elaborated in Sec. \ref{sec:tunnel_perception} and \ref{sec:tunnel_planning}, respectively.

\item \textbf{Ex-tunnel post-planning:}
As the quadrotor approaches the tunnel exit, a smooth B-spline trajectory is generated to safely decelerate it during the post-tunnel phase. This ensures a smooth transition from the narrow tunnel to the wider external area.
 
\end{itemize}

\section{Perception in Narrow Tunnels}
\label{sec:tunnel_perception}

The quadrotor must have perception capabilities to navigate narrow tunnels, where no assistive infrastructure is available. In such constrained space, efficient and accurate perception is necessary to ensure flight safety. Therefore, the virtual omni-directional perception module needs to fully utilize the information from the three RGB-D cameras oriented forward, upward, and downward. In this section, we will elaborate on the details of this module, covering state estimation, mapping, and tunnel center waypoint extraction.

\subsection{State Estimation in Narrow Tunnels}
\label{subsec:state_estimation}
%We develop an optimization-based asynchronous multi-camera-IMU state estimator for ego state estimation within narrow tunnels. The estimator comprises multiple parallel front-ends, a front-end coordinator, and an optimization back-end. The IMU front-end performs pre-integration using IMU data and directly outputs high rate odometry based on the latest optimization results. Each visual front-end processes the color and depth images from individual camera modules asynchronously to extract and track features. The processed feature and depth measurements are fed to the front-end coordinator, which dynamically redistributes the number of features processed by each front-end based on the collected features, and determines whether the measurements are forwarded to the sliding window in the back-end optimization module according to their priority. If the asynchronous measurements have sufficient priority, they are inserted into a state history buffer and chronologically ordered by the time stamp for optimization. The optimization objective is formulated as a combination of the prior factor, the IMU propagation factor and the visual and depth factor. The estimator efficiently leverage the color and depth information from the RGB-D camera modules to perform virtual omni-directional state estimation, which is functional inside narrow tunnels. Details about the estimator can be found in \cite{wang2024vinsmulti}.

\begin{figure}
    \centering
    \includegraphics[width=0.95\columnwidth]{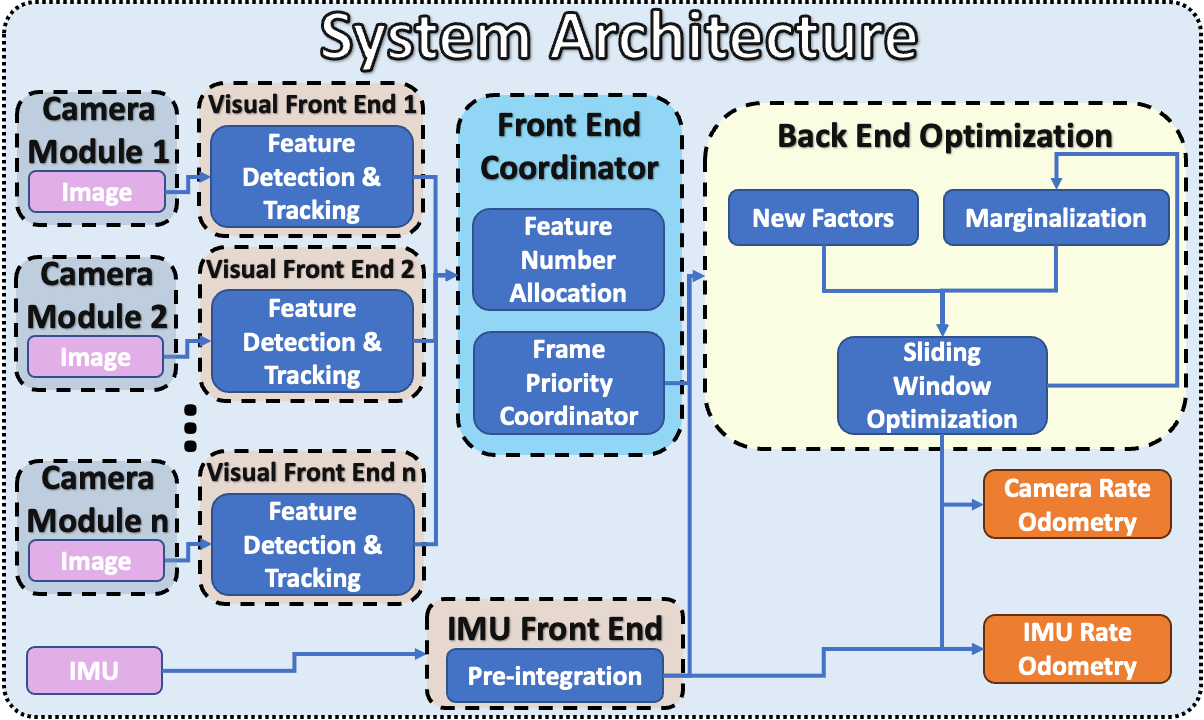}
    \caption{The system architecture of the state estimator in narrow tunnels. The estimator processes color and depth images from individual camera modules to produce RGBD-inertial odometries. It includes multiple parallel front ends, a front end coordinator, and an optimization back-end. In practice, it utilizes three RGB-D camera modules.}
    \label{fig:vins_multi_system}
\end{figure}

We develop an optimization-based multi-camera-IMU state estimator based on our previous works~\cite{qin2018vins,qin2019a}. The estimator processes color and depth images from three individual RGB-D camera modules to produce RGBD-inertial odometries. The system architecture of the state estimator is illustrated in Fig.~\ref{fig:vins_multi_system}. The estimator comprises three main components: parallel front ends, a front end coordinator, and a back end optimizer.

\subsubsection{Parallel Front End}
Each visual front end independently processes color and depth images from its respective camera, performing feature detection, tracking, and outlier rejection. Concurrently, the IMU front end conducts pre-integration of inertial data, providing high-frequency odometry updates based on the latest optimized states.

\subsubsection{Front End Coordinator}
Given the different perspectives in the tunnel, the feature qualities vary across cameras. In order to efficiently utilize the features and computation resources, the front end coordinator module is introduced to dynamically manages the feature allocation and frame prioritization across all camera front ends. By utilizing a dynamic feature number allocation strategy, it adjusts the number of features each camera processes based on the quality and quantity of features detected in the previous frame. This strategy helps maintain consistent feature tracking across different camera views.

\begin{figure}
    \centering
    \includegraphics[width=0.95\columnwidth]{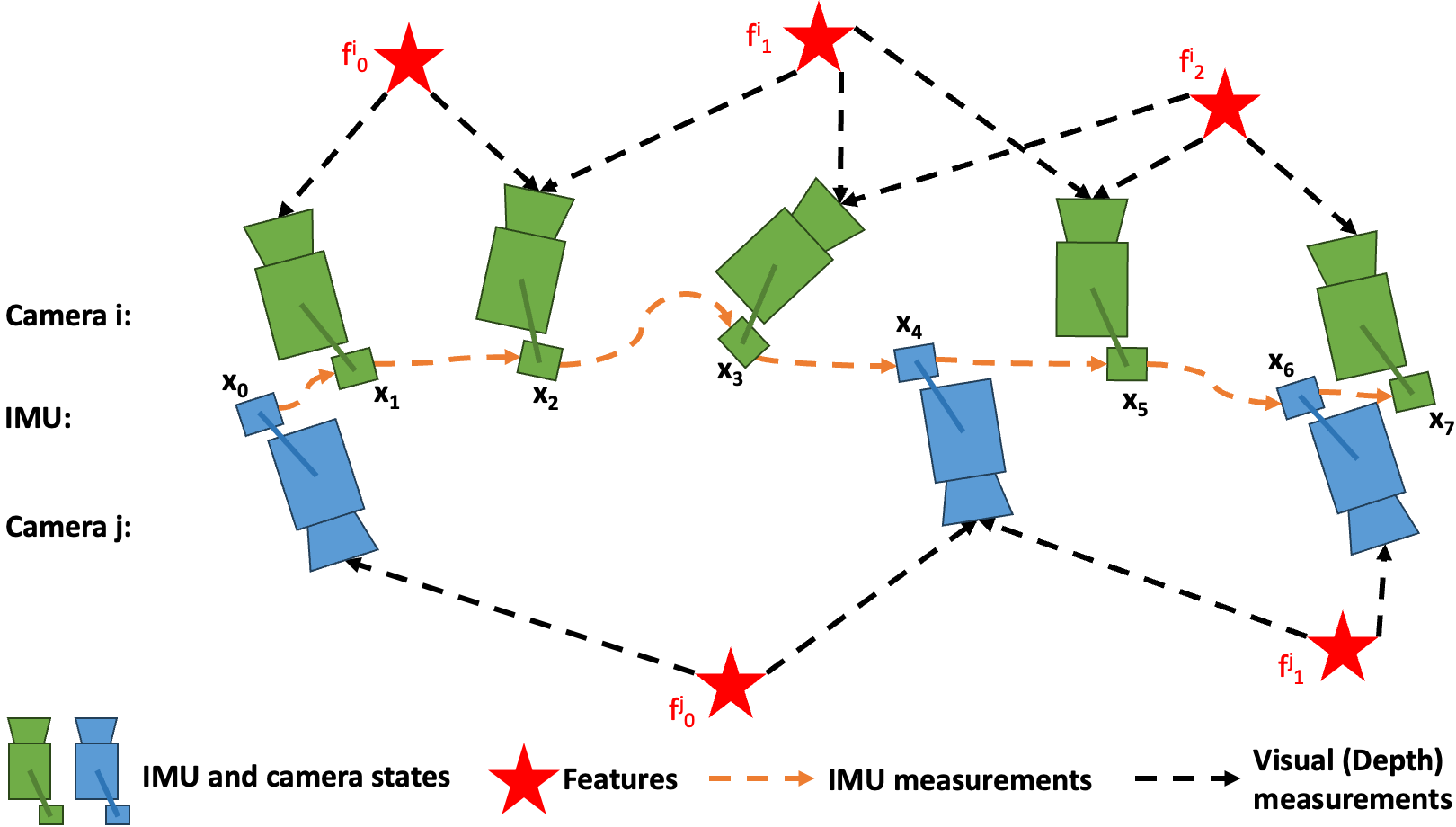}
    \caption{\label{fig:sliding_window} An illustration of the sliding window for optimization in the state estimation.}
\end{figure}

\subsubsection{Back End Optimizer}
The back end of the state estimator employs a sliding window optimization approach, integrating visual, depth, and IMU measurements to estimate the state vector, as illustrated in Fig.~\ref{fig:sliding_window}. This state vector includes the position, velocity, orientation, and sensor biases of the IMU. By chronologically ordering asynchronous frames within the sliding window, the optimizer ensures coherent and accurate state estimation. The optimization objective combines prior factors, IMU propagation factors, and visual-depth factors to refine the state estimates continuously.

By leveraging the three RGB-D camera modules and a efficient coordination mechanism, the optimization-based state estimation approach provides robust RGBD-inertial odometries during the navigation through narrow tunnels. For further details about the state estimator, we suggest the readers refer to \cite{wang2024vinsmulti}.

\subsection{Mapping in Narrow Tunnels}
\label{subsec:mapping}

To perform mapping in narrow tunnels in arbitrary directions, it is also essential to fully utilize the depth information. We developed a parallel and asynchronous mapping module based on \cite{han2019fiesta}. This module processes depth images from three cameras and RGBD-inertial odometry data from the state estimation module to construct occupancy and Euclidean Distance Field (EDF) maps inside tunnels. The projection and ray-casting procedures for each camera are assigned to separate threads, which asynchronously update the occupancy map with the corresponding camera odometries. Meanwhile, the EDF map, which stores the minimum Euclidean distance of each voxel to the nearest obstacle, is updated synchronously. This strategy efficiently manages the asynchronous depth images from the three cameras, enabling omni-directional mapping with yaw movements in narrow tunnels.

\subsection{Tunnel Center Waypoint Extraction}
\label{subsec:center_point_extraction}

\begin{figure*}[t]
\begin{center}         
{\includegraphics[width=0.99\columnwidth]{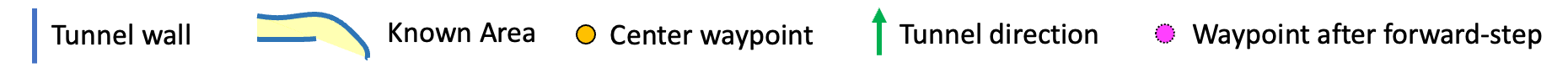}}
{\includegraphics[width=0.99\columnwidth]{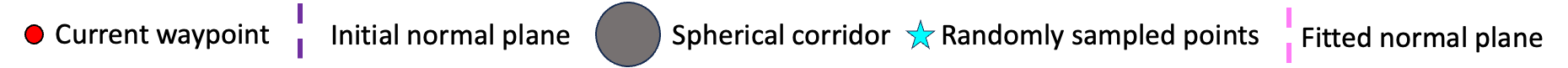}}
\subfigure[\label{fig:center_waypoint_1} Initial center waypoint and direction.]
{\includegraphics[width=0.39\columnwidth]{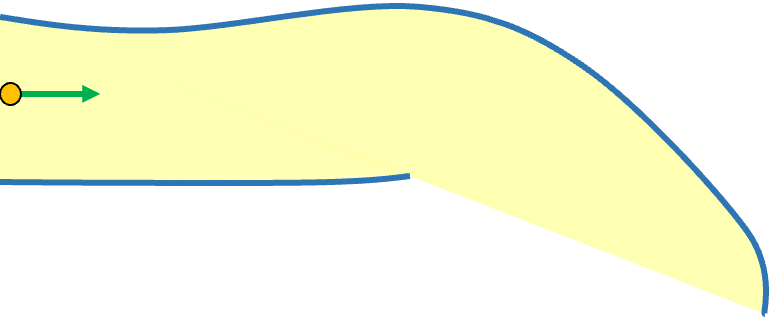}}
\subfigure[\label{fig:center_waypoint_2} Step forward and gradient ascend in the normal plane according to the EDF value.]
{\includegraphics[width=0.39\columnwidth]{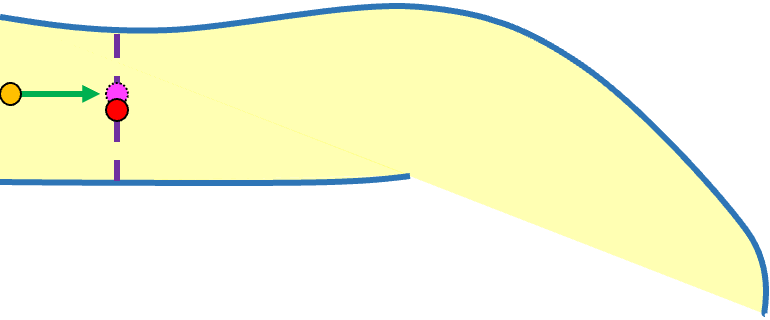}}
\subfigure[\label{fig:center_waypoint_3} Randomly sample points on the spherical corridor.]
{\includegraphics[width=0.39\columnwidth]{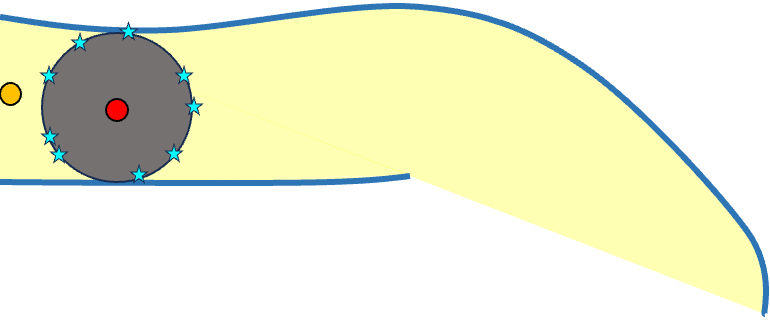}}
\subfigure[\label{fig:center_waypoint_4} Gradient descend the sampled points on the sphere according to the EDF value and fit the new tunnel direction.]
{\includegraphics[width=0.39\columnwidth]{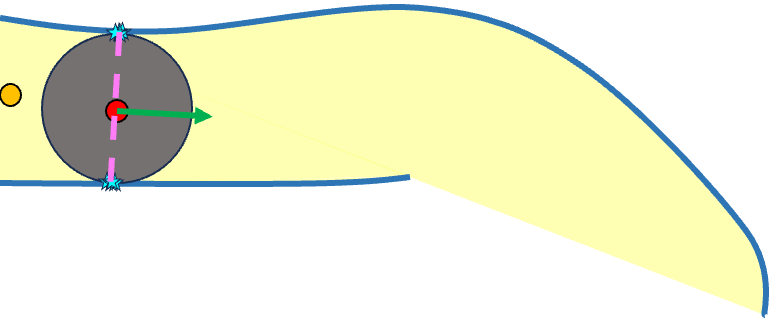}}
\subfigure[\label{fig:center_waypoint_5} Iterate (b) to (d) to obtain the final tunnel center waypoint set.]
{\includegraphics[width=0.39\columnwidth]{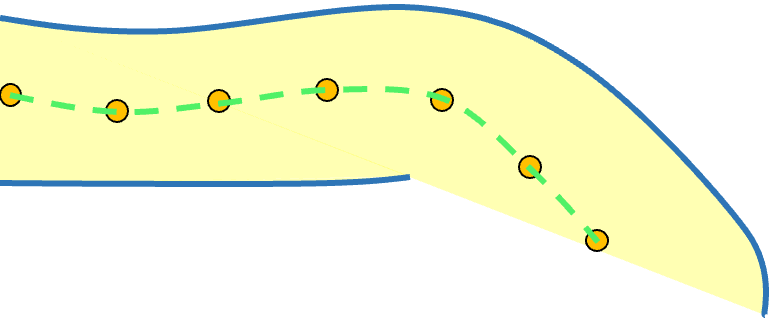}}
\end{center}
\vspace{-0.4cm}
\caption{\label{fig:center_waypoint}The illustrations of tunnel center waypoint extraction using Euclidean Distance Field (EDF). The illustration is in 2-D for a clear view.}
\vspace{-0.2cm}
\end{figure*}

\begin{algorithm}[t]
\caption{Tunnel Center Waypoint Extraction}  
\label{alg:center_waypoint}  
\begin{algorithmic}
\State \textbf{Notation}: Start point $p_0$, Start direction $dir_0$, Tunnel center waypoints $\mathcal{W}$, Point $p$, Quadrotor dimension $D_q$, Maximum tunnel dimension $D_{max}$, Search step length $S$, Search direction change threshold $DIR_{min}$, EDF value at a position $edf(\cdot)$, Point set $\mathcal{P}$, Plan distance $d_p$, Plan range $R_p$
\State \textbf{Input}: $p_0$, $dir_0$
\State \textbf{Output}: $\mathcal{W}$
\State Initialize :
\State $predict\_dir \leftarrow false$
\State $d_p \leftarrow 0$
\State $dir^- \leftarrow dir_0$
\State $dir^+ \leftarrow dir_0$
\State $d_{max} \leftarrow 0.5 \cdot D_{max}$
\State $\mathcal{W}$.push\_back($p_0$)
\State $p \leftarrow GradientAscend( p_0 + S \cdot dir^+, dir^+, d_{max})$
\While{$d_p \leq R_p \ \&\& \ is\_known(p) \ \&\& \  D_q \leq 2\cdot edf(p) \leq D_{max}$}
\State $\mathcal{W}$.push\_back($p$)
\State $\mathcal{P} \leftarrow SphereRandomSample(p)$
\ForEach {$p_i \in \mathcal{P}$}
\State $p_i \leftarrow SphereGradientDescend(p_i)$
\EndFor
\State $dir^- \leftarrow dir^+$
\State $dir^+ \leftarrow PlaneFit(\mathcal{P}, dir^-).normal()$
\If{$predict\_dir$}
\State $dir^+ \leftarrow dir^-$
\Else
\If{$dir^- \cdot dir^+ \leq DIR_{min}$}
\State $predict\_dir \leftarrow true$
\State $d_{max} \leftarrow edf(p)$
\EndIf
\EndIf
\State $p \leftarrow GradientAscend( p + S \cdot dir^+, dir^+, d_{max})$
\State $d_p \leftarrow d_p + dist(p,\mathcal{W}.back())$
\EndWhile
\State \Return $\mathcal{W}$
\end{algorithmic} 
\end{algorithm}

To ensure safety and control within narrow tunnels, it's crucial to maintain flight paths along the centerline, keeping as far from the walls as possible. This minimizes proximity effects that can cause strong external forces and unpredictable disturbances, reducing control difficulties and enhancing safety\cite{eberhart2017modeling,conyers2019empirical,wang2021estimation}.
Therefore, extracting tunnel center waypoints, which represent the geometric details of the tunnel centerline, is essential for tunnel flight operations.

The procedure for the tunnel center waypoint extraction is illustrated in Fig.~\ref{fig:center_waypoint} and detailed in Alg.~\ref{alg:center_waypoint}. It utilizes the EDF map generated by the mapping module to extract the center waypoints. By providing the start point $p_0$ and direction $dir_0$, the algorithm searches along the tunnel with a constant step length $S$ until the total planned distance $d_p$ reaches a preset plan range $R_p$, or the searching point $p$ reaches an unknown or impassable area. Specifically, traversal is considered invalid if twice the EDF value at $p$ is smaller than the quadrotor diameter $D_q$ or larger than the maximum tunnel diameter $D_{max}$. The maximum allowable EDF value $d_{max}$ is initialized to the max radius of the tunnel and the search point $p$ is initialized to the maximum EDF value position in the plane with the normal direction $dir_0$ and a distance $S$ ahead of the initial point $p_0$. Using the gradient ascent method and the maximum allowable EDF value $d_{max}$, a proper initialization can be achieved. During the gradient ascent, if the initial point has an EDF value larger than the maximum allowable EDF value, a gradient descent is first conducted until the EDF value reach the maximum allowable limit. This is particularly useful in the prediction stage when only part of the tunnel wall is observed in the local map, preventing the point from ascending unreasonably away from one side of the wall. During each iteration of the search, at a new search point $p$, a bunch of random points are sampled on the spherical corridor, which is a sphere centered at $p$ with a radius equal to the EDF value at $p$. Each sampled point $p_i$ undergoes gradient descent to the position on the sphere closest to the tunnel wall, minimizing the EDF value. The tunnel direction $dir^+$ is obtained through least-squares plane fitting of the descended points. Finally, a gradient ascent is performed on the plane with a normal direction $dir^+$ and a distance $S$ ahead of the current point to complete the loop. If the change in tunnel direction compared with the previous direction $dir^-$ is significant, it usually indicates incomplete tunnel observation. In such cases, predictions should be made from the current point along the previous direction, assuming the tunnel dimensions remain unchanged. The waypoints from each iteration form the tunnel center waypoint set $\mathcal{W}$, concluding the center waypoint extraction.

\section{Model Formulation of Perception and Ego Airflow Disturbance Factors}
\label{sec:disturbance_perception_cost}

After ensuring the perception capability, motion planning inside narrow tunnels can be discussed. However, prior to introducing the planning methodology, it is important to formulate the perception and ego airflow disturbance models, as these two factors significantly impact performance and safety during flights through narrow tunnels, as indicated in Sec.~\ref{sec:introduction}.

%As indicated in Sec. \ref{sec:introduction}, perception performance is critical to ensuring safe flights in confined narrow tunnels. Additionally, ego airflow disturbances can significantly affect control performance during flights in such environments. 

As analyzed in \cite{wang2022neither}, the perception and ego airflow disturbance factors are often coupled and exhibit opposing effects on flight performance with changes in speed, necessitating compensation during planning. However, the detailed formulation of these factors remains to be analyzed. Therefore, we first analyze the perception factor related to feature tracking, as its stability is crucial for effective perception, and model the mean optical flow speed. Next, we examine the ego airflow disturbance, which significantly impacts control performance, and model the disturbance level under different flight conditions. With these models in place, perception-and-disturbance-aware planning can be conducted afterwards.

\subsection{Perception Factor Model Formulation}
\label{subsec:perception_factor}

The experimental results in \cite{wang2022neither} demonstrate that as flight speed inside the tunnel increases, the number of tracking features decreases, indicating reduced perception performance. Since a vision-based state estimation system is used, feature tracking occurs on 2-D images. It's more accurate to conclude that the number of tracked features decreases as the optical flow speed on the images increases. Therefore, to ensure stable feature tracking and improve perception quality for safe flights, it is important to minimize the optical flow speed on the images.

\begin{figure}[t]
\begin{center}
\subfigure[\label{fig:square_6_of_sample}The quadrotor is flying in a horizontal rectangular tunnel.]
{\includegraphics[width=0.48\columnwidth]{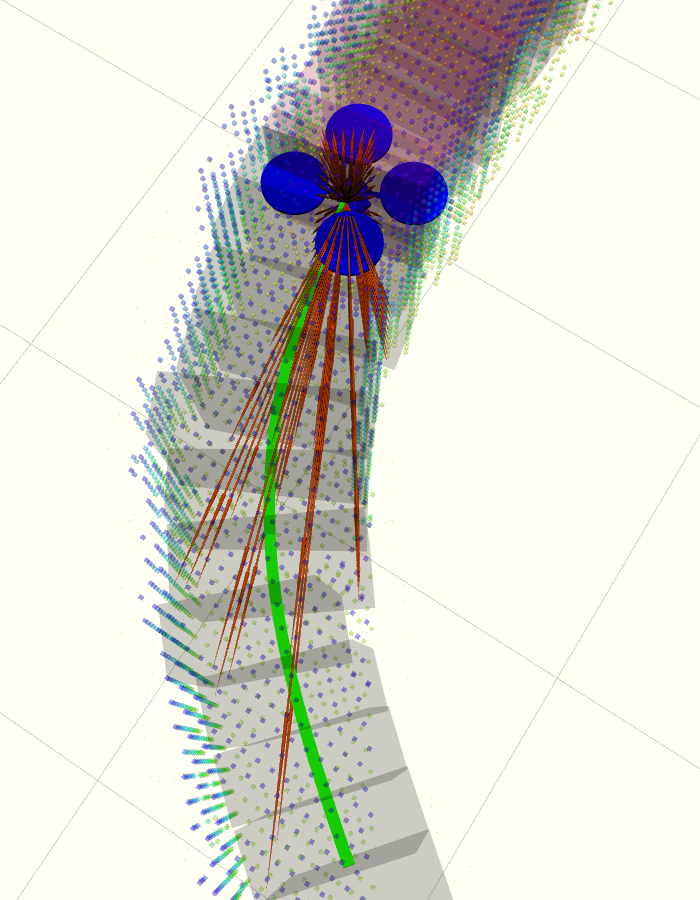}}             
\subfigure[\label{fig:circle_7_of_sample}The quadrotor is flying upward in a circular tunnel.]
{\includegraphics[width=0.48\columnwidth]{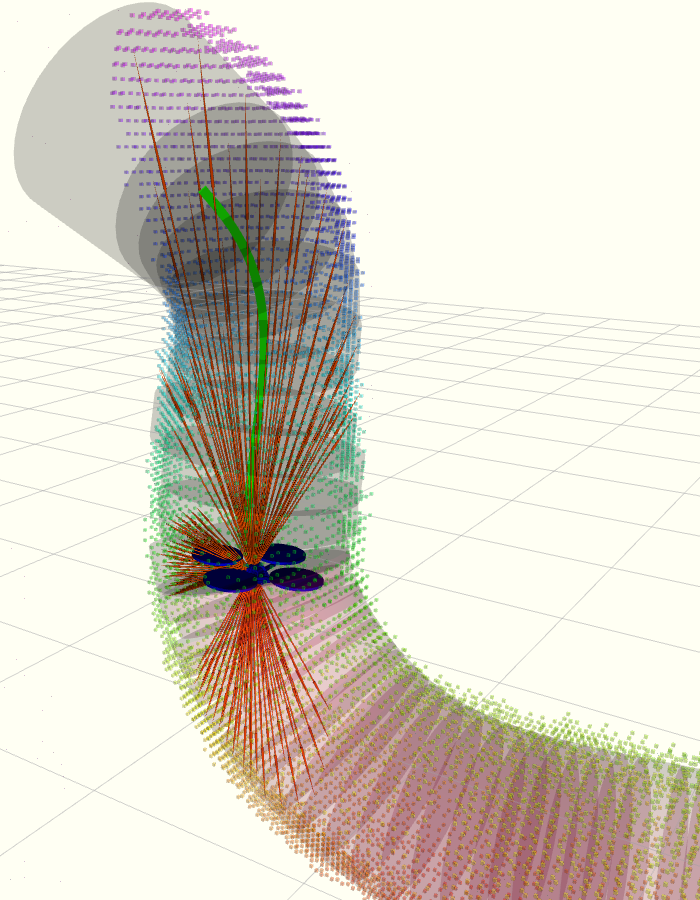}} 
\end{center}
\vspace{-0.3cm}
\caption{\label{fig:of_sample}The illustrations of the uniformly sampled feature points on the image planes and the corresponding 3-D points on the extracted tunnel walls in simulations. The green curves represent the tunnel-following trajectory, the transparent boxes and cylinders indicate the estimated tunnel cross-sections, and the red arrows indicates the corresponding 3-D positions of the sample points on the image planes. Note that the quadrotor is equipped with three cameras facing front, top, and down, and only the samples of the current position of the quadrotor is illustrated in each case.}
\vspace{-0.5cm}
\end{figure}

\begin{figure}[t]
\begin{center}
{\includegraphics[width=0.95\columnwidth]{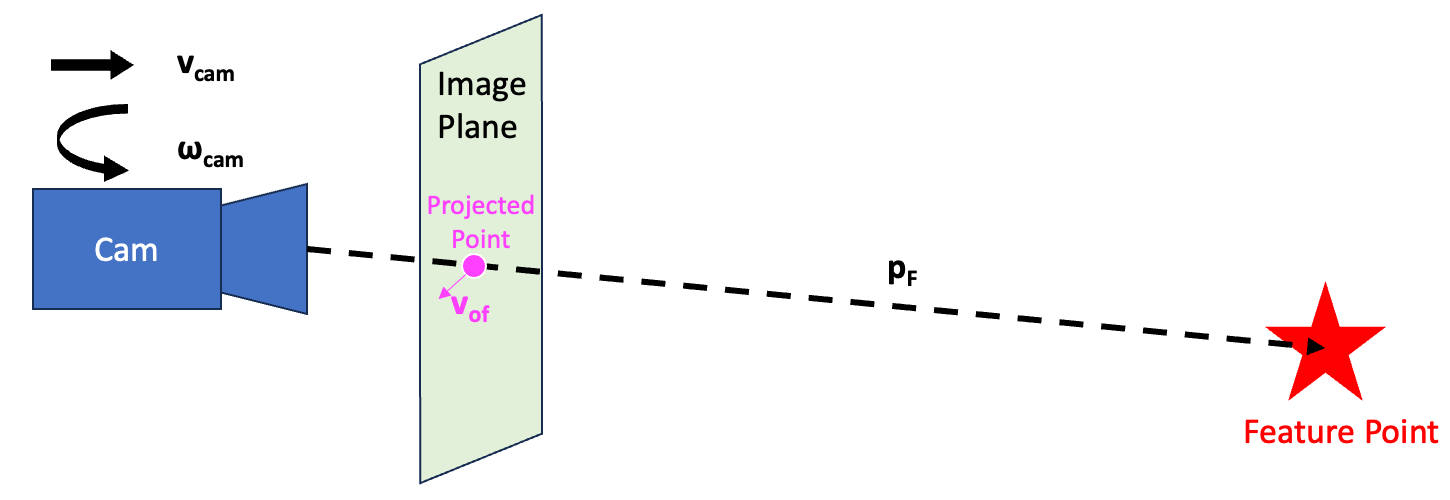}}
\end{center}
\vspace{-0.3cm}
\caption{\label{fig:projection}The illustration of the projection model for a feature point.}
\vspace{-0.1cm}
\end{figure}

Predicting the optical flow speed of future features is typically challenging, but the tunnel scenario simplifies this because the features are located on the tunnel walls. For a 2-D feature point on the image planes of the cameras along the tunnel-following trajectory (will be detailed in Sec. \ref{subsubsec:tunnel_traj_opt}), the corresponding 3-D points on the wall can be determined using ray-casting, as depicted in Fig. \ref{fig:of_sample}. For a 3-D feature point located at $\mathbf{p_f}$ in the camera frame, when the camera is moving at a linear velocity $\mathbf{v_{cam}}$ and an angular velocity $\mathbf{\omega_{cam}}$, as illustrated in Fig. \ref{fig:projection}, the optical flow velocity on the image plane $\mathbf{v_{of}}$ can be derived as:
\begin{equation}\label{eq:projection} 
	\mathbf{v_{of}}
	= proj(\mathbf{K_{cam}} \frac{-\mathbf{\omega_{cam}} \times \mathbf{p_F}- \mathbf{v_{cam}}}{p_{f_z}}),
\end{equation}
where the $proj$ function projects the vector inside the bracket onto the image plane by extracting the $x$ and $y$ elements for the conventional camera frame definition, $\mathbf{K_{cam}}$ is the projection matrix of the camera, and $p_{f_z}$ is the $z$ element of the position of the feature point in the camera frame, which also represents the depth. From the uniform sampling on the camera image plane, the mean optical flow speed can be approximated.

\subsection{Ego Airflow Disturbance Analysis and Modeling}
\label{subsec:ego_disturbance}

\begin{figure*}[t]
\begin{center}         
{\includegraphics[width=1.95\columnwidth]{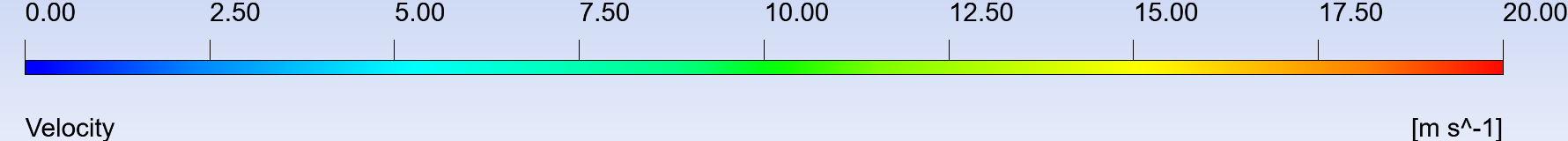}}
\subfigure[\label{fig:square_6_0deg0_cfd} Hover, horizontal tunnel with 0.6 m edge length square cross-section.]
{\includegraphics[width=0.64\columnwidth]{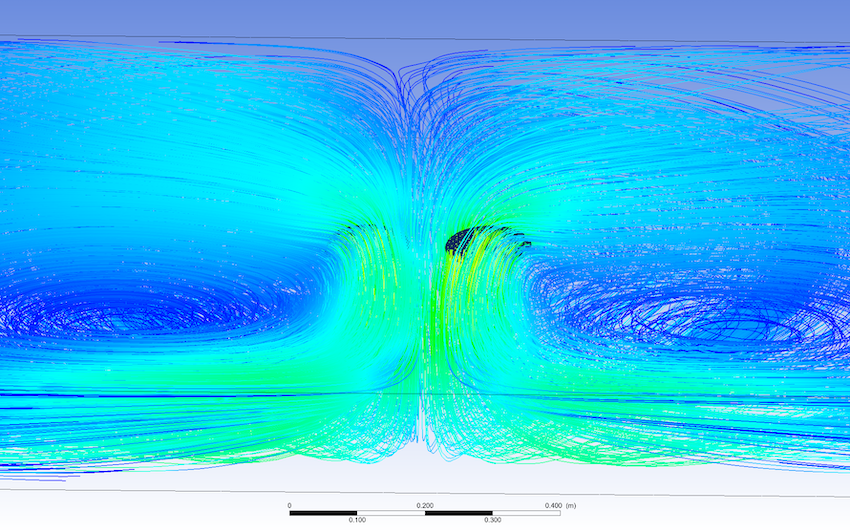}}
\subfigure[\label{fig:square_6_0deg10_cfd} 1 m/s towards left, horizontal tunnel with 0.6 m edge length square cross-section.]
{\includegraphics[width=0.64\columnwidth]{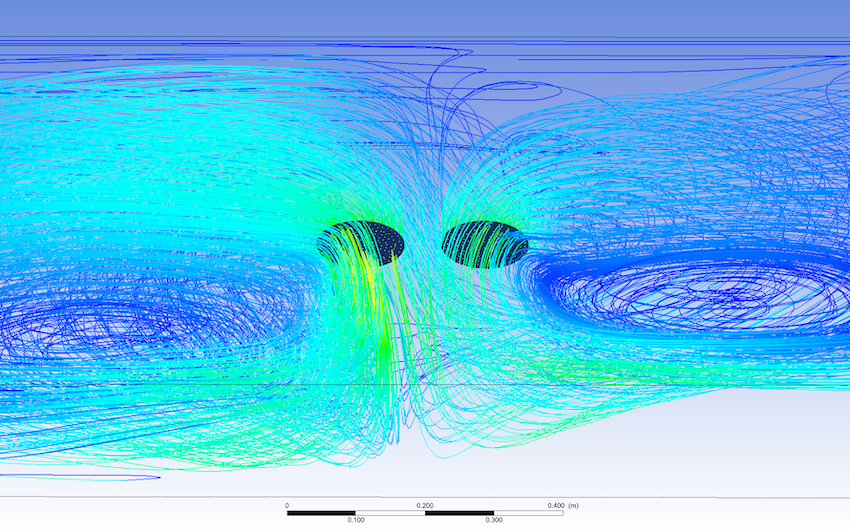}}
\subfigure[\label{fig:square_6_0deg20_cfd} 2 m/s towards left, horizontal tunnel with 0.6 m edge length square cross-section.]
{\includegraphics[width=0.64\columnwidth]{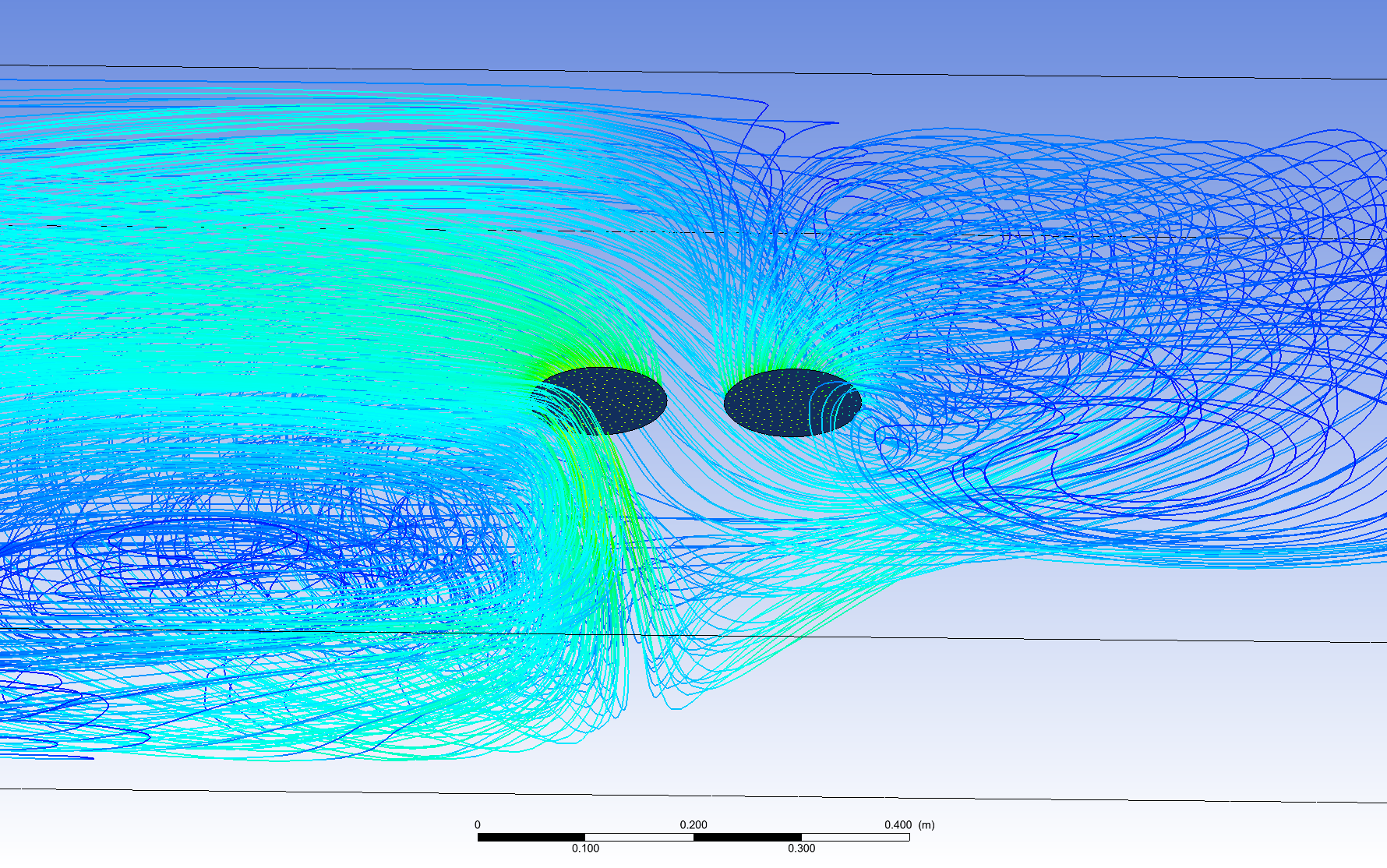}}
\subfigure[\label{fig:rect_5_9_45deg0_cfd} Hover, $45^\circ$ slope tunnel with 0.9 m $\times$ 0.5 m rectangular cross-section.]
{\includegraphics[width=0.64\columnwidth]{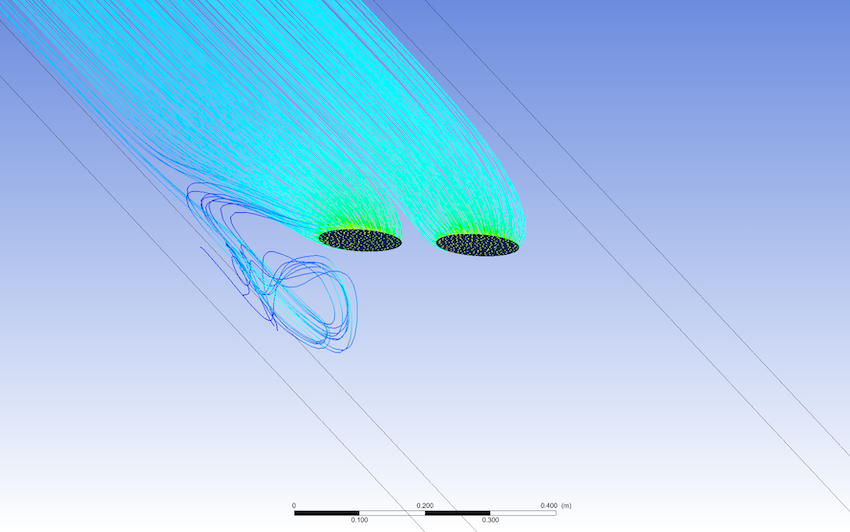}}
\subfigure[\label{fig:rect_5_9_45deg-20_cfd} 2 m/s towards bottom right, $45^\circ$ slope tunnel with 0.9 m $\times$ 0.5 m rectangular cross-section.]
{\includegraphics[width=0.64\columnwidth]{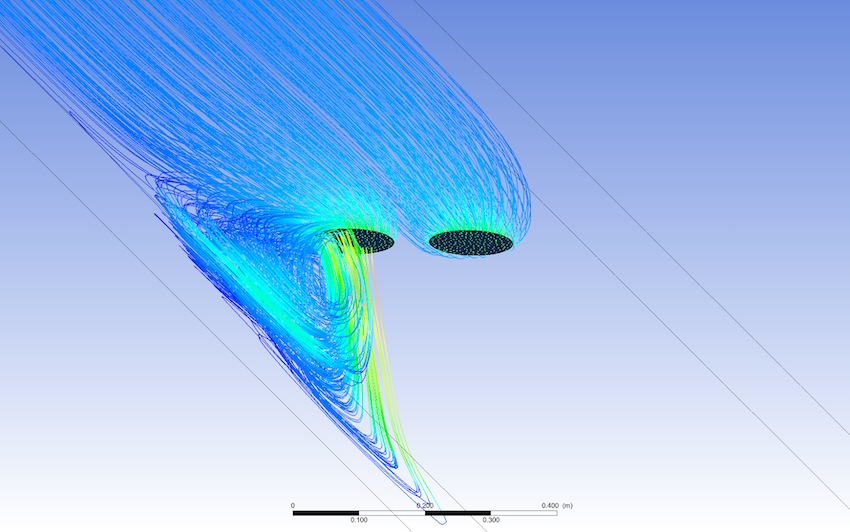}}
\subfigure[\label{fig:rect_5_9_45deg-35_cfd} 3.5 m/s towards bottom right, $45^\circ$ slope tunnel with 0.9 m $\times$ 0.5 m rectangular cross-section.]
{\includegraphics[width=0.64\columnwidth]{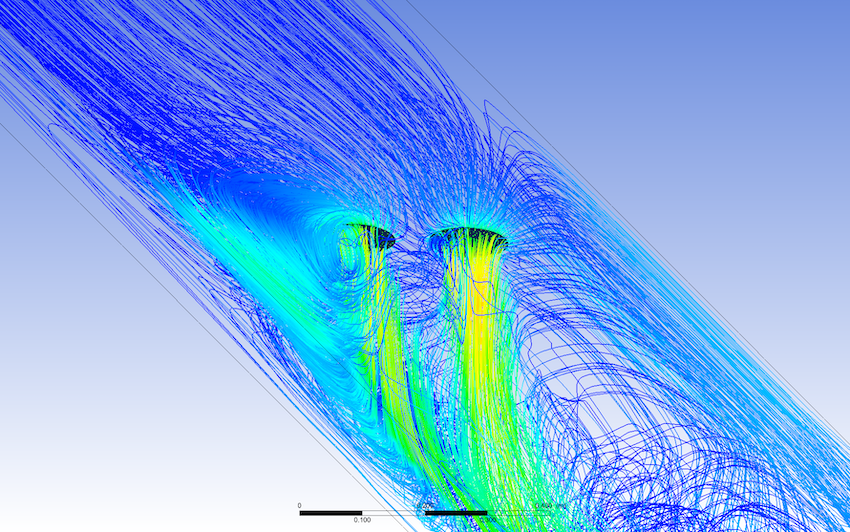}}
\subfigure[\label{fig:circle_5_15deg-5_cfd} 0.5 m/s towards bottom right, $15^\circ$ slope tunnel with 0.5 m diameter circular cross-section.]
{\includegraphics[width=0.64\columnwidth]{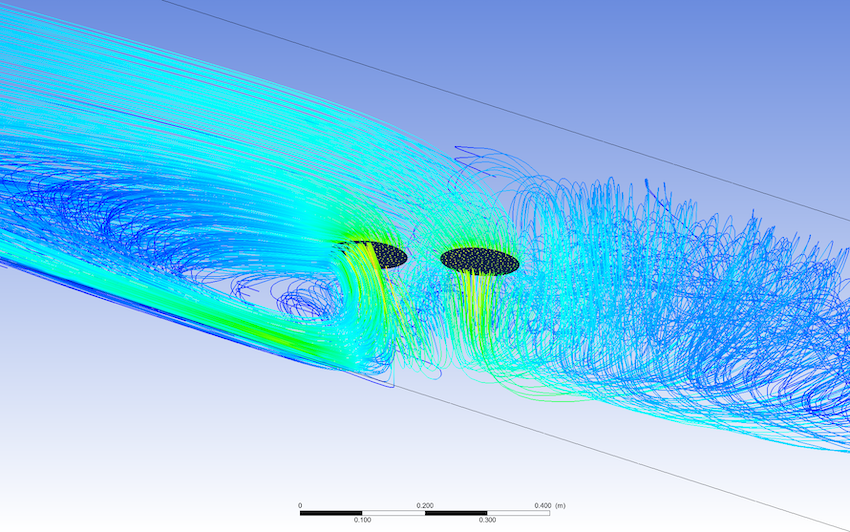}}
\subfigure[\label{fig:circle_5_45deg-5_cfd} 0.5 m/s towards bottom right, $45^\circ$ slope tunnel with 0.5 m diameter circular cross-section.]
{\includegraphics[width=0.64\columnwidth]{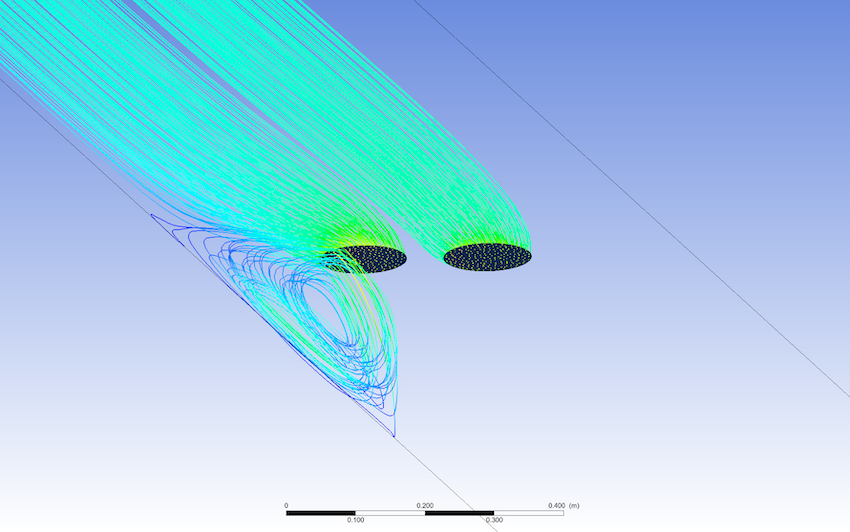}}
\subfigure[\label{fig:circle_5_90deg-5_cfd} 0.5 m/s towards bottom right, vertical tunnel with 0.5 m diameter circular cross-section.]
{\includegraphics[width=0.64\columnwidth]{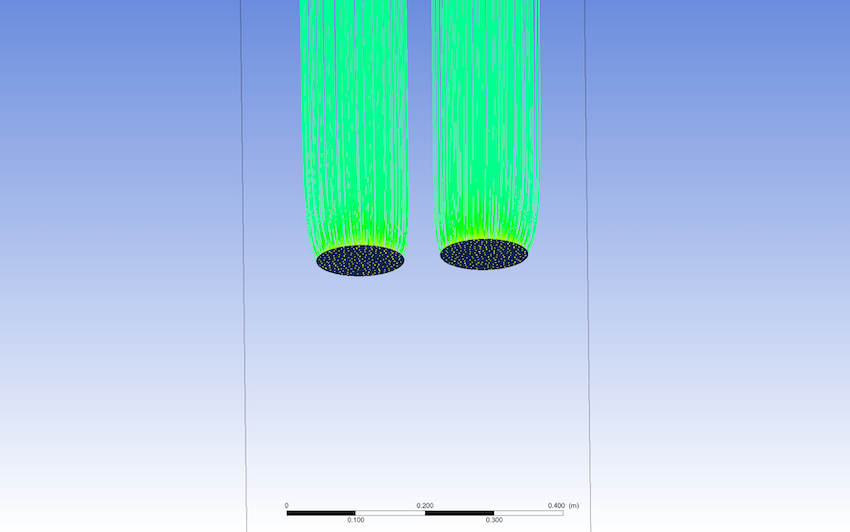}}
\end{center}
\vspace{-0.4cm}
\caption{\label{fig:cfd_result}The backward streamlines from the shadow of the propellers when the quadrotor is flying at different speeds in tunnels of different shapes from the computational fluid dynamics (CFD) result. The black circles represent the propellers and the color indicates the speed of the flow on the streamlines. Note that the illustrations of half of the space can fully represent the results on account of the symmetry.}
\vspace{-0.2cm}
\end{figure*}

\begin{figure*}[h]
\begin{center}         
\subfigure[\label{fig:square_6_disturbance} 0.6 m edge length square cross-section.]
{\includegraphics[width=0.64\columnwidth]{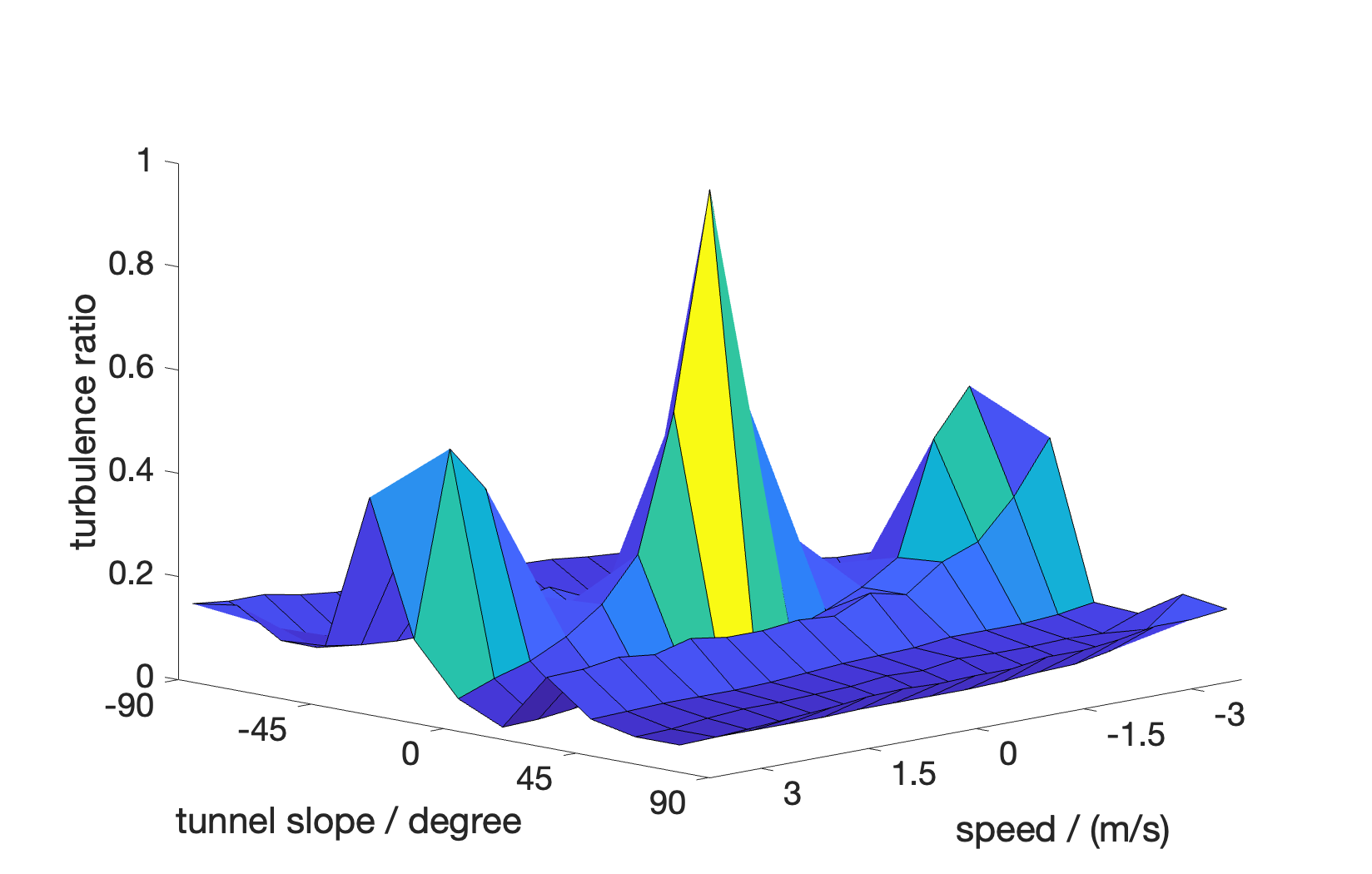}}
\subfigure[\label{fig:rect_6_9_disturbance} 0.5 m $\times$ 0.8 m rectangular cross-section.]
{\includegraphics[width=0.64\columnwidth]{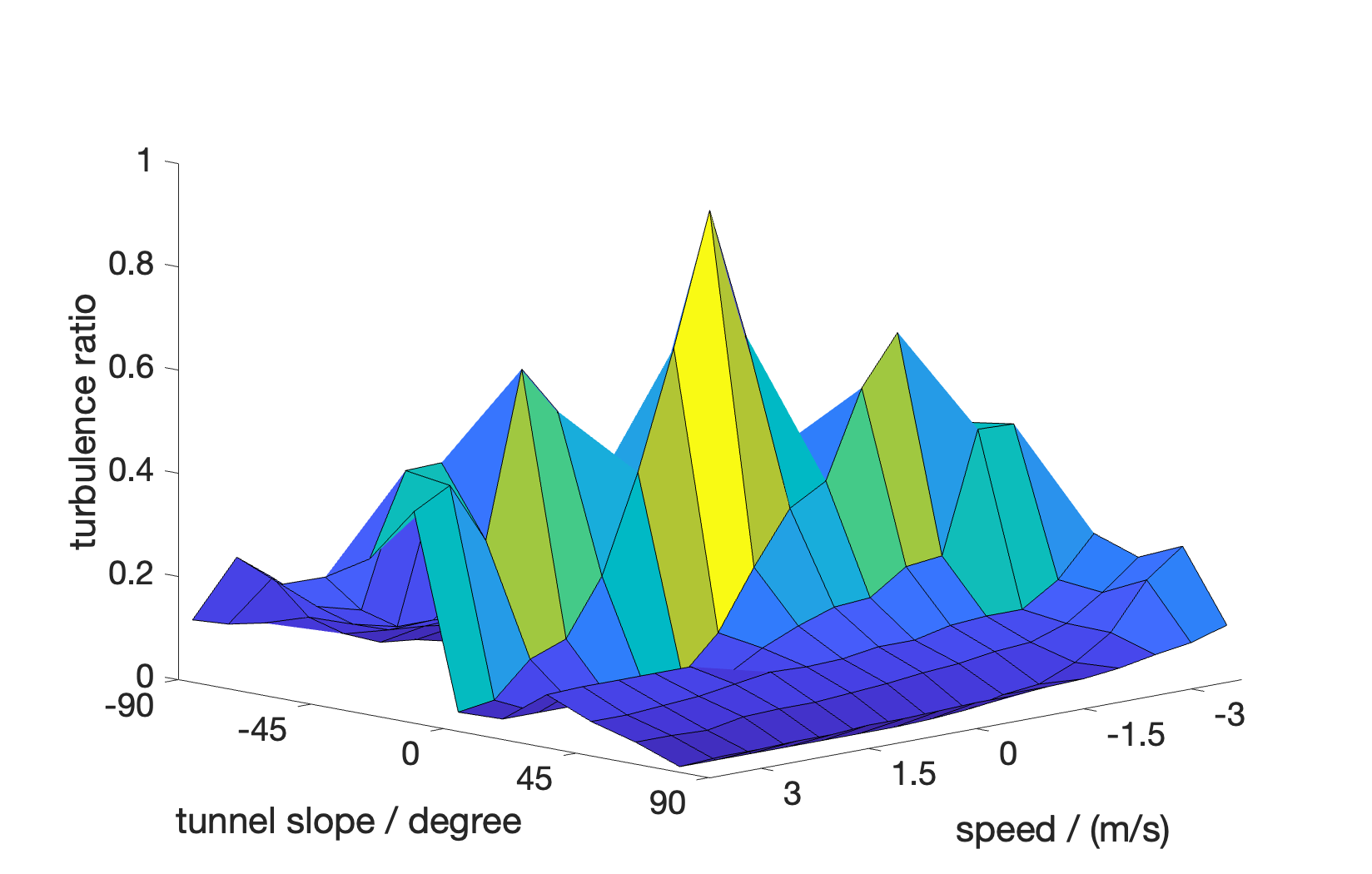}}
\subfigure[\label{fig:circle_5_disturbance} 0.5 m diameter circular cross-section.]
{\includegraphics[width=0.64\columnwidth]{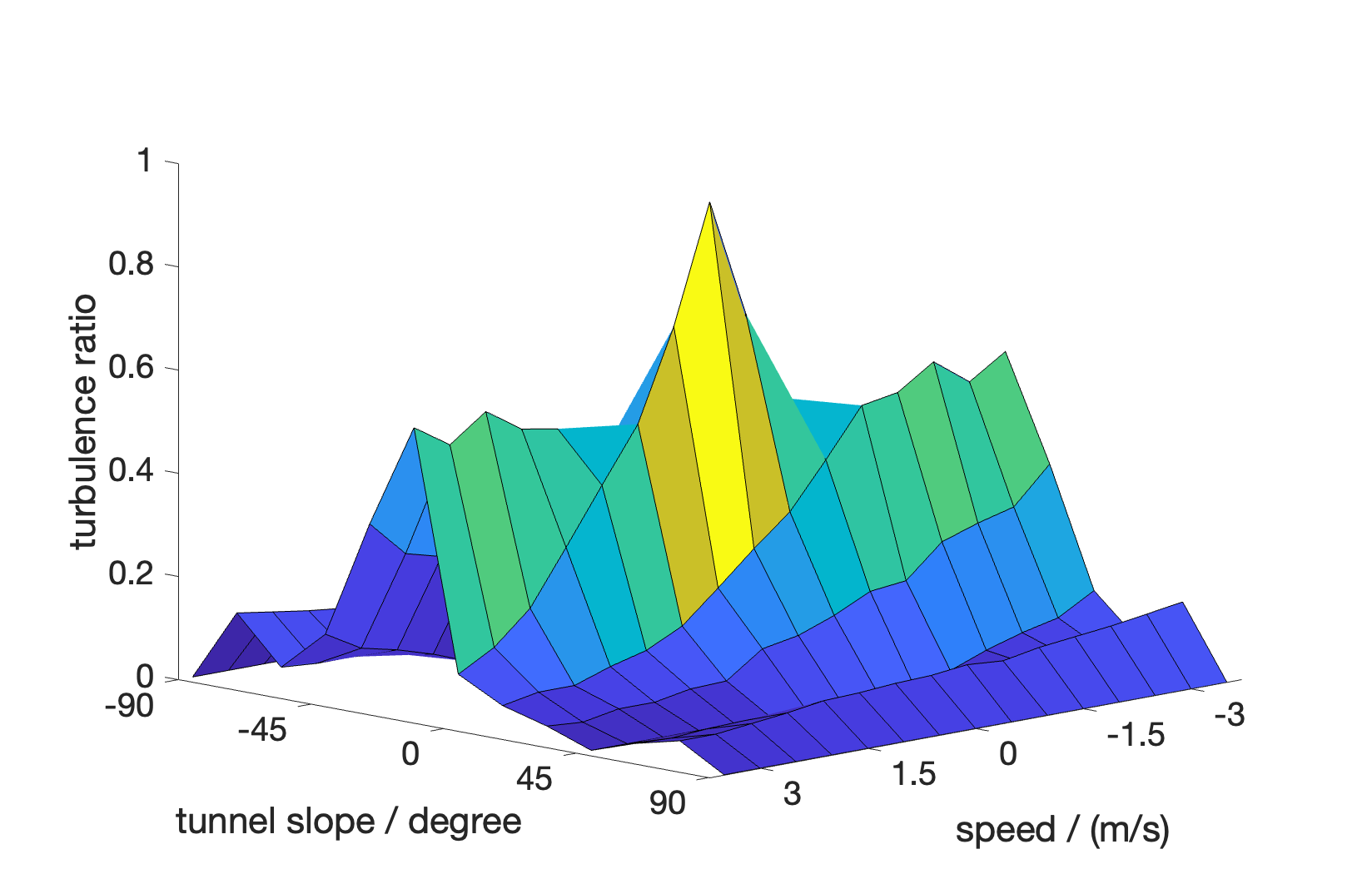}}
\end{center}
\vspace{-0.4cm}
\caption{\label{fig:disturbance_result}The disturbance level when the quadrotor is flying at different speeds in tunnels with different slopes and cross-sections.}
\vspace{-0.2cm}
\end{figure*}

\begin{table}[t]
\centering
\caption{\label{tab:disturbance_mlp}Information of the MLP neural network disturbance model}
\begin{tabular}{@{}lccc@{}}
\toprule
Model & Circle & Rectangle \\
\midrule
Structure  & \multicolumn{2}{c}{3 layers with 64, 64, 16 hidden sizes}\\
Loss function & \multicolumn{2}{c}{MSE}\\
Learning rate & \multicolumn{2}{c}{$1e-3$ to $1e-6$}\\
Optimizer & \multicolumn{2}{c}{Adam}\\
Total epoch & 50000 & 70000 \\
Final training loss & $1.1e-5$ & $1.8e-5$ \\
Final test loss  & $1.3e-5$ & $1.9e-5$\\
\bottomrule
\end{tabular}
\vspace{-0.6cm}
\end{table}

The other crucial factor to analyze is the ego airflow disturbance, as mitigating its effects can enhance control performance. While it might seem that higher flying speeds reduce the impact of airflow disturbances, this intuition mainly applies to horizontal flights. A comprehensive analysis is still necessary to fully understand the effect. Therefore, detailed computational fluid dynamics (CFD) analyses at various angles, flight speeds, and tunnel shapes are conducted to gain a thorough understanding of airflow disturbances during tunnel flights. Due to the impracticality of performing CFD analyses for every possible scenario, a range of settings representing typical real flight conditions is adopted. The cases encompass speeds from 0 m/s to 3.5 m/s with 0.5 m/s intervals, tunnel pitch direction from $-90 ^\circ$ (upward) to $90 ^\circ$ (downward) with $15 ^\circ$ intervals, circular cross-sections with diameters from 0.5 m to 0.9 m with 0.1 m intervals, and rectangular cross-sections with edge lengths from 0.5 m to 0.9 m with 0.1 m intervals, culminating in a total of 2940 cases. In each case, a near-hovering model of the proposed 1.085 kg quadrotor platform with four 5-inch propellers is simplified to four 5-inch thin fans generating a pressure jump of 210 Pa each. The quadrotor is placed in the middle of a 7-meter long tunnel computation region. Since the quadrotor follows the tunnel centerline, symmetry considerations allow the computational region to be reduced by half, expediting the analyses. In each case, a tetrahedral mesh sized from 0.5 to 4 cm is used, depending on the distance from the quadrotor. A moving frame along the quadrotor is set to simulate the flights at various speeds. The airflow together with the tunnel wall moves backward against the static quadrotor, while both ends of the tunnel maintain atmospheric pressures. The cases are solved sing a pressure-based, standard k-epsilon turbulence model and standard wall model using the SIMPLE algorithm in Ansys Fluent.

The backward streamlines from the propellers in selected cases are illustrated in Fig. \ref{fig:cfd_result}. The airflow varies significantly in different situations, underscoring the importance of comprehensive CFD analyses. Since the primary airflow disturbances are attributable to the turbulent flow generated by the propellers, the disturbance level is formulated as the proportion of the turbulent flow intake relative to the total intake of the propellers. According to the streamline results, this formulation can be approximated by dividing the number of backward streamlines from the propellers by the total number of sampled streamlines, which is 200 for each case in practice. The final disturbance level results for selected cases are depicted in Fig. \ref{fig:disturbance_result}. Generally, the disturbance level decreases when flying faster in a horizontal tunnel. However, this trend does not necessarily hold when the tunnel has a slope. At specific flight speeds and tunnel slopes, for instance as shown in Fig. \ref{fig:rect_5_9_45deg0_cfd}-\ref{fig:rect_5_9_45deg-35_cfd}, the quadrotor can easily fly into the turbulent flow it generates, leading to a high disturbance level. This finding contradicts to the simple intuition that faster flights reduce disturbances. To ensure safe flights, these situations should be avoided, necessitating the reduction of disturbance levels as much as possible.

The raw disturbance level data, while derived, cannot be directly used for planning. This is because the flight conditions typically do not match specific CFD cases, and conducting infinite CFD analyses for every scenario is impractical. To address this, a general regression neural network (GRNN) \cite{specht1991general} is employed to regress the raw disturbance level data, thereby generating a continuous model. During the training procedure, the raw disturbance level data are firstly divided into training and test sets with a 4:1 ratio. Then, we apply the GRNN on the test data based on training data and find the minimum mean squared error and the corresponding optimal value of the network parameters. Despite the applicability of this continuous model in subsequent planning, a significant drawback of the GRNN is that its computational cost increases with the size of the raw data sample, leading to slower computation as more CFD results are provided. To address this issue, a learning-based regression model acceleration technique is employed, utilizing a multilayer perceptron (MLP) neural network to represent the model. We uniformly sample 1000000 points within the domain and compute the disturbance level using the GRNN model. The data are then divided into training, validation, and test sets in a 3:1:1 ratio. The final MLP model achieves a speed improvement of over ten times with a negligible loss of 1e-5 level compared to the GRNN when the raw disturbance level data is in the thousands' level. Information about the MLP models is detailed in Tab.~\ref{tab:disturbance_mlp}.

\section{Perception-and-disturbance-aware Planning}
\label{sec:tunnel_planning}

With the formulated perception and disturbance models, perception-and-disturbance-aware planning can be executed, leveraging the mapping results and extracted tunnel center waypoints from the perception module. This planning process includes tunnel-following trajectory and corridor generation, perception-aware active yaw planning, and perception-and-disturbance-aware speed profile planning. The planned trajectories and speed profiles are used to generate position, velocity, and acceleration commands, which are then sent to the controller.

\subsection{Tunnel-following Trajectory and Corridor Generation}
\label{subsec:tunnel_traj}

To ensure safe flights through narrow tunnels, it's essential to have a collision-free trajectory that follows the tunnel centerline while accounting for system dynamics. Additionally, as discussed in Sec. \ref{sec:introduction}, the quality of feature tracking and the level of disturbances, which affect perception and control, vary with tunnel shapes, sizes, and flight speeds. Therefore, identifying the tunnel corridor, including the shape and size of cross-sections, is crucial for subsequent active yaw planning and speed profile planning.

As a result, we propose a tunnel-following trajectory and corridor generation approach that comprises tunnel-following trajectory optimization, cross-section recognition, and trajectory refinement. This integrated approach ensures that the generated trajectories and corridors accurately capture the necessary geometric data of the tunnel while considering smooth system dynamics.

%Since the space inside a narrow tunnel is constrained, the proximity effects are significant. As mentioned in \cite{eberhart2017modeling,conyers2019empirical,wang2021estimation}, the proximity effects, i.e., the ground effect, the near-wall effect, and the ceiling effect, can generate not only additional mean forces but also disturbances. In narrow tunnels, turbulent downwashes bounce back and forth, inducing complicated and unpredictable effects rather than a simple superposition of the proximity effects, and thus bring tremendous difficulties to flights. With consideration of these effects, it is crucial to reserve as large a clearance as possible from the tunnel walls; thus, following the tunnel center line is a desirable solution. Additionally, on account of the 0.25 m minimum working range of the LiDAR, keeping sufficient clearance from the walls is also beneficial to the perception module, further justifying the solution.

\subsubsection{Tunnel-following Trajectory Optimization}
\label{subsubsec:tunnel_traj_opt}

%\subsubsection{Tunnel Centerline Optimization}
%\label{subsubsec:centerline_optimization}
With the acquired tunnel center waypoints mentioned in Sec.~\ref{subsec:center_point_extraction} , a uniform B-spline through them can be generated to represent the initial estimation of the tunnel centerline. However, due to perception noise, this initial centerline is typically jerky and may deviate from the true centerline, making it unsuitable for smooth flight. Additionally, the cross-sections on the jerky initial centerline can significantly differ from the true shape due to the inaccurate directions. Therefore, we introduce a trajectory optimization method that leverages initial waypoints and EDF information from the fused map, while considering system dynamics, to determine the tunnel-following trajectory. This trajectory, represented by a B-spline curve, also facilitates subsequent smooth cross-section recognition.

To enhance the quality of the B-spline trajectory, we propose a cost function $f_{cl}$ for its evaluation and optimization. This function is a weighted sum of the smoothness cost $f_s$, the waypoint constraint cost $f_w$, the interval constraint cost $f_i$, the distance constraint cost $f_d$, and the end state constraint costs $f_e$:
\begin{equation}\label{eq:3d_cost}
	f_{cl} = \lambda_{s} f_{s} + \lambda_{w} f_{w} + \lambda_{i} f_{i} + \lambda_{d} f_{d} +  \lambda_{e} f_{e}.
\end{equation}

Taking account of the smooth dynamics of the multirotor system, a third-order elastic band cost \cite{quinlan1993elastic,zhu2015convex} along the trajectory is adopted:
\begin{equation}\label{eq:elastic} 
	f_{s} 
	= \sum\limits_{i=p_b}^{N-p_b} \Vert -\mathbf{Q}_{i} + 3\mathbf{Q}_{i+1} - 3\mathbf{Q}_{i+2} + \mathbf{Q}_{i+3} \Vert^{2},
\end{equation}
where $p_b \geq 3$ is the order of the B-spline, $\mathbf{Q}_{i}$ represents the position of the $i$-th control point, and $N$ is the total number of control points.

Moreover, the constraints on the waypoints are essential for preserving the original shape of the centerline by employing the waypoint constraint cost $f_w$, which is defined as the summation of the squared deviations of the waypoints on the current trajectory from the original waypoints $W_i$:
\begin{equation}\label{eq:waypoint} 
	f_{w} 
	= \sum\limits_{i=p_b}^{N-p_b} \Vert EvalBspline(\mathbf{Q}_{i},...,\mathbf{Q}_{i+p_b-1}) - W_{i} \Vert^{2},
\end{equation}
where the $EvalBspline$ function evaluates the waypoints on the B-spline according to the corresponding control points.

Furthermore, an interval constraint cost $f_i$ is added to prevent fluctuating distances between control points by penalizing deviations in control point speed from a virtual speed $v_{virtual}$:
\begin{equation}\label{eq:interval}
	f_{i} = \sum\limits_{i=p_b}^{N-1} (\Vert \frac{\mathbf{Q}_{i+1} - \mathbf{Q}_{i}}{\delta t} \Vert  - v_{virtual})^2,
\end{equation}
where $\delta t$ is the virtual time interval between the control points.

Additionally, due to occasional partial observation of the tunnel wall, a distance constraint cost $f_d$ is implemented to enhance safety by quadratically penalizing control points that are closer to the tunnel wall than a predefined distance $d_0$:
\begin{equation}\label{eq:dist}
	f_{d} = \sum\limits_{i=p_b}^{N-p_b} F_{d}(edf(\mathbf{Q}_{i})),
\end{equation}
where $ edf(\mathbf{Q}_{i}) $ represents the EDF value at $\mathbf{Q}_{i}$ and $ F_{d} $ is a differentiable quadratic cost function defined by:
\begin{equation}\label{eq:dist_cost}
	F_{d}(edf(\mathbf{Q}_{i})) = \left\{
	\begin{array}{cl}
	(edf(\mathbf{Q}_{i})-d_{0})^{2} & edf(\mathbf{Q}_{i}) \le d_{0} \\
	0 & edf(\mathbf{Q}_{i}) > d_{0}.
	\end{array}
	 \right.
\end{equation}
Note that $d_0$ is typically set slight larger than the radius of the quadrotor to ensure a safety margin.

Finally, the end state constraint cost $f_e$ is included to preserve the original terminal states:

\begin{equation}\label{eq:end}
	f_{e} = \sum\limits_{i=0}^{k} \Vert EvalBspline_i(\mathbf{Q}_{N-p_b},...,\mathbf{Q}_{N-1}) - E_i \Vert^{2},
\end{equation}
where the $EvalBspline_i$ function computes the $i$-th derivative of the point on the B-spline based on the corresponding control points, and $E_i$ denotes the $i$-th derivative of the original ending state.

A limited-memory BFGS optimization is applied to the control points $\mathbf{Q}_i$'s based on the total cost $f_{cl}$. This process establishes a smooth tunnel-following trajectory, represented by the optimal control points, for subsequent planning procedures.

\subsubsection{Tunnel Cross-section Recognition}
\label{subsubsec:cross-section_recognition}

\begin{figure}[t]
\begin{center}
\subfigure[\label{fig:cs_rviz} The visualization during flight in the tunnel shown in Fig. \ref{fig:tunnel_2d} when cross-sections of different shapes appear in the planning horizon. The color coding indicates the height of the real-time constructed map, and the black arrows show the estimated tunnel entrance. The axes indicate the current pose of the quadrotor. The blue and red curves indicate the initial tunnel centerline from the extracted waypoints and the tunnel-following trajectory respectively.]
{\includegraphics[width=0.9\columnwidth]{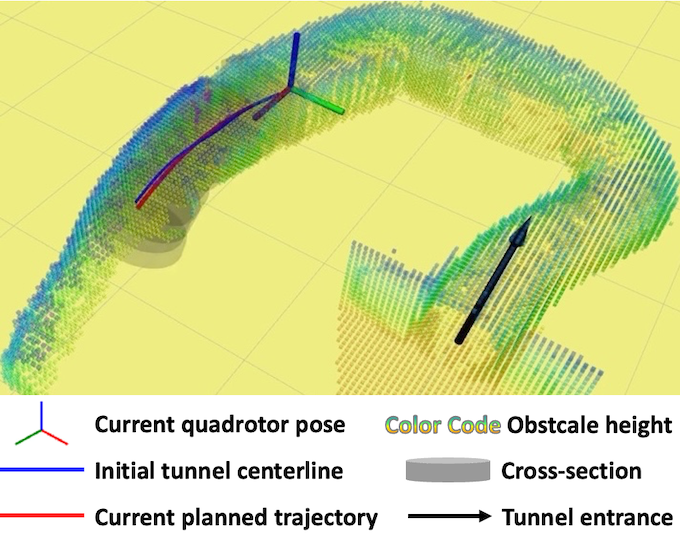}} 
\subfigure[\label{fig:cs_ori} The raw cross-section images.]
{\includegraphics[width=0.23\columnwidth]{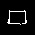} \includegraphics[width=0.23\columnwidth]{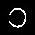}} 
\subfigure[\label{fig:cs_detect} The detected rectangle and circle.]
{\includegraphics[width=0.23\columnwidth]{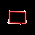} \includegraphics[width=0.23\columnwidth]{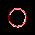}}
\end{center}
\vspace{-0.4cm}
\caption{\label{fig:cs_detect_viz} The visualization and two of the recognized cross-sections when the quadrotor fly about the position where the tunnel shape and size changes in the 2-D tunnel case 3 shown in Fig. \ref{fig:tunnel_2d}. The detected rectangular and circular cross-sections are marked in red and shown half-transparently in the visualization.}
\vspace{-0.6cm}
\end{figure}

Based on the obtained smooth tunnel-following trajectory, the cross-section planes can be determined for recognition, enabling the generation of the corridor. Initially, raw cross-section images are generated along the tunnel-following trajectory according to the local EDF map, as illustrated in Fig. \ref{fig:cs_ori}. Subsequently, these cross-section images are extracted for shape recognition. Two example results are depicted in Fig. \ref{fig:cs_detect}. The recognition pipeline is divided into two stages: shape classification and pattern detection. During this process, we assume that all the shapes of the cross-sections belong to a close set and we recognize the shape patterns within this set. In this paper, we focus on rectangles and circles for the shape set, as these cover a wide range of real-world tunnels. Furthermore, the pipeline can be easily extended to arbitrary shapes with minor modifications.

%\subsubsection{Cross-section Shape Classification}
%\label{subsubsec:cross_section_classification}

\begin{figure}[t]
\begin{center}
\subfigure[\label{fig:cs_train} The generated image examples from the training and validation sets.]
{\includegraphics[width=0.23\columnwidth]{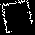} \includegraphics[width=0.23\columnwidth]{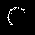}} 
\subfigure[\label{fig:cs_test} The extracted cross-sections from real flights for testing.]
{\includegraphics[width=0.23\columnwidth]{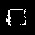} \includegraphics[width=0.23\columnwidth]{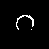}}
\end{center}
\vspace{-0.4cm}
\caption{\label{fig:cs_classification} The example images for training, validation, and testing of the cross-section shape classification CNN.}
%\vspace{-0.9cm}
\end{figure}

\paragraph{Cross-section Shape Classification}
In the shape classification stage, we adopt a convolutional neural network (CNN) with two convolutional layers, two pooling layers, and two fully connected layers. The CNN takes the cross-section images as input and outputs the shape class. The input images are generated with resolutions of $35 \times 35$ in practice. The training set and validation set are generated from perfect circles and rectangles with random noise and occlusions on the edges. The training set consists of 24000 images of rectangles and circles in different sizes and orientations at different positions, while the validation set comprises 6000 images of generated rectangles and circles. The test set is formed from 2000 cross-section images collected from real flights. After training, the accuracy of the classification reaches 99.61\% on the validation set and 99.55\% on the test set. For extendibility, the CNN can be easily modified to support a larger set of shape classes with more generated images of different shapes.

%\subsubsection{Cross-section Pattern Detection}
%\label{subsubsec:cross_section_detection}
\paragraph{Cross-section Pattern Detection}
Once the cross-section shape class is determined by the classification CNN, the detection of the shape pattern can be executed. For rectangles, we adopt a Hough line based detection method modified from \cite{jung2004rectangle}. After acquiring the parallel line pairs through an enhanced Hough line transform, the rectangle patterns are identified based on right angle and edge length criteria. The final rectangle selection criterion is modified to be a weighted sum of edge intensity, parallelism, and perpendicularity of the edges. For circles, a circle Hough transform is employed for detection. If no pattern of the classified shape is detected, this typically indicates an incomplete map, and the cross-section shape and size are assumed to match the previous one. If a cross-section is of an unknown shape that is neither rectangle nor circle, it is treated as a circle centered at the maximum EDF value point to maintain sufficient distance from the tunnel wall. The detection process is also easily extendable to more complex shapes using the generalized Hough transform.

\subsubsection{Tunnel-following Trajectory Refinement}
\label{subsubsec:traj_refinement}
Following the cross-section recognition, the recognized cross-section centers may deviate from the previous extracted tunnel center waypoints. In such instances, the waypoints are adjusted to match the recognized cross-section centers. Subsequently, the procedures starting from the tunnel-following trajectory optimization can be repeated to acquire a refined tunnel-following trajectory. This ensures that the generated trajectories accurately capture the necessary geometric data of the tunnel while considering the smooth system dynamics.

\subsection{Perception-aware Active Yaw Planning}
\label{subsec:yaw_planning}
With the extracted tunnel-following trajectory, the flight path is determined. However, active yaw planning is still needed for the virtual omni-directional perception module to sense the environment ahead. This planning involves two stages: yaw waypoint selection and yaw trajectory optimization.

\subsubsection{Yaw Waypoint Selection}
\label{subsubsec:yaw_waypoint_selection}

\begin{figure}[t]
\begin{center}
{\includegraphics[width=0.9\columnwidth]{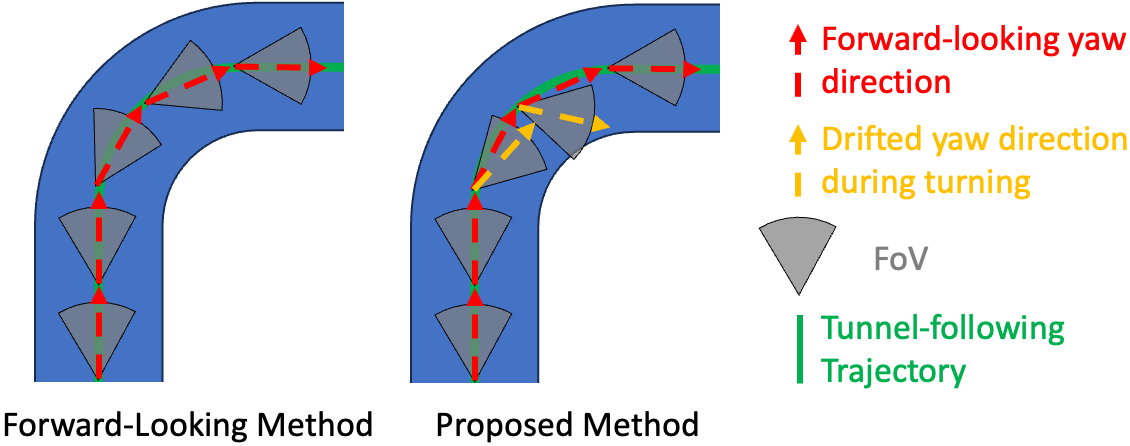}} 
\end{center}
\vspace{-0.4cm}
\caption{\label{fig:active_yaw}The illustration of the traditional forward-looking yaw method and the proposed active yaw method. The proposed active yaw method can perceive more information on the inner side at the turn.}
%\vspace{-1.0cm}
\end{figure}

In most general planning frameworks, the yaw direction is set to point towards a position at a constant distance or time ahead on the planned trajectory. This strategy is generally effective in open areas. However, in extremely restricted environments such as narrow tunnels, this approach is inadequate, particularly when the quadrotor needs to perform sharp turns. As illustrated in Fig. \ref{fig:active_yaw}, the limited FoV often cannot cover the inner side of a turning trajectory. To address this, we employ a side-drifting method for the quadrotor to navigate corners, enhancing visibility of the inner side.

For each sampled waypoint on a given tunnel-following trajectory, a forward-looking 2-D direction $\mathbf{y_0}_i$ can be easily derived, pointing towards a position a constant distance ahead. The drifted yaw direction $\mathbf{y_d}_i$ is then calculated based on the radius $\mathbf{r_h}_i$ of the curvature at the sampled waypoint on the 2-D tunnel-following trajectory projected on the horizontal plane and the forward-looking direction $\mathbf{y_0}_i$ as follows:

\begin{equation}\label{eq:yaw_dir} 
	\mathbf{y_d}_i
	= \mathbf{y_0}_i + k_r \frac{\mathbf{r_h}_i}{||\mathbf{r_h}_i||^2},
\end{equation}
where $k_r$ is a constant used to adjust the drift ratio. Utilizing this formulation, the larger the curvature, or the smaller the radius at a point on the trajectory, the greater the yaw drift at a turn. Additionally, when the quadrotor reaches a vertical section of the tunnel, the yaw remains unchanged until the 3-D forward-looking direction deviates sufficiently from the vertical direction. With the 2-D yaw directions $\mathbf{y_d}_i$, the 1-D yaw waypoints ${wy}_{i}$ can be simply obtained.

\subsubsection{Yaw Trajectory Optimization}
\label{subsubsec:yaw_trajectory_optimization}
From the 1-D yaw waypoints ${wy}_{i}$, a B-spline can be parametrized for further optimization. Similar to the approach in Sec. \ref{subsubsec:tunnel_traj_opt}, the total objective is formulated as a weighted sum of the yaw smoothness cost $f_{sy}$, the yaw waypoint cost $f_{wy}$, and the yaw end state cost $f_{ey}$:

\begin{equation}\label{equ:yaw_cost}
	f_{y} = \lambda_{sy} f_{sy} + \lambda_{wy} f_{wy} + \lambda_{ey} f_{ey}.
\end{equation}

The smoothness yaw cost is defined as a third order elastic band cost:
\begin{equation}\label{eq:yaw_elastic} 
	f_{sy} 
	= \sum\limits_{i=p_b}^{N-p_b} (-qy_{i} + 3{qy}_{i+1} - 3{qy}_{i+2} + 3{qy}_{i+3})^{2},
\end{equation}
where $qy_{i}$ represents the $i$-th control point of the parametrized B-spline.

The yaw waypoint cost is similar defined as:
\begin{equation}\label{eq:yaw_waypoint}
	\begin{aligned}
	f_{wy} 
	= &\sum\limits_{i=p_b}^{N-p_b} k_{wyi} \cdot (EvalBspline({qy}_{i},...,{qy}_{i+p_b-1}) - {wy}_{i})^{2},\\
	&k_{wyi} = 
	\left\{
	\begin{array}{cl}
	0 & {wy}_{i}\ in\ vertical\ sections, \\
	1 & otherwise,
	\end{array}
	 \right.
	\end{aligned}
\end{equation}
where the $EvalBspline$ function evaluates the yaw waypoint on the B-spline according to the corresponding control points, except that the deviations from the yaw waypoints in the vertical sections are not penalized.

Finally, the yaw end state cost is defined as:
\begin{equation}\label{eq:yaw_end}
	f_{ey} = \sum\limits_{i=0}^{k} (EvalBspline_i({qy}_{N-p_b},...,{qy}_{N-1}) - ey_i)^{2},
\end{equation}
where the $EvalBspline_i$ function evaluates the $i$-th derivative of the point on the B-spline according to the corresponding control points, and $ey_i$ indicate the $i$-th derivative of the original yaw ending state, respectively.

\subsection{Perception-and-disturbance-aware Speed Profile Planning}
\label{subsec:speed_planning}
Using the obtained tunnel-following trajectory and the yaw trajectory, the perception-and-disturbance-aware speed profile planning can be executed by leveraging the models formulated in Sec. \ref{sec:disturbance_perception_cost} to address both factors. Given the fixed trajectory shape, the planning is conducted within the 1-D distance space of the tunnel-following trajectory to generate speed profiles. This process involves two key phases: initial speed profile search and speed profile optimization.

\subsubsection{Initial Speed Profile Search}
\label{subsubsec:speed_search}

For the initial speed profile search, a hybrid-state A* search \cite{dolgov2010path} along the tunnel-following trajectory is adopted. Multiple discretized acceleration inputs are utilized to generate motion primitives at different time steps according to the state transition equation:
\begin{equation}\label{eq:state_transition}
	\mathbf{x}(t) = \exp({\mathbf{A}t})\mathbf{x}(0) + 
	\int_{0}^{t} \exp{\mathbf{A}(t-\tau)}\mathbf{B}\mathbf{u}(\tau) \ d\tau.
	\vspace{-0.1cm}
\end{equation}

The total actual cost $\mathcal{G}$ is defined as a weighted combination of the input, perception, ego airflow disturbance, and time cost:
\begin{equation}\label{equ:cost_search}
	\mathcal{G}(T) = \int_{0}^{T} \Vert \mathbf{u}(t) \Vert^{2} + \lambda_{of} \hat{v}_{of}(t)^{2} + \lambda_{ed}\Vert EDL(t) \Vert^{2} dt + \rho T,
\end{equation}
where $\lambda_{of}$, $\lambda_{ed}$ and $\rho$ are the weight constants, $\hat{v}_{of}(t)$ is the mean optical flow speed, and $EDL(t)$ is the estimated ego airflow disturbance level at a position on the tunnel-following trajectory at time $t$. Note that the integral is approximated using the summation of the interpolated $\hat{v}_{of}(t)$ and $EDL(t)$ with small time steps during the search process. Additionally, since the yaw trajectory is already set, considering the near-hover movement of the quadrotor, the angular velocity of the camera $\omega_{cam}$ in Eq. \ref{eq:projection} only changes with flight speed at a given position on the tunnel-following trajectory. Consequently, $\hat{v}_{of}(t)$ only depends on position and flight speed, which is the state during the search process.

When using $k$ cameras, the total estimated optical flow speed $\hat{v}_{of}(t)$ is calculated similarly to a parallel circuit, as the cameras operate in parallel:

\begin{equation}\label{eq:of_parallel}
	\hat{v}_{of}(t) = \frac{1}{\sum\limits_{i=1}^{k} \frac{1}{\hat{v}_{of_i}(t)}},
\end{equation}
where $\hat{v}_{of_i}(t)$ represents the approximated mean optical flow speed of camera $i$, as determined using the perception factor model elaborated in Sec. \ref{subsec:perception_factor}.

The heuristic cost $\mathcal{H}$ is defined as the minimized cost of the input and the time term in $\mathcal{G}$ utilizing Pontryagin’s minimum principle\cite{mueller2015computationally}:
\begin{equation}
\label{equ:poly}
\begin{gathered}
\begin{bmatrix}
	{\alpha} \\ {\beta}
\end{bmatrix}
= \frac{1}{T^{3}}
\begin{bmatrix}
	 {-12} & {6 T} \\ {6 T} & {-2 T^{2}}
\end{bmatrix}
\begin{bmatrix}
p_{g}-p_{s}-v_{s} T \\ v_{g}-v_{s}
\end{bmatrix} 
\\
\mathcal{H}(T) = \min_{T}(\rho T + (\frac{1}{3}\alpha^{2}T^{3} + \alpha \beta T^{2} + \beta^{2} T)),
\end{gathered}
\end{equation}
where $p_{s}$, $p_{g}$ and $ v_{s}$, $v_{g}$ represents the searching and goal positions and velocities, respectively. 
The total cost used for the hybrid-state A* search, denoted as $\mathcal{F}$, is simply $\mathcal{G}+\mathcal{H}$. For safety considerations, the ending velocity $v_{g}$ at the end of the tunnel-following trajectory is always set to $0$. After executing the hybrid A* search, the states of the initial 1-D waypoints can be retrieved.

\subsubsection{Speed Profile Optimization}
\label{subsubsec:speed_optimization}

After the initial speed profile search, an optimization similar to the approach in Sec. \ref{subsubsec:tunnel_traj_opt} is performed to determine the optimum speed profile. The searched 1-D waypoints are first parametrized into a uniform B-spline to facilitate the optimization.

The total cost of the 1-D trajectory $f_{1}$ is defined as the weighted sum of the smoothness jerk and acceleration cost $f_{sj}$ and $f_{sa}$, the perception cost $f_{p}$, the ego airflow disturbance cost $f_{ed}$ and the ending state cost $f_{e1}$ as:
\begin{equation}\label{equ:cost_1d}
	f_{1} = \lambda_{sj} f_{sj} + \lambda_{sa} f_{sa} + \lambda_{p} f_{p} + \lambda_{ed} f_{ed} + \lambda_{e1} f_{e1}.
\end{equation}

The smoothness jerk and acceleration cost $f_{sj}$ and $f_{sa}$ are included to prevent abrupt maneuvers along the tunnel-following trajectory and are formulated similarly to the elastic band cost used in previous sections:
\begin{equation}\label{eq:elastic_1d}
\begin{gathered}
	f_{sj} 
	= \sum\limits_{i=p_b}^{N-p_b} (-q_{i} + 3{q}_{i+1} - 3{q}_{i+2} + 3{q}_{i+3})^{2},
	\\
	f_{sa} 
	= \sum\limits_{i=p_b}^{N-p_b} (q_{i} - 2{q}_{i+1} + {q}_{i+2})^{2},
\end{gathered}
\end{equation}
where $q_{i}$ represents the $i$-th control point of the parametrized B-spline.

The perception cost $f_{p}$ and the ego airflow disturbance cost $f_{ed}$ are defined on the waypoints:

\begin{equation}\label{eq:of_1d}
	f_{p} = \sum\limits_{i=p_b}^{N-p_b} \hat{v}_{of}(EvalBspline({q}_{i},...,{q}_{i+p_b-1}))^2,
\end{equation}

\begin{equation}\label{eq:edl_1d}
	f_{ed} = \sum\limits_{i=p_b}^{N-p_b} EDL(EvalBspline({q}_{i},...,{q}_{i+p_b-1}))^2,
\end{equation}
where $\hat{v}_{of}$ and $EDL$ represent the disturbance level and the calculated optical flow speed at the waypoint, as described in Sec. \ref{sec:disturbance_perception_cost} and \ref{subsubsec:speed_search}, and the $EvalBspline$ function evaluates the waypoint and speed on the B-spline according to the corresponding control points.

The ending state cost $f_{e1}$ is also added to constrain the ending states similar to the previous sections:
\begin{equation}\label{eq:end_1d}
	f_{e1} = \sum\limits_{i=0}^{k} (EvalBspline_i({q}_{N-p_b},...,{q}_{N-1}) - e1_i)^{2},
\end{equation}
where the $EvalBspline_i$ function evaluates the $i$-th derivative of the point on the B-spline according to the corresponding control points, and $e1_i$'s indicate the $i$-th derivative of the original ending state, which are the original tunnel-following trajectory length and the 0 ending speed in practice.

%Although the CFD analyses can verify the intuition that flying at higher speeds mitigates the effect of ego airflow disturbances, the errors from modeling, discretization, etc., as well as complex scenarios in real-world environments, make it hard to generate precise quantitative results for direct usage. Therefore, an experiment-based speed selection workflow is still necessary. Experiments for speed selection are conducted in a straight narrow tunnel of around 0.6 m in both width and height, as shown in Fig. \ref{fig:tunnel_0}. The proposed autonomous tunnel flight system traverses the tunnel with desired speeds from 0.1 m/s to 2.5 m/s, with an interval of 0.1 m/s. The flight at each desired speed is repeated 10 times for data collection. The control error data are compared with data of straight-line flights recorded at the same flight speeds in a broader area outside the tunnel. Additionally, the feature tracking data of the VIO system are also collected during the tunnel flights for further analyses. Finally, the practical speed range is selected according to the root-mean-square errors (RMSEs) of the positions and the minimum number of features tracked by the VIO system.

\section{Experiments and Results}
\label{sec:result}

\subsection{Detailed Implementation}
\label{subsec:implementation}
We employ three Intel Realsense L515 Lidar cameras as our primary perception sensors, which capture depth and color images at a frequency of 30 Hz each. Additionally, a Bosch BMI088 IMU, integrated into the NxtPX4 flight controller, provides data at a rate of 500 Hz to assist in control and state estimation. The comprehensive software system, depicted in Fig. \ref{fig:system}, encompasses modules for perception, planning, and high-level control. This software system is implemented in C++ and runs on an Nvidia Orin NX onboard computer, which operates at 2.0 GHz with a maximum power consumption of 25 w. The RGBD-inertial state estimation module outputs 30~Hz camera rate odometries and 500 Hz IMU rate odometries. The mapping module updates at 15 Hz, while the tunnel center waypoint extraction and planning modules both run at 10 Hz. The high-level control module generates thrust and attitude commands at a frequency of 100 Hz, which are then transmitted to the NxtPX4 flight controller to produce the necessary low-level commands executed by the four T-Motor F60 PRO KV2550 motors with 5-inch 5045 propellers.

\subsection{Experiment Sets Configuration}
\label{subsec:exp_setup}

\begin{table}[t]
\centering
\caption{\label{tab:tunnel_case}Experiment Sets Configuration}
\begin{tabular}{@{}lccc@{}}
\toprule
Case & Length & Cross-section shape & Cross-section dimension\\
\midrule
Straight   & 6 m  & Square & 0.6 m  \\
2-D case 1 & 10 m & Circle & 0.6 m - 0.7 m\\
2-D case 2 & 10 m & Circle & 0.6 m - 0.7 m \\
2-D case 3 & 20 m & Circle \& rectangle & 0.5 m - 0.7 m \\
3-D case 1 & 10 m & Circle & 0.6 m - 0.7 m \\
3-D case 2 & 23 m & Circle \& rectangle & 0.5 m - 0.9 m \\
Rigid vent pipe & 20 m & Rectangle & 0.5 m - 0.8 m\\
Construction site & 20 m & Circle & 0.7 m - 0.8 m \\
\bottomrule
\end{tabular}
%\vspace{-0.4cm}
\end{table}

Multiple sets of experiments are conducted in various scenarios (Fig.~\ref{fig:tunnel}), covering both 2-D and 3-D cases, with details provided in Tab.~\ref{tab:tunnel_case}. We use flexible and rigid pipes made from plastic and steel, featuring various cross-sectional shapes and dimensions to represent a wide range of real-world situations.

%, including a 6-m-long straight tunnel with a squared cross-sectional shape of 0.6 m edge length (Fig. \ref{fig:tunnel_straight}), two 10-m-long 2-D tunnels with circular cross-sections of 0.7 m diameter (Fig. \ref{fig:tunnel_n}-\ref{fig:tunnel_ra}), a 20-m-long 2-D tunnel with both circular and rectangular cross-sections ranging from 0.5 m to 0.9 m in diameter (Fig. \ref{fig:tunnel_2d}), a 10-m-long 3-D tunnel with a circular cross-section of 0.7 m diameter (Fig. \ref{fig:tunnel_3d_circle_7}), a 23-m-long 3-D tunnel with both circular and rectangular cross-sections ranging from 0.5 m to 0.7 m in diameter (Fig. \ref{fig:tunnel_3d}), a 20-m-long rigid rectangular vent pipe with diameters ranging from 0.5 m to 0.8 m (Fig. \ref{fig:vent}), and a 20-m-long vent pipe with circular cross-sections of 0.8 m diameter on a construction site from indoor to outdoor areas (Fig. \ref{fig:tunnel_ssl}).

\begin{figure}[t]
\begin{center}
{\includegraphics[width=0.8\columnwidth]{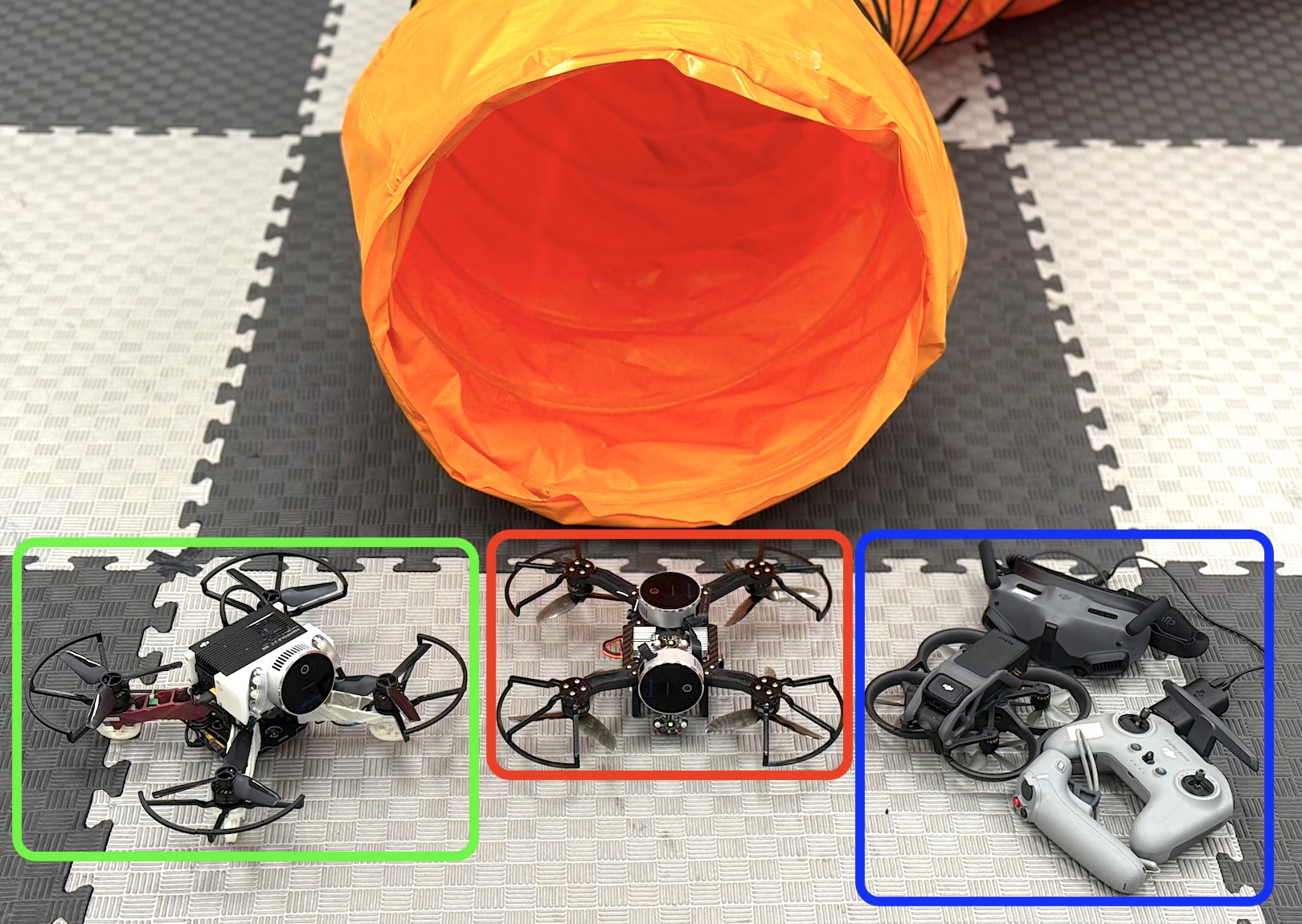}} 
\end{center}
\vspace{-0.3cm}
\caption{\label{fig:old_drone_fpv}The autonomous tunnel flight system (red frame), the tunnel flight system in \cite{wang2022neither} (green frame), and the commercial DJI Avata FPV system for manual flights (blue frame).}
\vspace{-1.7cm}
\end{figure}

Additionally, we conducted a series of ablation studies on the proposed planning method in selected representative cases. We also performed comparisons with the system in \cite{wang2022neither} and manual flights performed by an experienced pilot, in both 2-D and 3-D tunnels. For manual flights, we use the DJI Avata, which is one of the most advanced and compact commercial FPV systems, featuring a 25 cm diameter. The system in \cite{wang2022neither}, the commercial DJI Avata FPV system, and the proposed system are illustrated in Fig. \ref{fig:old_drone_fpv}.

%Experiments are conducted in multiple narrow tunnels of 0.6 m cross-sectional edge length and two vent pipes of 0.7 m diameter, as shown in Fig. \ref{fig:tunnel}. The customized 1.23 kg quadrotor platform with 5-inch propellers, shown in Fig. \ref{fig:drone}, has a diameter of 40 cm, meaning there is only around 10 cm of clearance on each side in the tunnels and vent pipes.
%
%The first set of experiments is conducted in a straight tunnel of 6 m length for speed selection, as stated in Sec. \ref{subsubsec:speed_workflow}.
%The second set of experiments is conducted in three differently curved tunnels, as shown in Fig. \ref{fig:tunnel_3}, \ref{fig:tunnel_1}, and \ref{fig:tunnel_2}. The flight speeds are set to be 0.2 m/s, 0.5 m/s, 1 m/s, 1.5 m/s, and 2 m/s, and six flights are performed for data collection at each speed. Then, the flight data are analyzed to verify the speed selection results, as well as the robustness of the autonomous flight system.
%The third set of experiments is conducted in two vent pipes with a circular cross-section of around 0.7 m in diameter, as shown in Fig. \ref{fig:vent_case_1} and \ref{fig:vent}. These vent pipes, which are commonly used in factories, are adopted to validate the proposed system at the selected speed in more realistic scenarios. Comparison experiments with a SOTA motion planning method\cite{zhou2019robust} and manual flights performed by an experienced pilot using the commercial FPV drone shown in Fig. \ref{fig:drone_fpv} are also performed.

\subsection{Straight Tunnel}
\label{subsec:straight_tunnel_result}

\begin{figure}[t]
\begin{center}
\includegraphics[width=0.8\columnwidth]{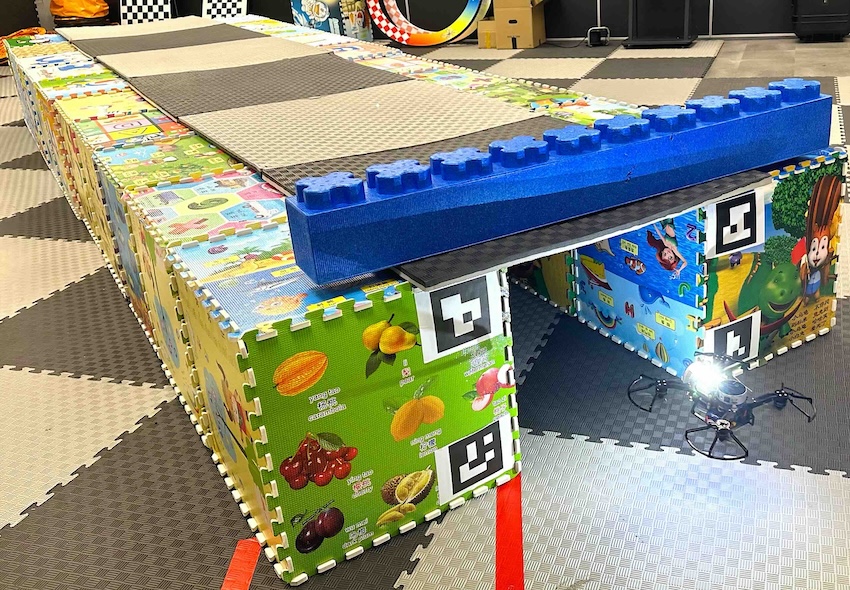}
\vspace{-0.2cm}
\caption{\label{fig:tunnel_straight}The straight tunnel with a squared cross-sectional shape of 0.6 m edge length for flight tests.}
\end{center}
%\vspace{-0.7cm}
\end{figure}

\begin{figure}[t]
\begin{center}         
\subfigure[\label{fig:straight_error} The maximum control tracking error.]
{\includegraphics[height=0.38\columnwidth]{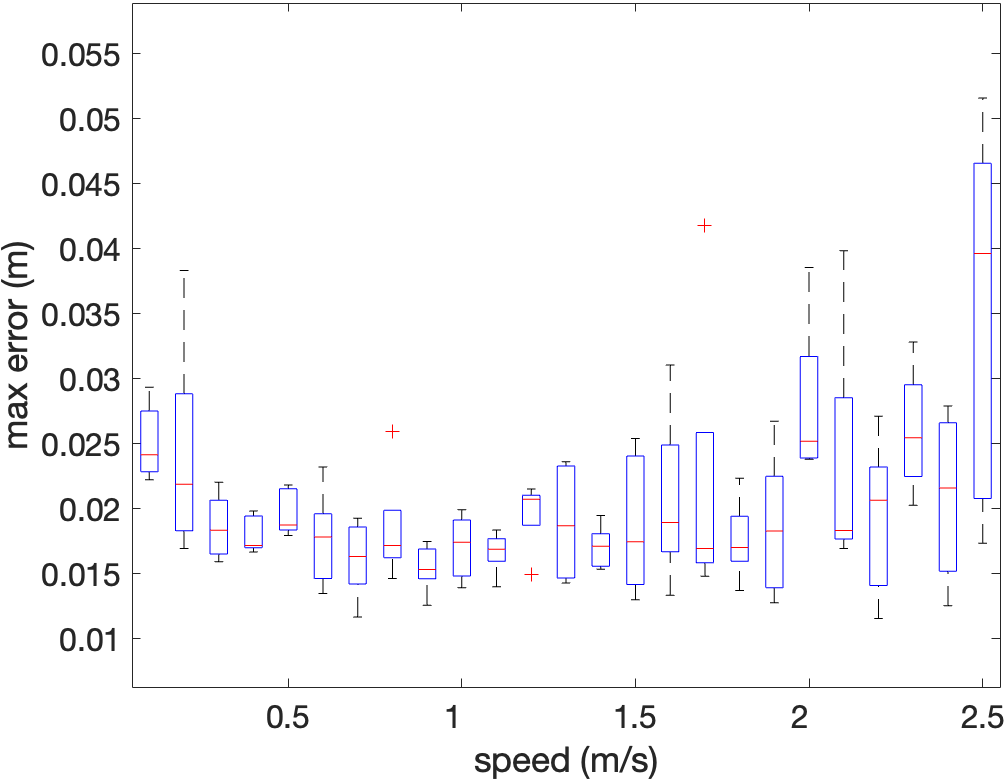}}
\subfigure[\label{fig:straight_feature} The minimum number of features.]
{\includegraphics[height=0.38\columnwidth]{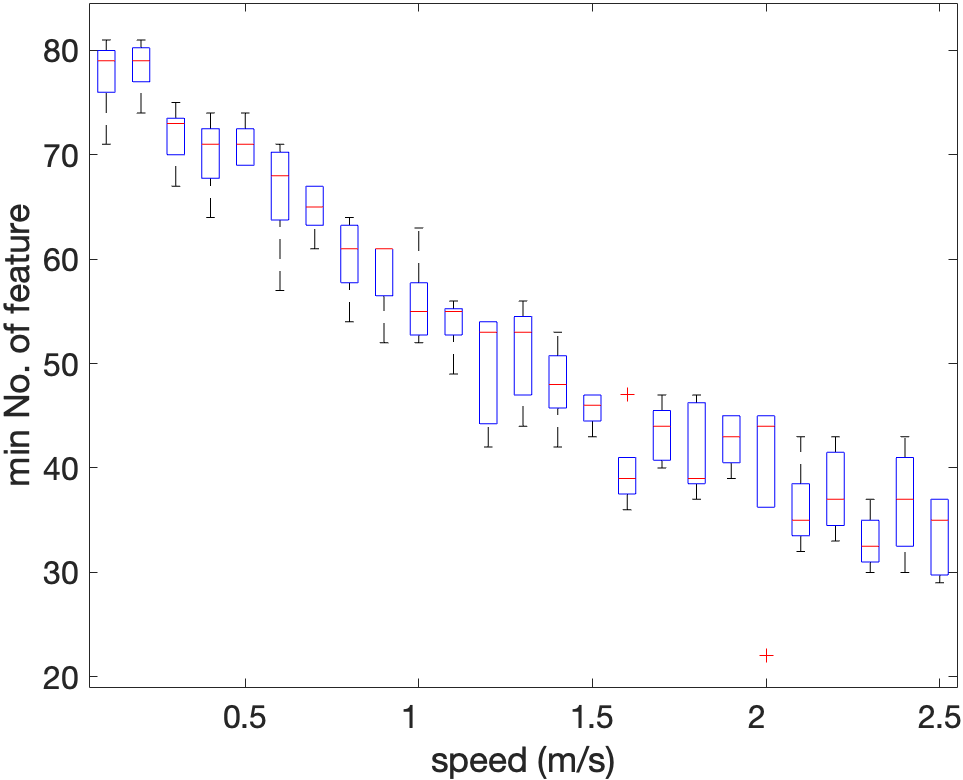}}
\end{center}
\vspace{-0.5cm}
\caption{\label{fig:straight_error_feature}The box plots of the maximum control tracking error on the cross-section planes and the minimum number of features that can be tracked by the state estimation system during the flights in the straight tunnel shown in Fig. \ref{fig:tunnel_straight}.}
\vspace{-1.0cm}
\end{figure}

\begin{figure}[t]
\begin{center}
\subfigure[\label{fig:straight_tunnel_v02_viz} Slow flight speed at 0.2 m/s.]
{\includegraphics[width=0.47\columnwidth]{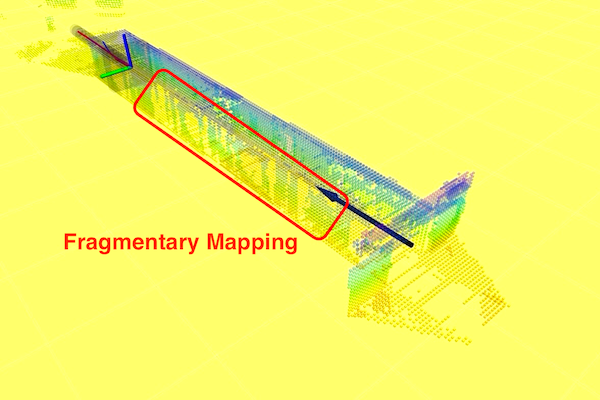}} 
\subfigure[\label{fig:straight_tunnel_viz} Proposed method.]
{\includegraphics[width=0.47\columnwidth]{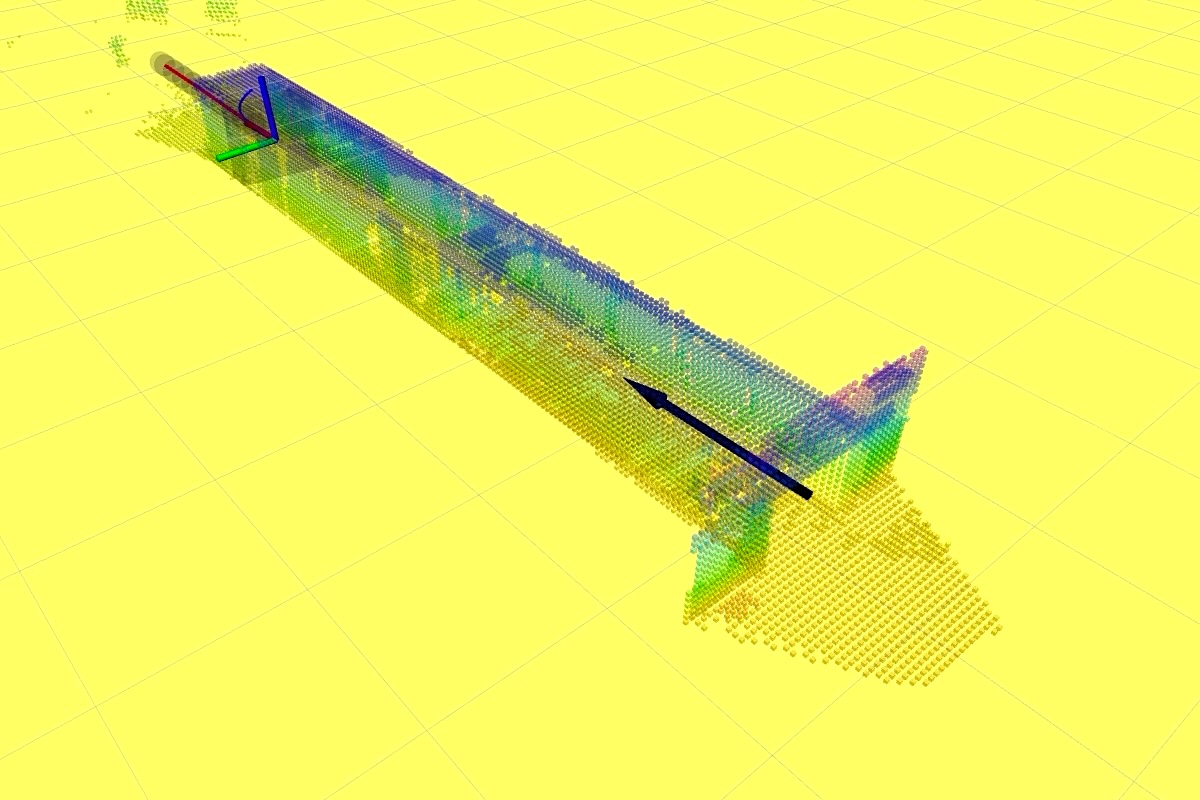}} 
\end{center}
\vspace{-0.4cm}
\caption{\label{fig:straight_tunnel_comp_viz} The real-time constructed map of the quadrotor flying through the straight tunnel shown in Fig. \ref{fig:tunnel_straight}. Fragmentary mapping results during the slow speed flight can be easily recognized in the red rectangle.
%Visualization screenshots of the quadrotor flying through the straight tunnels shown in Fig. \ref{fig:tunnel_straight}. The color coding indicates the height of the real-time constructed map, and the black arrows show the estimated tunnel entrance. The axes indicate the current pose of the quadrotor.
}
%\vspace{-0.9cm}
\end{figure}

\begin{figure}[t]
\begin{center}
{\includegraphics[width=1.0\columnwidth]{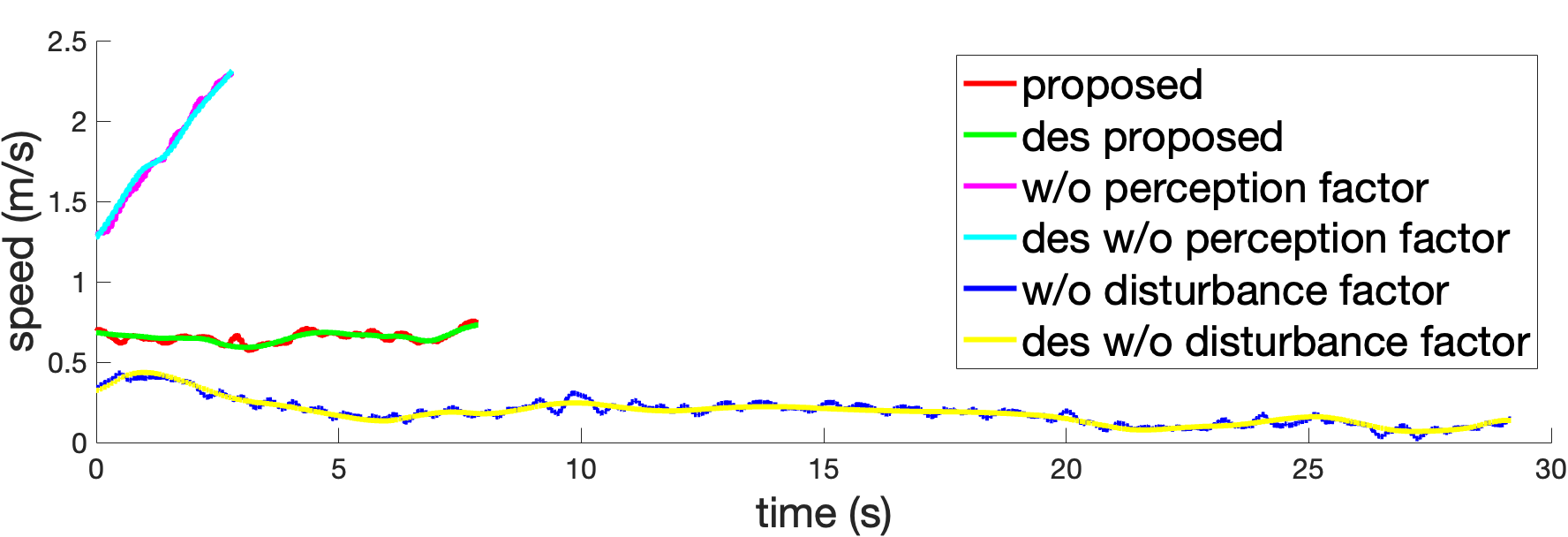}} 
\end{center}
\vspace{-0.3cm}
\caption{\label{fig:straight_speed}The speed profiles in the ablation study during straight tunnel flights.}
\vspace{-0.3cm}
\end{figure}

The first two sets of experiments are conducted in a straight 6-meter-long tunnel with a squared cross-sectional shape of 0.6 m edge length to examine the characteristics of the proposed system.

The first set of experiments consist of 125 flights are performed at constant flight speeds ranging from 0.1 m/s to 2.5 m/s in 0.1 m/s intervals, with each speed tests five times on the proposed quadrotor platform. As depicted in the control tracking error and feature number plots (Fig. \ref{fig:straight_error_feature}), the results align with previous studies \cite{wang2022neither} despite configuration differences. Large tracking errors occur at slow speeds due to ego airflow disturbances and at high speeds due to modeling errors and actuator limitations. Additionally, the minimum number of tracked features decreases as speed increases due to large parallax and motion blur. Furthermore, the experiments highlights that slow speeds can cause shaking movements, leading to fragmentary mapping results, as shown in Fig. \ref{fig:straight_tunnel_v02_viz}, which are unfavorable during tunnel flights. The consensual results underscore the necessity of compensating for perception and ego airflow disturbance factors, validating the motivations behind this work.

The second set of experiments involve ablation studies on the perception and ego airflow disturbance costs in the same straight tunnel. The proposed method and the ablation of either cost are each tested five times to gather control tracking error and feature tracking results. As shown in Tab. \ref{tab:traj_benchmark}, the proposed method achieves the minimum cross-section tracking error compared to both ablation scenarios. Note that we mainly focus on the errors in cross-sections since small longitudinal errors generally do not cause safety issues. Notably, the results also indicate that the proposed method attains comparable feature tracking numbers with the ablation of the disturbance cost and significantly higher numbers compared to the ablation of the perception cost. This is attributed to the automatic compensation of the perception and disturbance factors, ensuring that the quadrotor flies neither too fast or too slow through the tunnel, as depicted in Fig. \ref{fig:straight_speed}. The appropriate flight speed leads to a moderate optical flow speed and trajectory duration. These results preliminarily verify the functionality of the system in straight tunnels.

\subsection{2-D Tunnels}
\label{subsec:2d_tunnel_result}

\begin{table*}[h]
\centering
\caption{\label{tab:traj_benchmark}Ablation Study on Perception and Ego Airflow Disturbance Costs}
\begin{tabular}{@{}lccccccccc@{}}
\toprule
\multirow{4}{*}{Seq.} & \multicolumn{3}{c}{Straight} & \multicolumn{3}{c}{2-D tunnel case 1}  & \multicolumn{3}{c}{3-D tunnel case 1} \\
\cmidrule(lr){2-4} \cmidrule(lr){5-7} \cmidrule(lr){8-10}
& Proposed & $\lambda_{p} = 0$ & $\lambda_{ed} = 0$  & Proposed & $\lambda_{p} = 0$ & $\lambda_{ed} = 0$ & Proposed & $\lambda_{p} = 0$ & $\lambda_{ed} = 0$ \\
\midrule
Max error in cross-section (m)  & \textbf{0.0203}  & 0.0301 & 0.0228 & \textbf{0.0401} & $\times$ & 0.0407 & \textbf{0.0537} & $\times$ & 0.0554 \\
Min No. feature  & 72.8  & 43.1 & \textbf{78.4}  & \textbf{32.2} & $\times$ & 29.4 & 36.2 & $\times$ & \textbf{53.2} \\
Avg. optical flow speed (pixel/s) & 0.3935 & 0.8085 & \textbf{0.2811} & 0.5071 & $\times$ & \textbf{0.423} & 0.1858* & $\times$ & \textbf{0.1497}* \\
Traj. duration (s) & 13.411 & \textbf{5.9076} & 32.7925 & \textbf{24.103} & $\times$ & 52.6244 & \textbf{20.2722} & $\times$ & 31.207 \\
\bottomrule
\multicolumn{4}{l}{$\times$: fail. \textbf{Bold}: best results. *:Using three cameras and calculated by Eq. \ref{eq:of_parallel}.}
\end{tabular}
\vspace{-0.4cm}
\end{table*}

\begin{figure}[t]
\begin{center}
\subfigure[\label{fig:ra_shape_circle_7_rviz} Visualization.]
{\includegraphics[width=0.6\columnwidth]{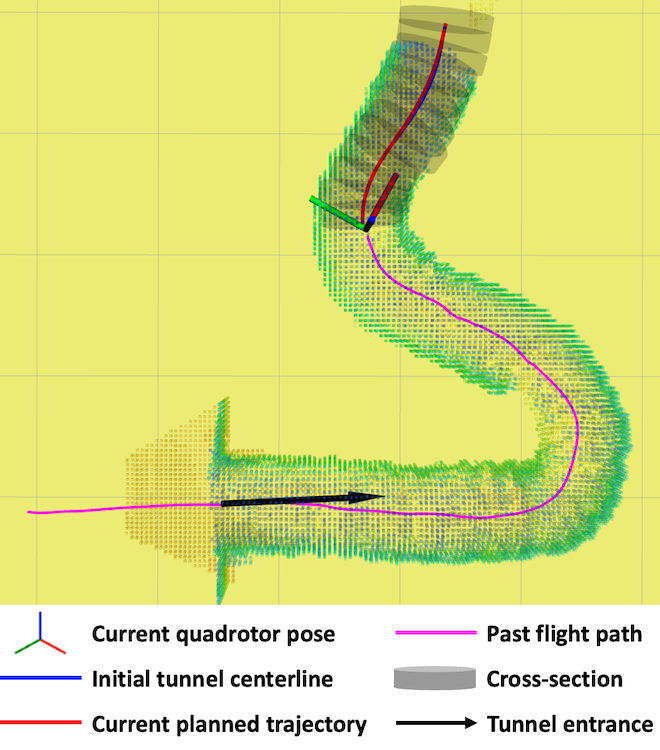}}
\subfigure[\label{fig:ra_shape_circle_7_speed} Speed profile.]
{\includegraphics[width=0.6\columnwidth]{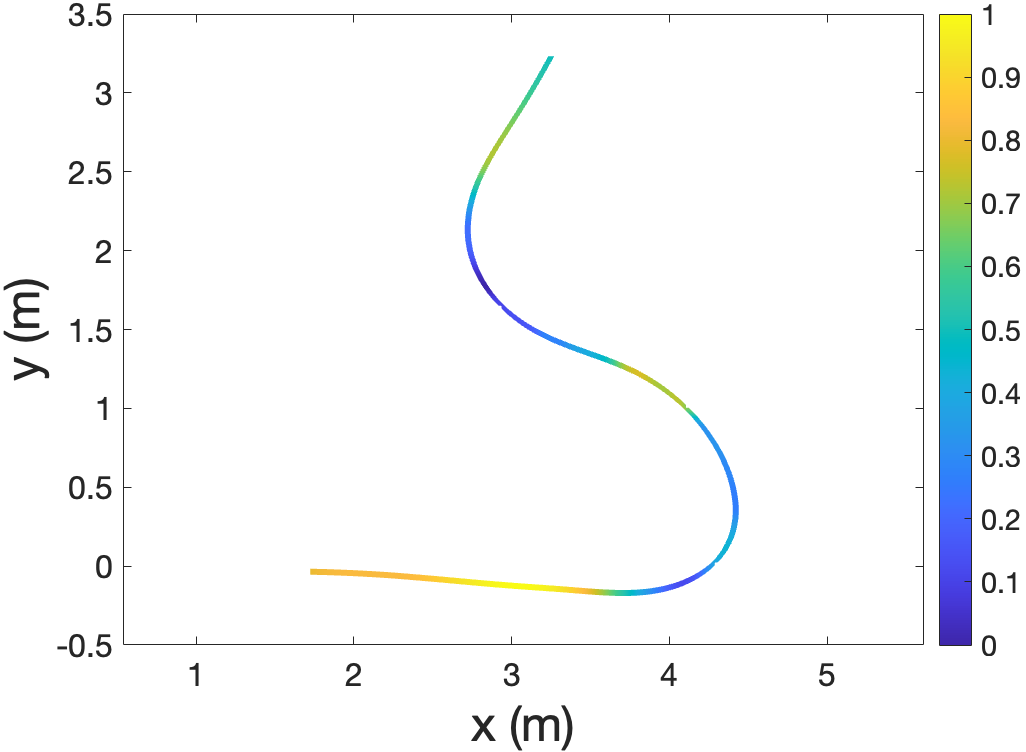}}
\end{center}
\vspace{-0.4cm}
\caption{\label{fig:ra_shape_circle_7_viz_speed} Visualization screenshots and the speed profile of the quadrotor flying through the 2-D tunnel case 1 shown in Fig. \ref{fig:tunnel_ra}. The color coding in the visualization indicates the height of the real-time constructed map, and the black arrow show the estimated tunnel entrance; the axes indicate the current pose of the quadrotor, while the pink, blue and red lines indicates the past flight path, the initial tunnel centerline and the current planned tunnel-following trajectory respectively. The extracted and predicted cross-sections are shown half-transparently.}
%\vspace{-0.9cm}
\end{figure}

\begin{figure}[t]
\begin{center}
\subfigure[\label{fig:ra_shape_circle_7_view_1} Proposed method.]
{\includegraphics[height=0.35\columnwidth]{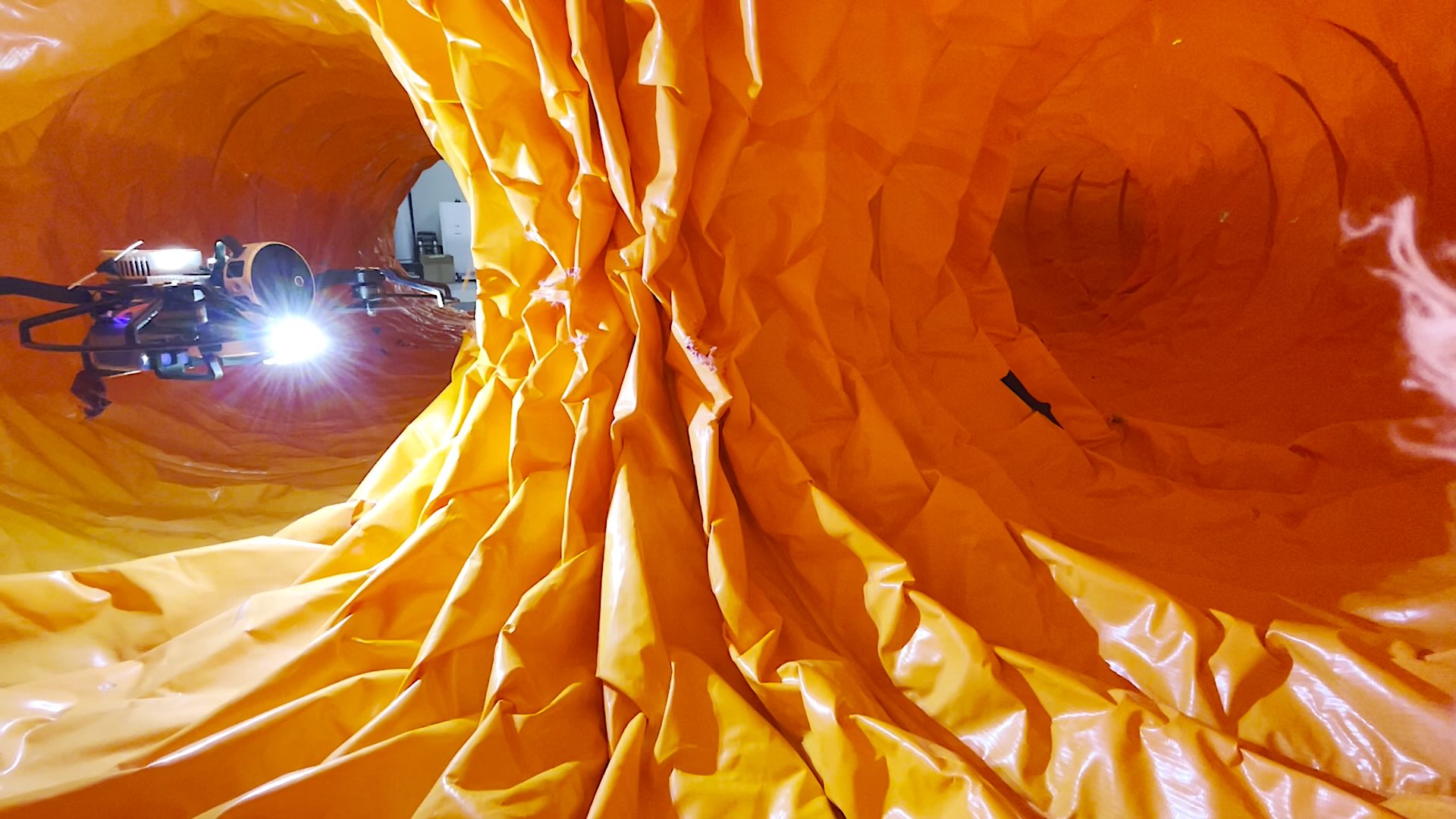}
\includegraphics[height=0.35\columnwidth]{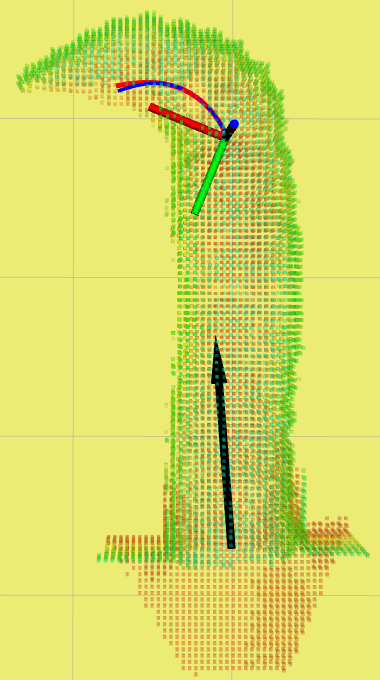}} 
%\subfigure[\label{fig:ra_shape_circle_7_viz} Visualization.]
%{\includegraphics[height=0.35\columnwidth]{ra_shape_circle_7_rviz}} 
\subfigure[\label{fig:ra_shape_circle_7_wo_yaw_view_1} Without active yaw planning.]
{\includegraphics[height=0.35\columnwidth]{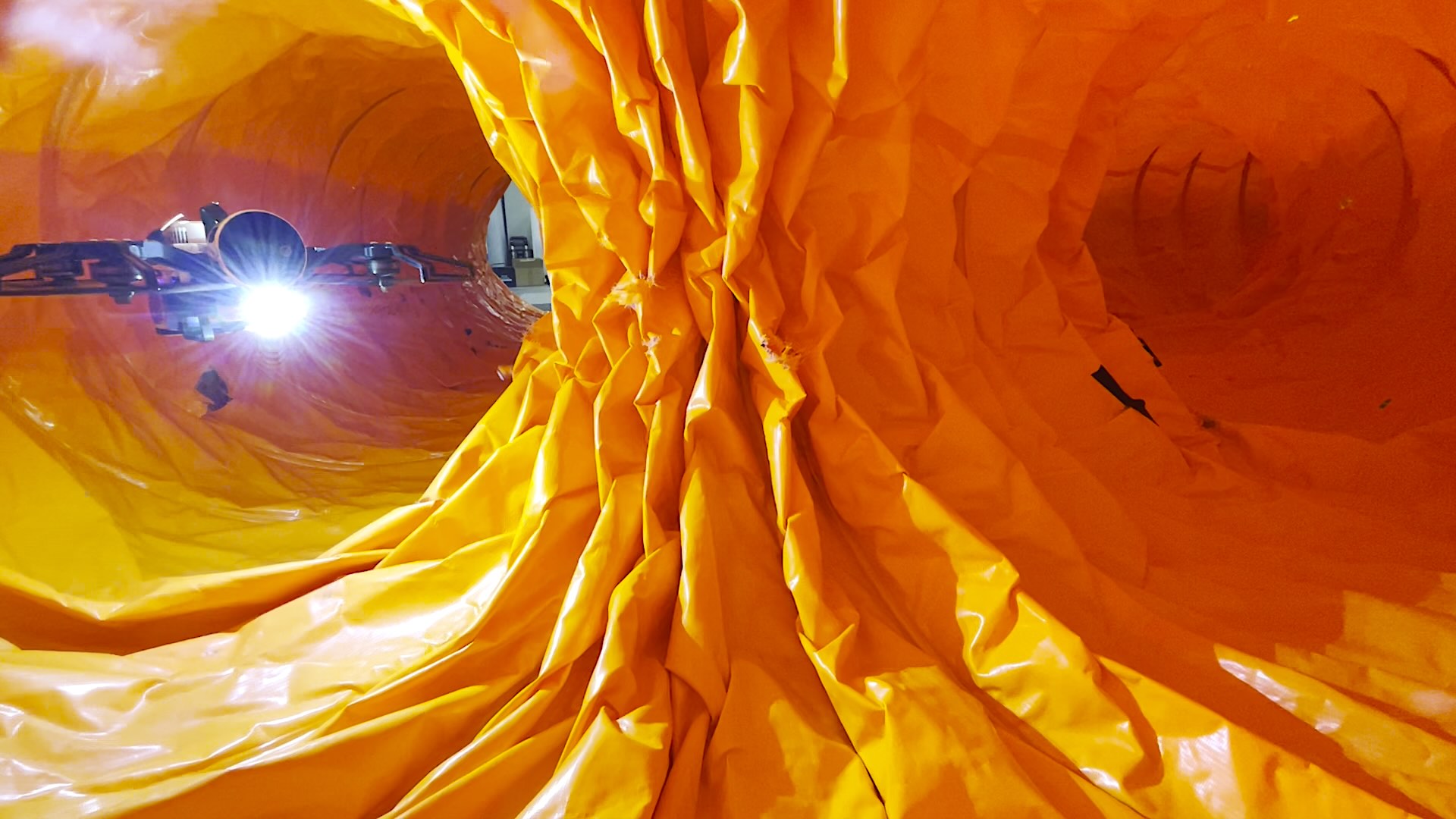}
\includegraphics[height=0.35\columnwidth]{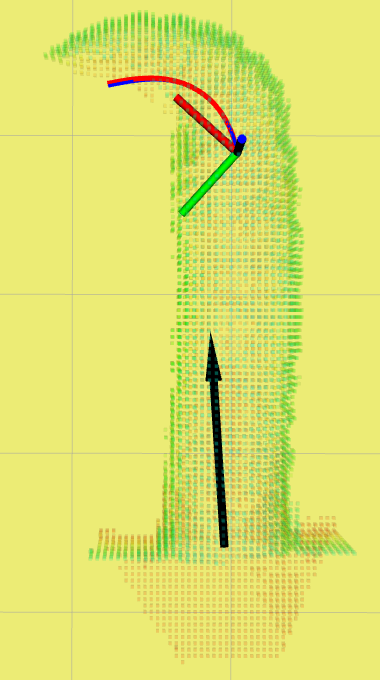}}
%\subfigure[\label{fig:ra_shape_circle_7_wo_yaw_rviz} Visualization.]
%{\includegraphics[height=0.35\columnwidth]{ra_shape_circle_7_wo_yaw_rviz}} 
\end{center}
\vspace{-0.4cm}
\caption{\label{fig:ra_shape_circle_7_yaw} Snapshots of the quadrotor flying through the first corner using the proposed method and the method without active yaw planning in the 2-D tunnel case 1 shown in Fig. \ref{fig:tunnel_ra}. The markers are the same as previous figures. It can be clearly seen that the quadrotor using the proposed method turns the yaw at a larger angle.}
%\vspace{-0.9cm}
\end{figure}

The next three sets of experiments are conducted in three different 2-D tunnels constructed using one or more flexible vent pipes commonly employed in industrial settings, as shown in Fig. \ref{fig:tunnel_ra} - \ref{fig:tunnel_2d}.

We first perform an ablation study focusing on ego airflow disturbance, perception costs, and perception-aware active yaw planning in the 2-D tunnel case 1. This scenario, shown in Fig.~\ref{fig:tunnel_ra}, features a flexible vent pipe with circular cross-sections ranging from 0.6 to 0.7 meters in diameter. Each configuration is tested five times to avoid accidental results, and an overview on the visualization result using the proposed method is shown in Fig. \ref{fig:ra_shape_circle_7_rviz}.

According to the results in Tab. \ref{tab:traj_benchmark}, removing the perception cost consistently leads to failures at the first turn. This is because the increased ego airflow disturbance causes the quadrotor to accelerate until it completely avoids the ego airflow, regardless of the tunnel's curvature. Conversely, removing the ego airflow disturbance cost leads to unnecessarily slow flight speeds within the tunnel, causing the quadrotor to shake due to the turbulent flow it generates. The proposed perception-and-disturbance-aware speed profile planning automatically compensate for the two factors by reducing speed at turns and appropriately increasing speed in sections with small curvatures (Fig. \ref{fig:ra_shape_circle_7_speed}). This approach achieve a minimum control tracking error and maintain a sufficient number of tracked features for state estimation. Additionally, it achieves higher efficiency in terms of traversal time compared to configurations without ego airflow disturbance cost. These results demonstrate the effectiveness of the proposed speed profile planning. Furthermore, we compared the proposed method with the ablation of perception-aware active yaw planning, where the yaw direction always points towards a point at a constant distance ahead on the trajectory. In five trails without the active yaw planning, the quadrotor consistently stops at the first turn because the yaw headings never turn enough to provide an adequate view of the tunnel ahead. In contrast, the proposed method leverages active yaw planning to successfully traverse the tunnel in all five trials. The heading differences during one of the comparison experiments are clearly illustrated in Fig.~\ref{fig:ra_shape_circle_7_yaw}.

\begin{figure}[t]
\begin{center}
\subfigure[\label{fig:n_shape_circle_7_rviz} Visualization.]
{\includegraphics[width=0.8\columnwidth]{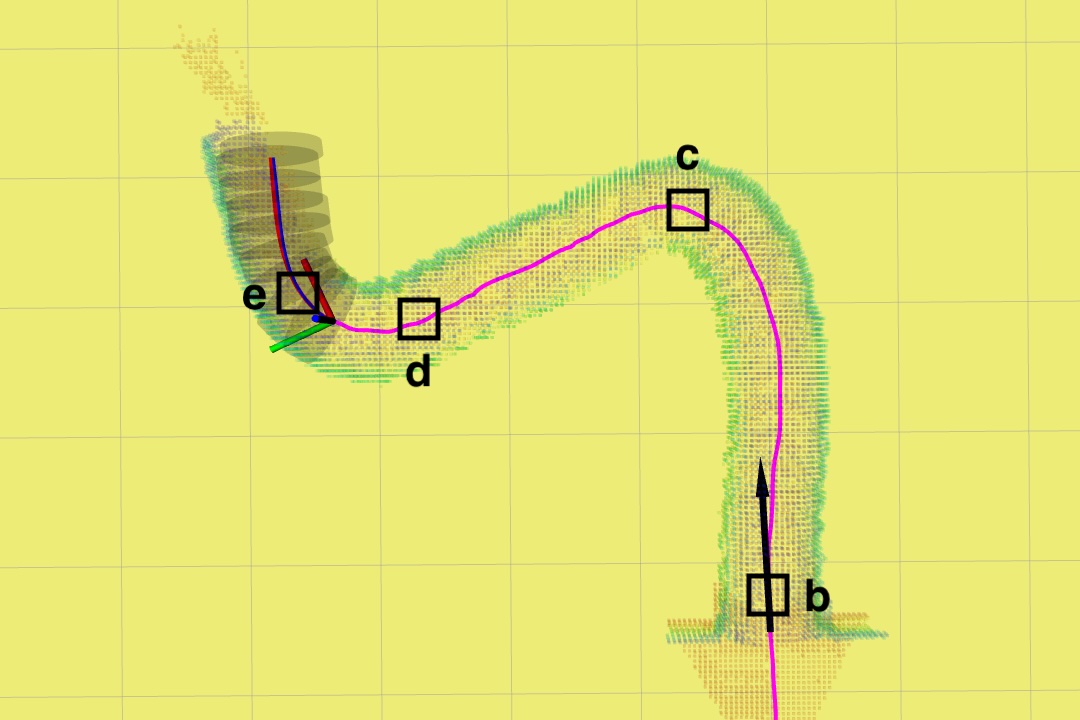}} 
\subfigure[\label{fig:n_shape_circle_7_view_0} Flying into the entrance.]
{\includegraphics[width=0.47\columnwidth]{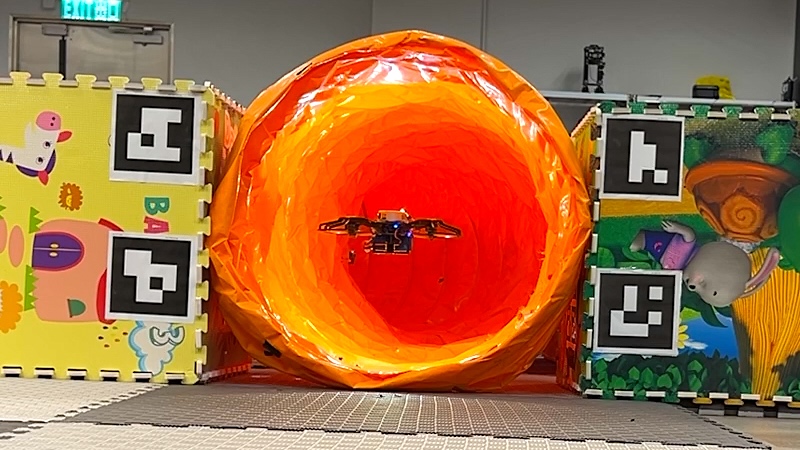}}
\subfigure[\label{fig:n_shape_circle_7_view_1_1} Flying through turn 1.]
{\includegraphics[width=0.47\columnwidth]{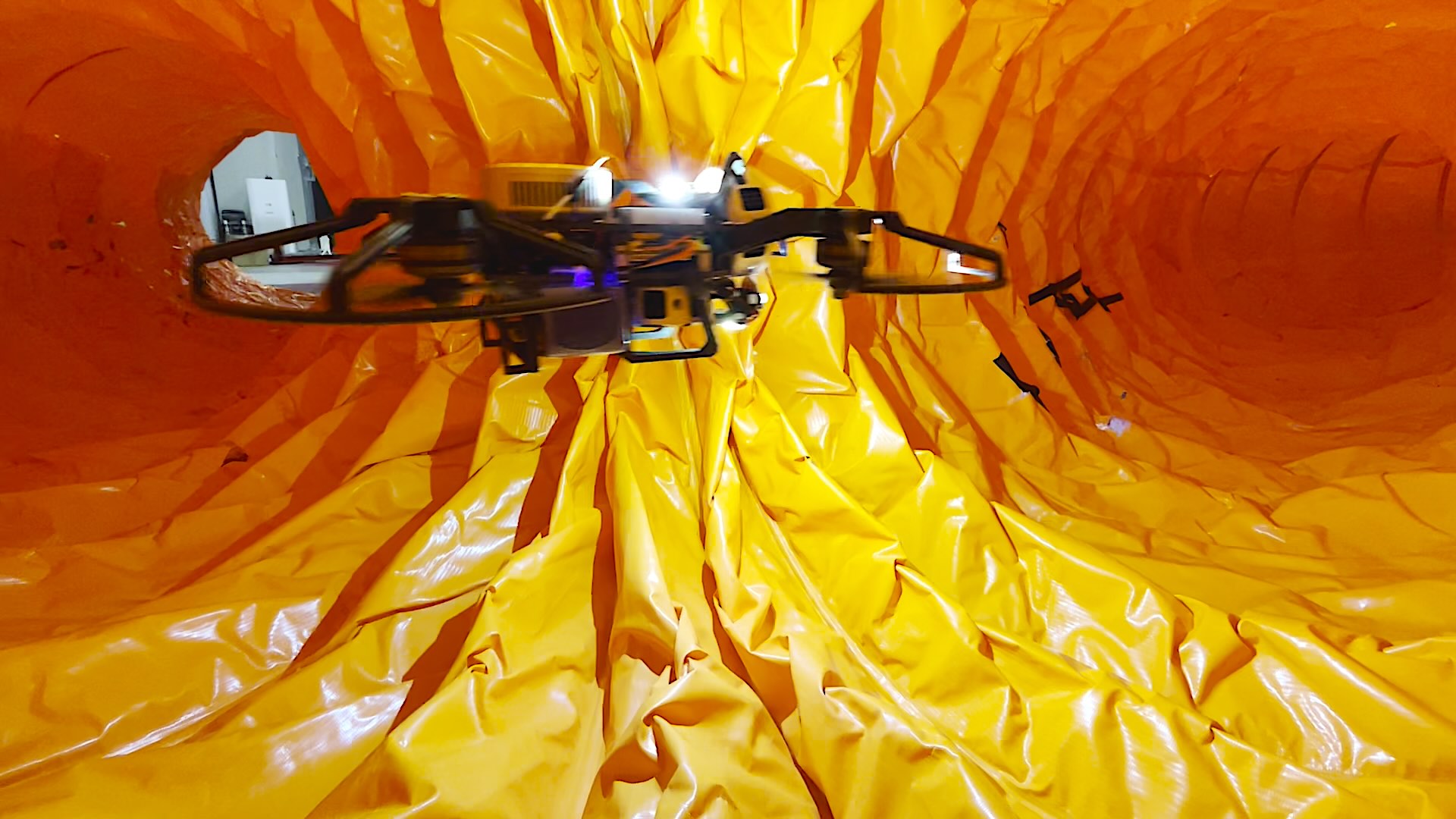}} 
\subfigure[\label{fig:n_shape_circle_7_view_1_2} Flying through turn 2.]
{\includegraphics[width=0.47\columnwidth]{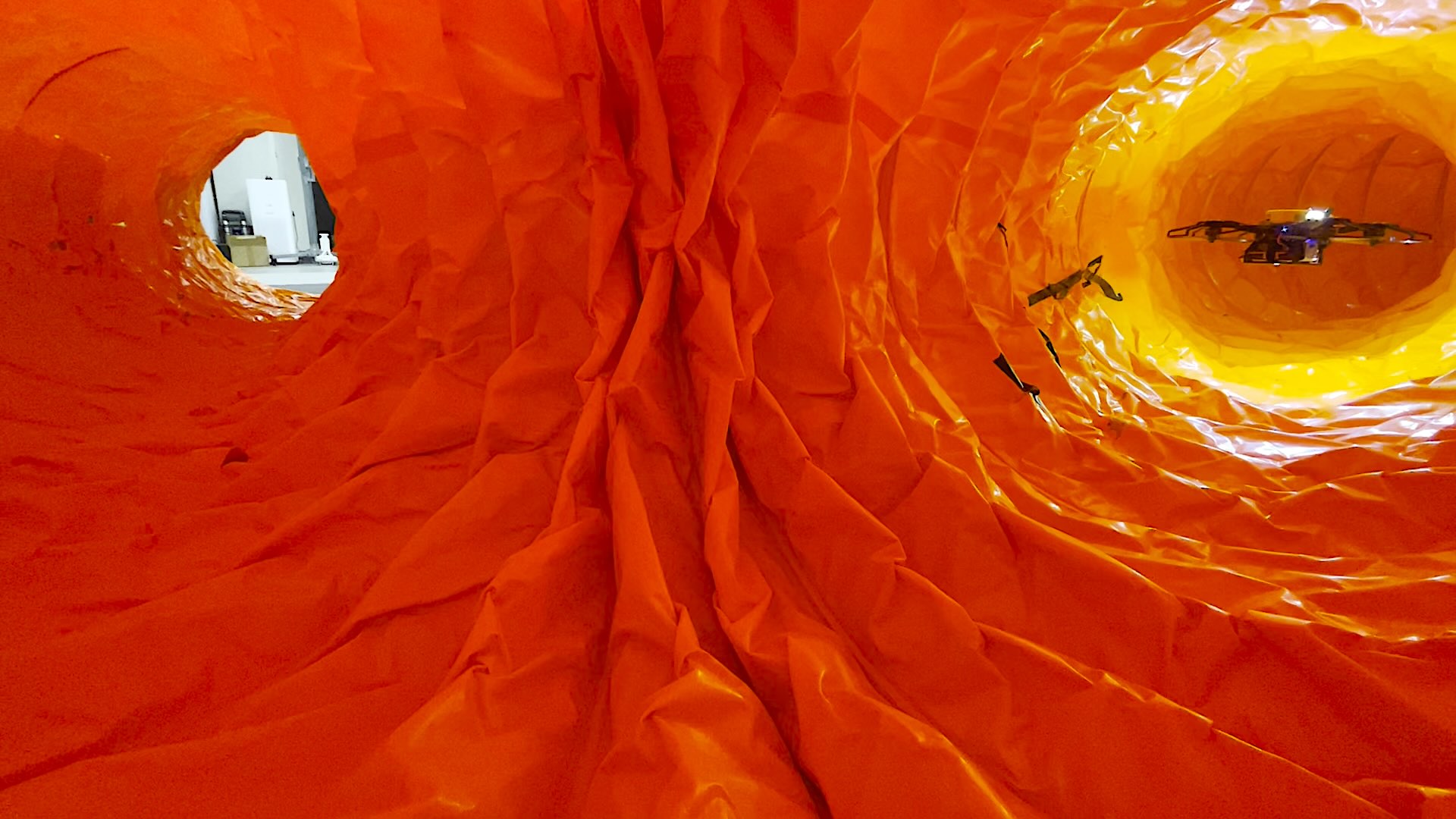}}
\subfigure[\label{fig:n_shape_circle_7_view_2} Flying out of the exit.]
{\includegraphics[width=0.47\columnwidth]{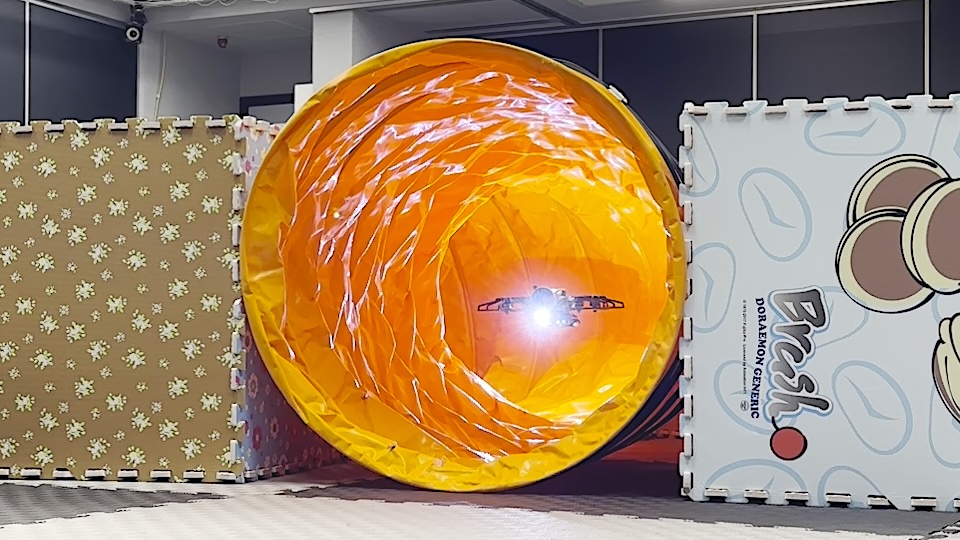}}
\end{center}
\vspace{-0.4cm}
\caption{\label{fig:n_shape_circle_7} Snapshots and visualization results when the quadrotor flying through the 2-D tunnel shown in Fig. \ref{fig:tunnel_n}. The markers are the same as previous figures and the quadrotor positions in the snapshots are labelled in the visualization.}
\vspace{-0.5cm}
\end{figure}

We then compare the proposed system with the system in \cite{wang2022neither} and manual flight using a commercial FPV system, as shown in Fig. \ref{fig:old_drone_fpv}, in another 2-D tunnel using the same vent pipe yet of a different shape, as shown in Fig. \ref{fig:tunnel_n}. Each configuration is tested five times. The snapshots and visualization results of the proposed method are shown in Fig. \ref{fig:n_shape_circle_7}.

During the five flights using the proposed system, no failures are reported, while the system in \cite{wang2022neither} always crashes at the first shape turn despite the known 0.7 m tunnel diameter and a proper flight speed of 1 m/s from its previous results. The system in \cite{wang2022neither} is unable to handle the large curvature and narrower dimensions caused by the folds of the pipe at the turn in all five trails, as it lacks cross-section recognition, active yaw planning, and speed profile planning. Additionally, the FPV drone operated by an experienced pilot is kept sucking to the tunnel wall during the manual flight and eventually crashes inside the tunnel in all five trails, despite its smaller size compared to the proposed system. This set of experiment demonstrates the superior performance of the proposed system against the system in \cite{wang2022neither} and experienced human pilots.

\begin{figure}[t]
\begin{center}
\subfigure[\label{fig:2d_multi_dim_rviz} Overview from the proposed system.]
{\includegraphics[width=0.8\columnwidth]{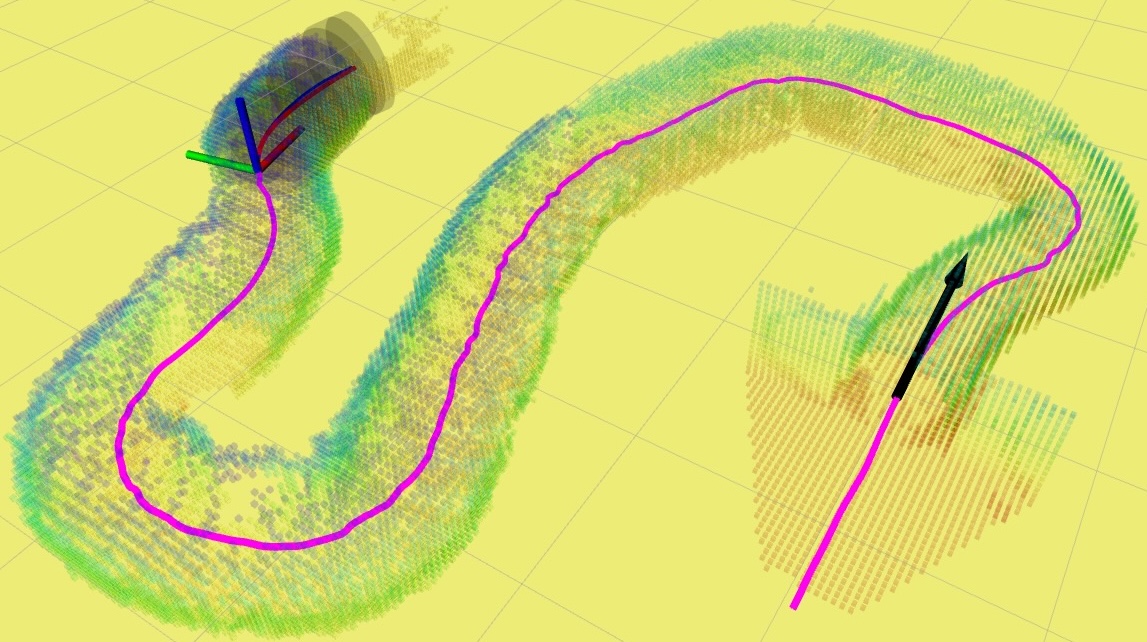}} 
\subfigure[\label{fig:2d_multi_dim_ori_rviz} \cite{wang2022neither}.]
{\includegraphics[height=0.5\columnwidth]{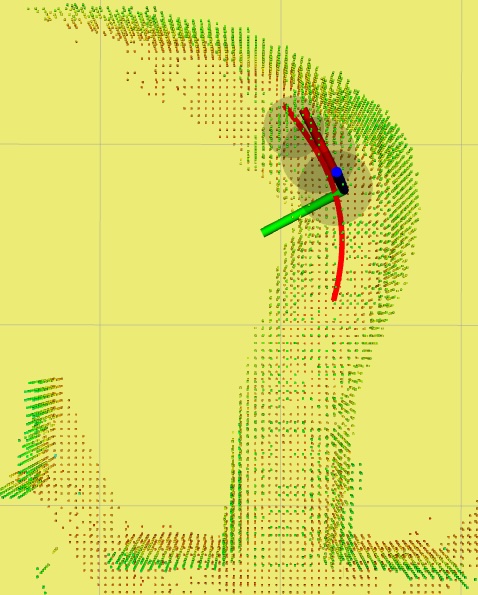}}
\subfigure[\label{fig:2d_multi_dim_v07_rviz} Proposed system with constant speed of 0.7 m/s.]
{\includegraphics[height=0.5\columnwidth]{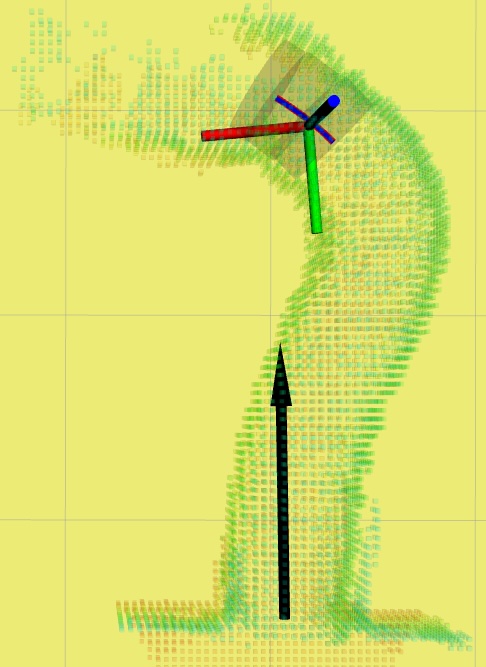}}
\end{center}
\vspace{-0.4cm}
\caption{\label{fig:2d_multi_dim} Visualization screenshots of the quadrotors flying through the 2-D tunnel shown in Fig. \ref{fig:tunnel_2d}. The markers are same as previous figures. The system in \cite{wang2022neither} crashes at the first turn and the constant speed method emergency stops at the first turn.}
%\vspace{-0.9cm}
\end{figure}

The final set of 2-D tunnel experiments is conducted using a series of flexible vent pipes with varying cross-sectional shapes: a circular section with diameters of 0.5 to 0.55 m, a rectangular section measuring 0.5 m by 0.65 m, and another circular section with diameters of 0.6 to 0.7 m,  as illustrated in Fig.~\ref{fig:tunnel_2d}. The performance of the proposed system is evaluated against the system described in \cite{wang2022neither}, manual flights by an experienced pilot, and the ablation of ego airflow disturbance and perception-aware speed profile planning, also known as the constant speed method. Each configuration is tested five times, except that more trails are conducted on manual flights.

As shown in Fig. \ref{fig:2d_multi_dim_rviz}, the proposed system successfully navigates through different sections and sharp turns in all five trials, demonstrating its robustness and adaptability. In contrast, as shown in Fig. \ref{fig:2d_multi_dim_ori_rviz}, the system in \cite{wang2022neither} fails at the first turn due to its inability to manage changes in cross-sectional dimensions and sharp turns, even at a flight speed of 1 m/s. Furthermore, despite the pilot's experience from previous tunnels, the manual flight crashes in all ten trials before the first turn of the tunnel, regardless of using a joystick motion controller or a conventional radio controller. This failure is attributed to stronger ego-airflow disturbances caused by the smaller cross-section size of the tunnel. The quadrotor is immediately sucked to the tunnel wall even when it is just several centimeters away from the centerline and there is almost no chance to recover. Although the constant speed method using the proposed system at 0.7 m/s does not result in crashes, it requires five emergency stops at the first turn due to safety concerns. The flight speed at the sharp turn is excessively fast, preventing timely map updates. These results highlight the proposed system's capability to handle tunnels with varying cross-sections and sharp turns effectively. Comparisons with the system in \cite{wang2022neither}, manual flight and the constant speed method further underscore the necessity and effectiveness of cross-section recognition, active yaw, and speed profile planning.

\subsection{3-D Tunnels}
\label{subsec:3d_tunnel_result}

\begin{figure}[t]
\begin{center}    
%\subfigure[\label{fig:vent_rviz} The minimum number of features tracked by the VIO system.]
{\includegraphics[width=0.8\columnwidth]{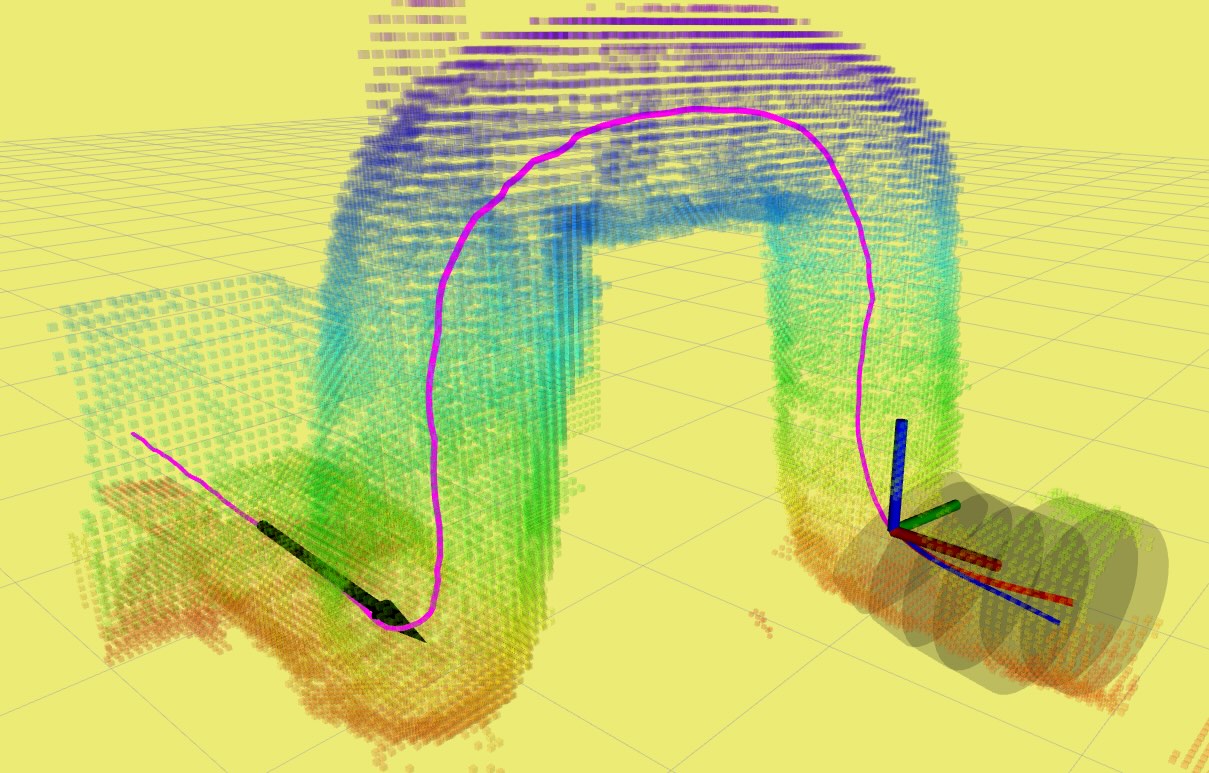}}
\caption{\label{fig:3d_circle_7_rviz}A screenshot of the visualization during flight of the proposed system inside 3-D tunnel case 1 shown in Fig. \ref{fig:tunnel_3d_circle_7}. The markers are same as previous figures.}
\end{center}
\end{figure}

\begin{figure}[t]
\begin{center} 
\subfigure[\label{fig:3d_multi_dim_rviz} Visualization.]
{\includegraphics[width=0.8\columnwidth]{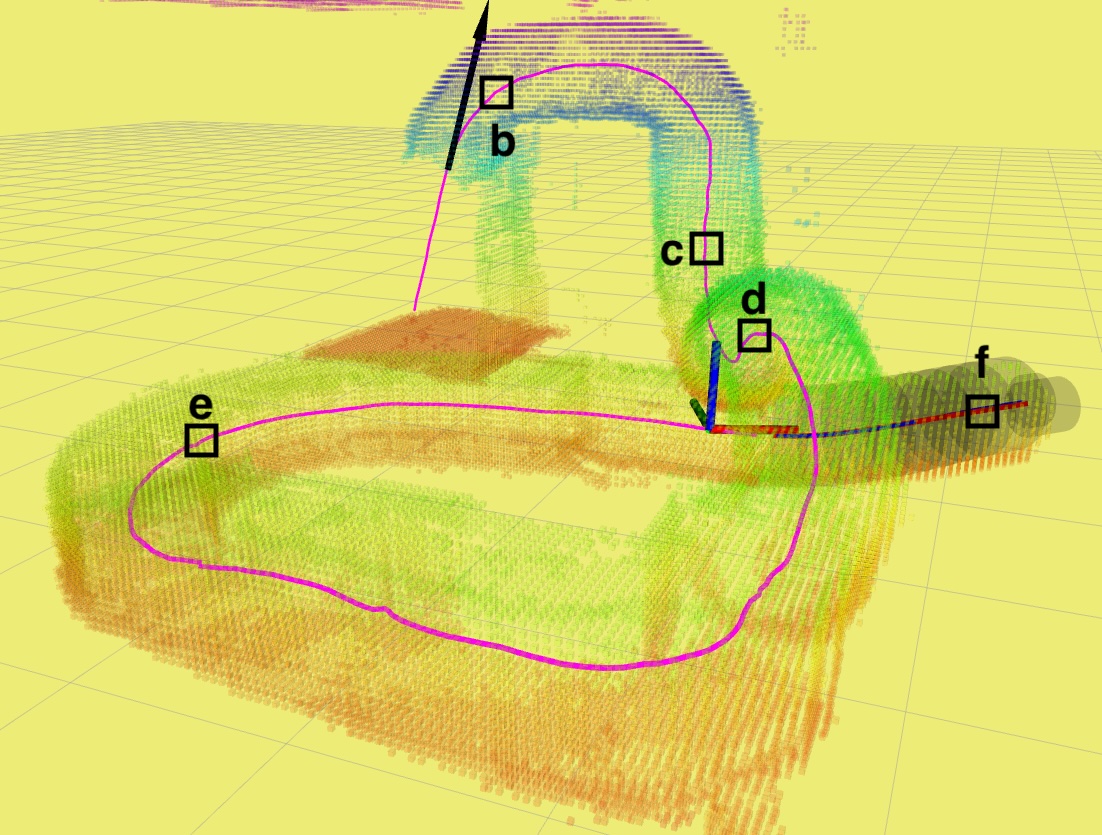}}        
\subfigure[\label{fig:3d_multi_dim_view_0} Flying into the entrance.]
{\includegraphics[width=0.32\columnwidth]{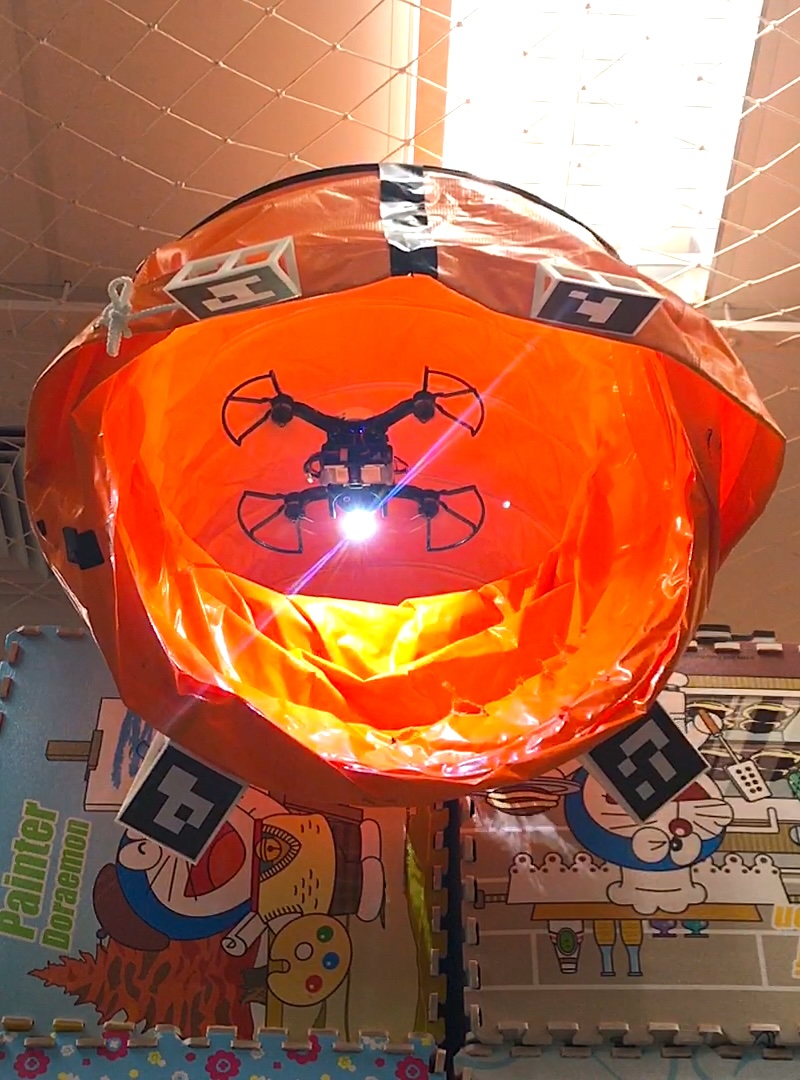}}
\subfigure[\label{fig:3d_multi_dim_view_1} Flying down the vertical section.]
{\includegraphics[width=0.32\columnwidth]{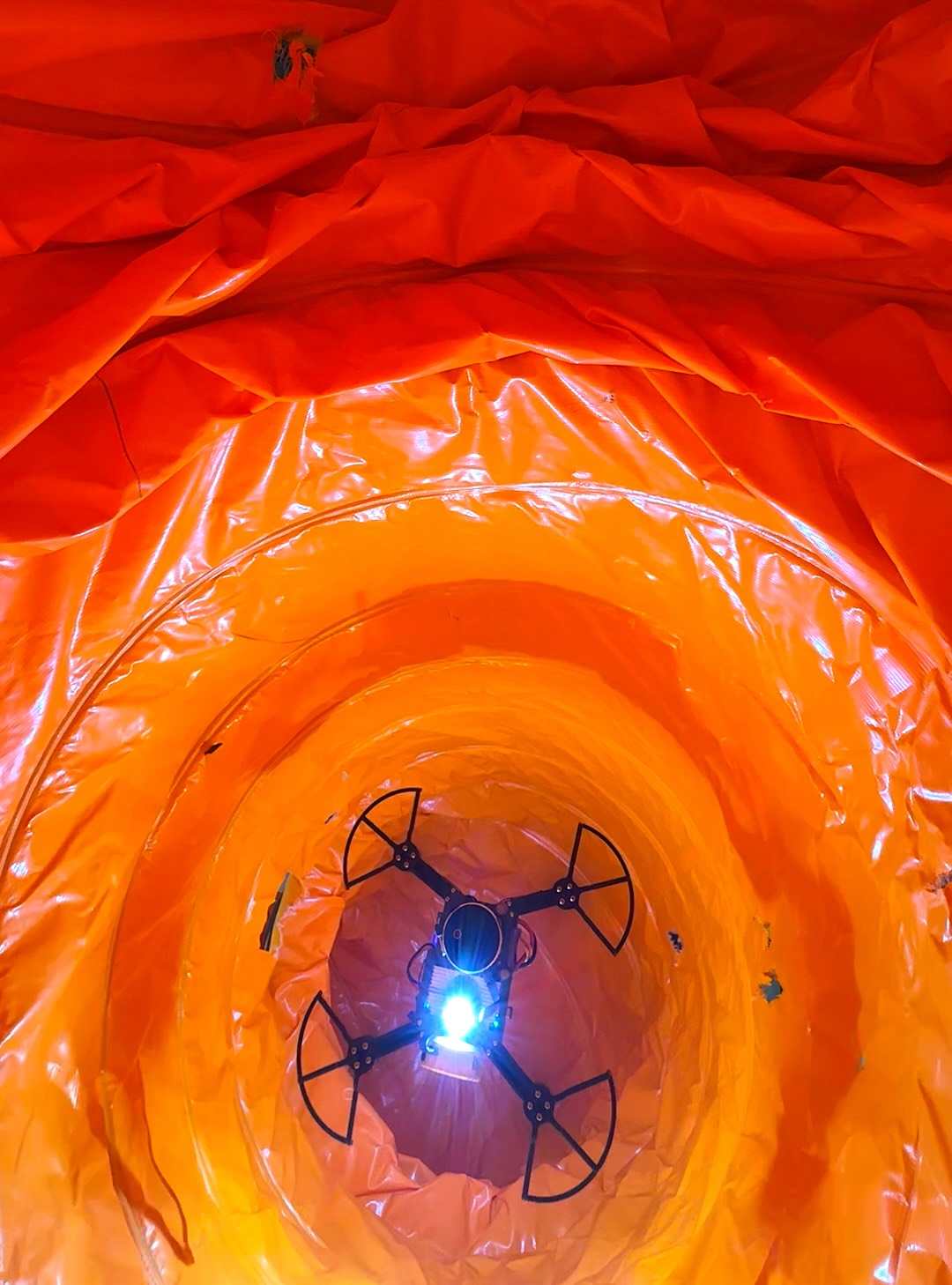}}
\subfigure[\label{fig:3d_multi_dim_view_2} Flying upwards along the slope.]
{\includegraphics[width=0.32\columnwidth]{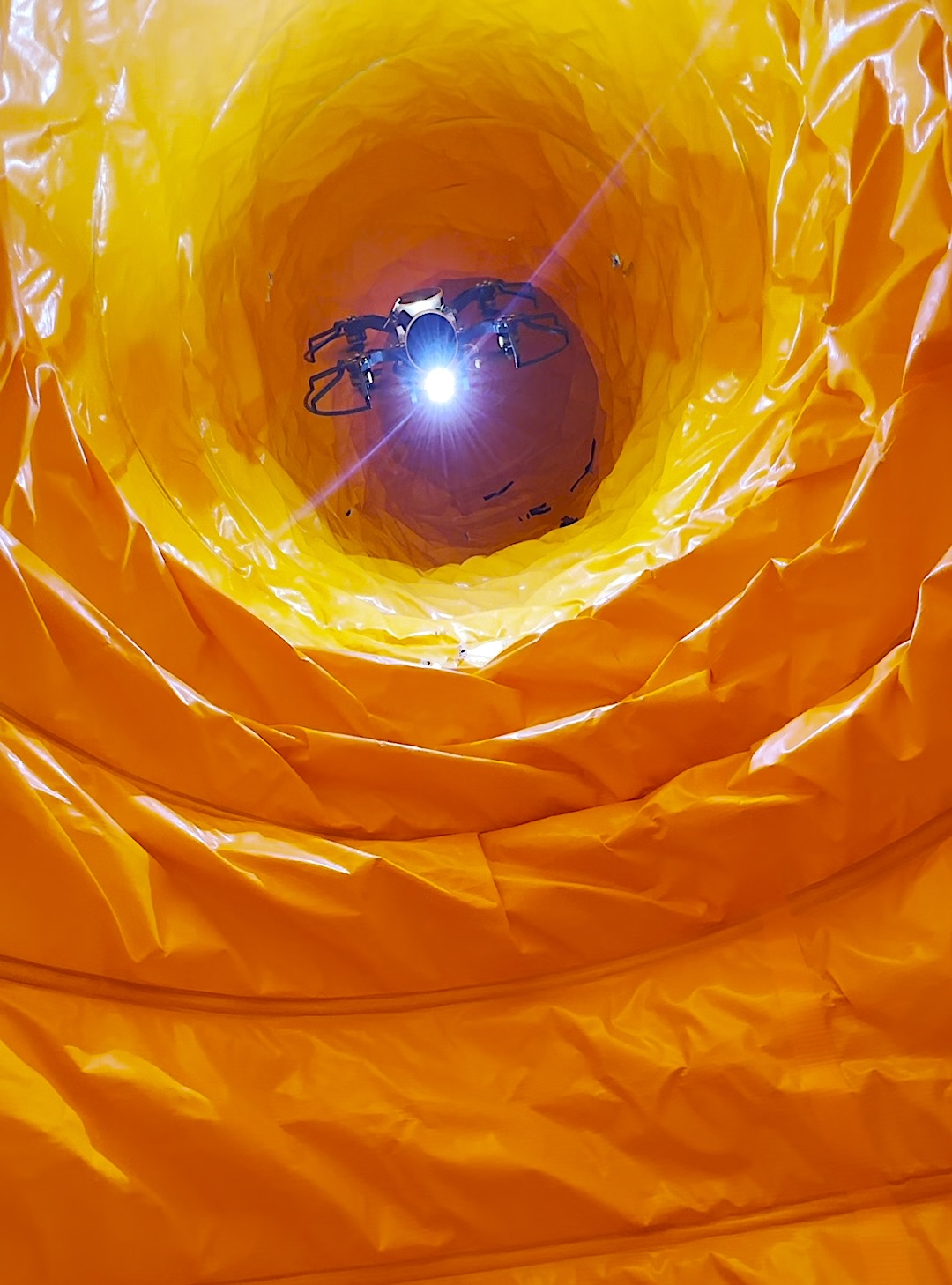}}
\subfigure[\label{fig:3d_multi_dim_view_4} Traversing the dark rectangular tunnel section.]
{\includegraphics[width=0.48\columnwidth]{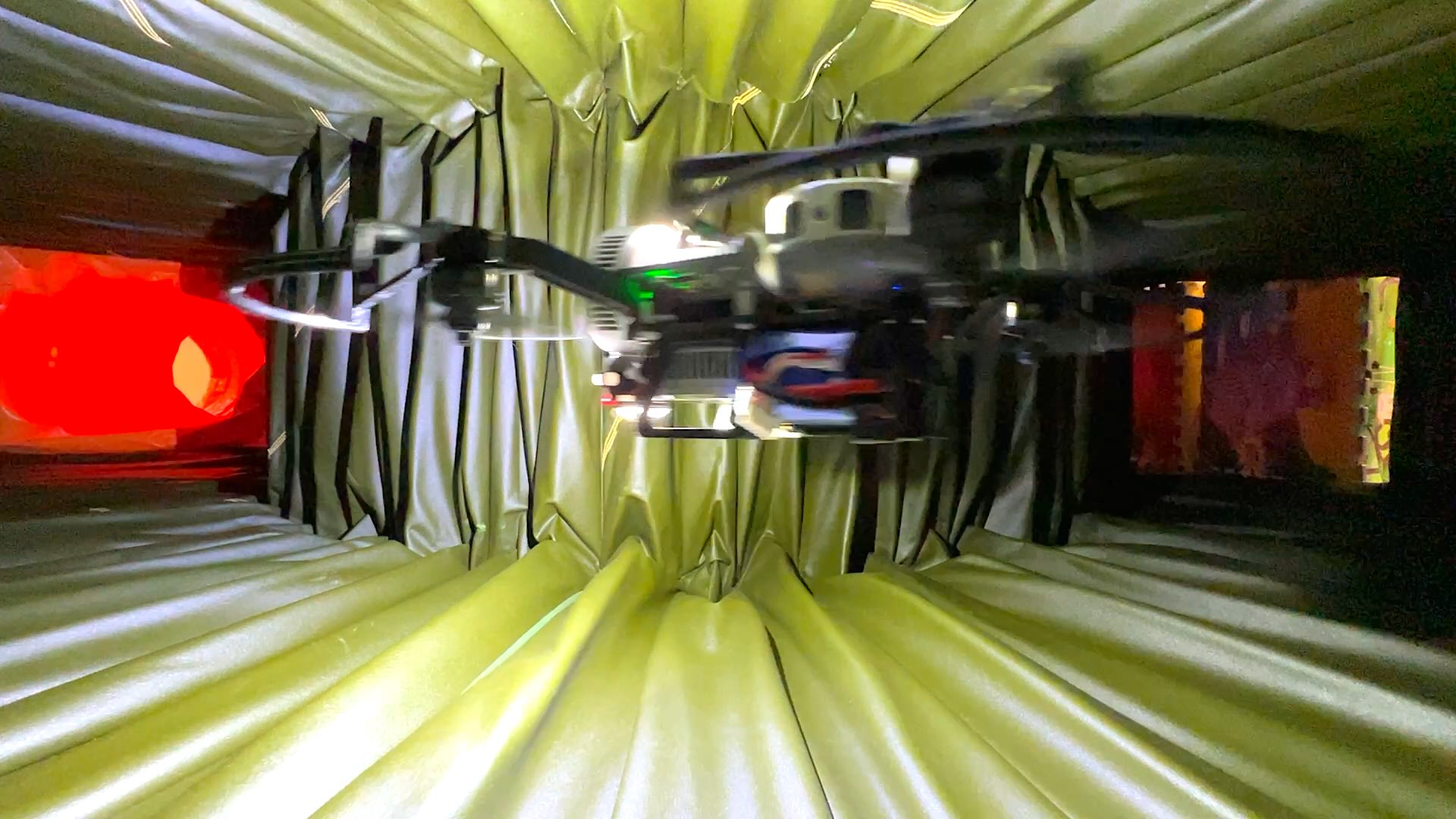}}
\subfigure[\label{fig:3d_multi_dim_view_5} Flying out of the exit.]
{\includegraphics[width=0.48\columnwidth]{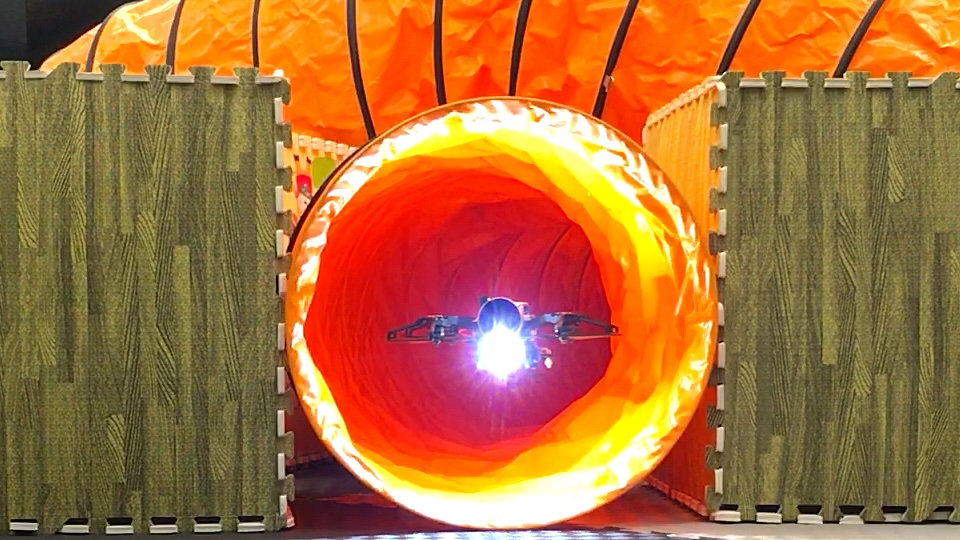}}
\end{center}
\vspace{-0.5cm}
\caption{\label{fig:3d_multi_dim}Snapshots and visualization results when the quadrotor flies through the 3-D tunnel case 2 shown in Fig. \ref{fig:tunnel_3d}. The markers are same as previous figures and the quadrotor positions in the snapshots are labelled in the visualization.}
\vspace{-0.3cm}
\end{figure}

Using the proposed system, we conduct further two sets of experiments in 3-D narrow tunnels with level variations, including vertical sections, as illustrated in Fig. \ref{fig:tunnel_3d_circle_7} - \ref{fig:tunnel_3d}.

The first set of flights involves an ablation study on the perception and ego airflow disturbance costs within a flexible vent pipe of 0.6 to 0.7 m diameter, previously used and reshaped to include two vertical sections as shown in Fig. \ref{fig:tunnel_3d_circle_7}. Each configuration is tested five times, similar to prior experiments, and a visualization screenshot is provided in Fig. \ref{fig:3d_circle_7_rviz}. As indicated in Tab. \ref{tab:traj_benchmark}, removing the perception cost results in failures in all five trials due to excessive speed at the first turn. Conversely, the ablation of the ego airflow disturbance cost leads to unnecessary slow flight speeds, thereby reducing efficiency in terms of traversal time. The proposed method automatically adjusts the speed by balancing both factors, achieving optimal control performance and the shortest flight time while maintaining sufficient tracked features for state estimation. This results in smooth and efficient traversal in all five trials. This set of ablation studies demonstrate the effectiveness of the proposed perception and ego airflow disturbance models and the entire system in tunnels with level variations, including vertical sections.

The second set of flights involves a comparison with manual flight in a more complex 3-D tunnel, as depicted in Fig.~\ref{fig:tunnel_3d}. This tunnel consists of sections with varying slopes and cross-sectional shapes, including circles and rectangles with dimensions ranging from 0.5 m to 0.9 m. The proposed quadrotor system successfully traverses the tunnel in 38 seconds, with snapshots and visualization results shown in Fig.~\ref{fig:3d_multi_dim}. Conversely, the experienced pilot operating the smaller commercial FPV drone, as shown in Fig.~\ref{fig:old_drone_fpv}, finds it extremely challenging to navigate the same tunnel, with the drone frequently being pulled toward the tunnel walls. The pilot takes 55 seconds to reach the downslope shown in Fig. \ref{fig:3d_multi_dim_view_2}, covering only about half of the tunnel length from the entrance, and eventually crashes at that slope. These comparisons demonstrate the superior robustness and adaptivity of the proposed system in complex scenarios, as well as its efficiency, outperforming even experienced pilots.

\subsection{Rigid Vent Pipe}
\label{subsec:vent_pipe_result}

\begin{figure}[t]
\begin{center}         
\subfigure[\label{fig:vent_rviz} Visualization.]
{\includegraphics[width=0.75\columnwidth]{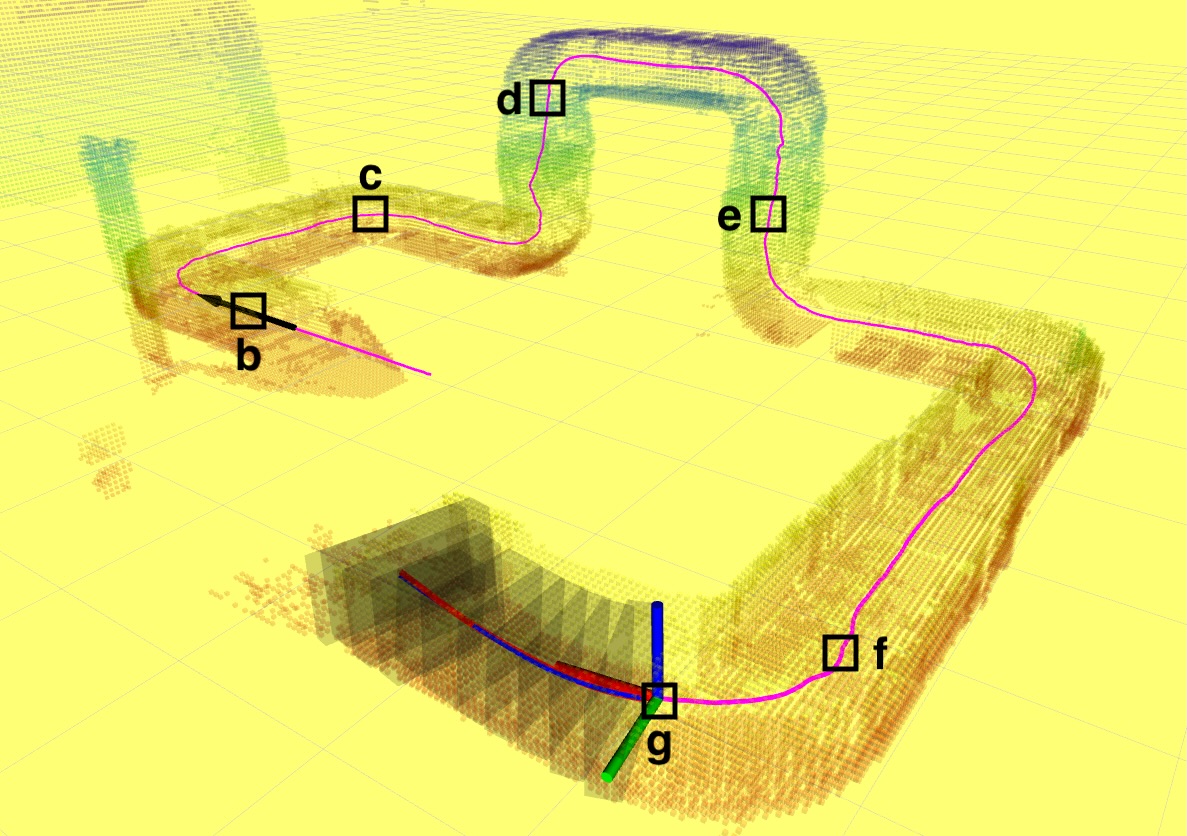}}
\subfigure[\label{fig:vent_view_0} Flying into the entrance.]
{\includegraphics[width=0.48\columnwidth]{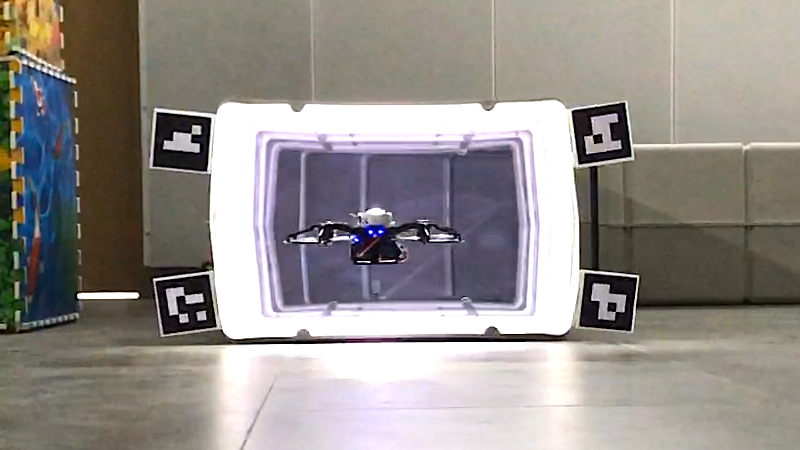}}
\subfigure[\label{fig:vent_view_1} Flying through turn 2.]
{\includegraphics[width=0.48\columnwidth]{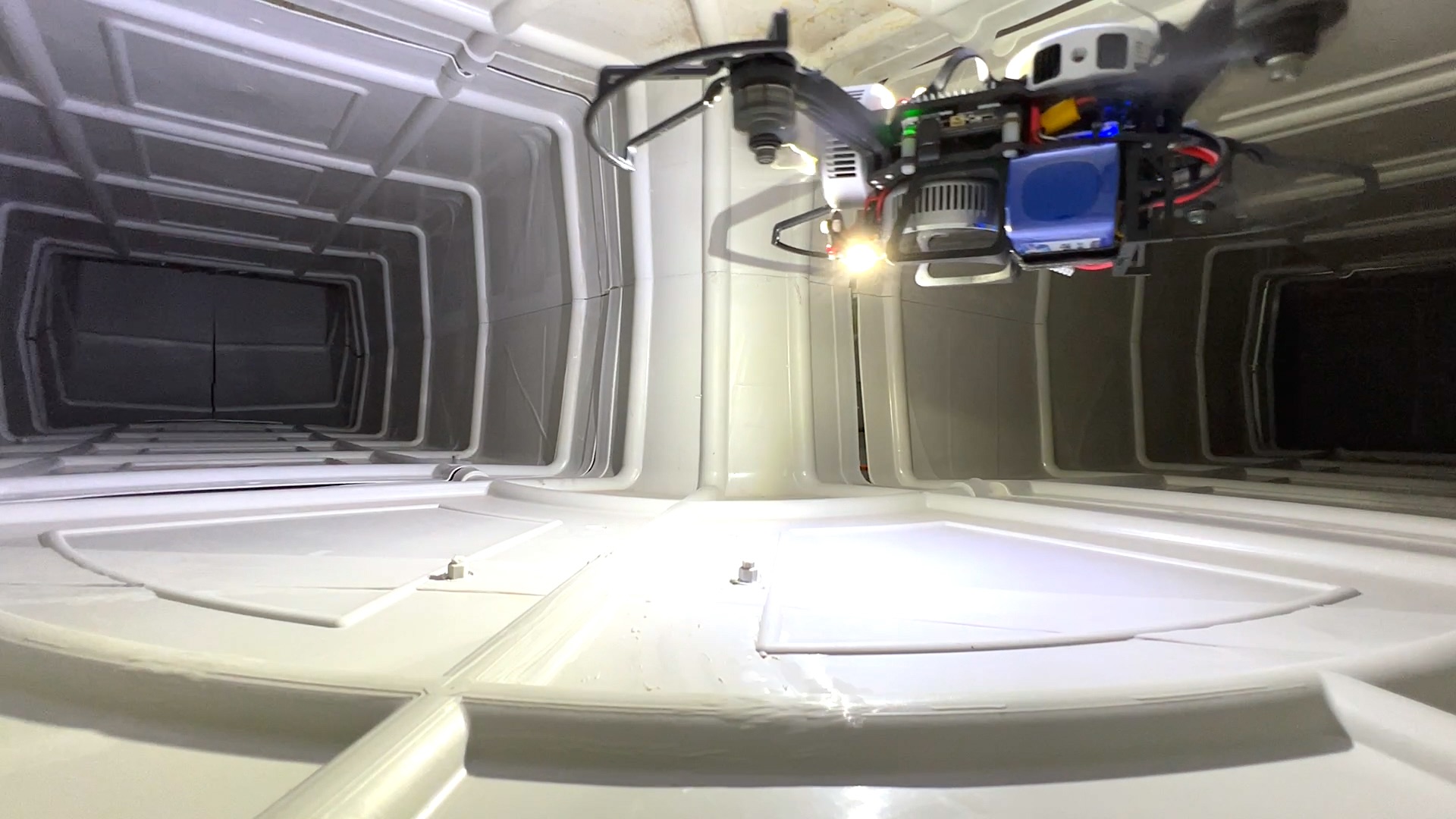}}
\subfigure[\label{fig:vent_view_2} Flying up the first vertical section.]
{\includegraphics[width=0.48\columnwidth]{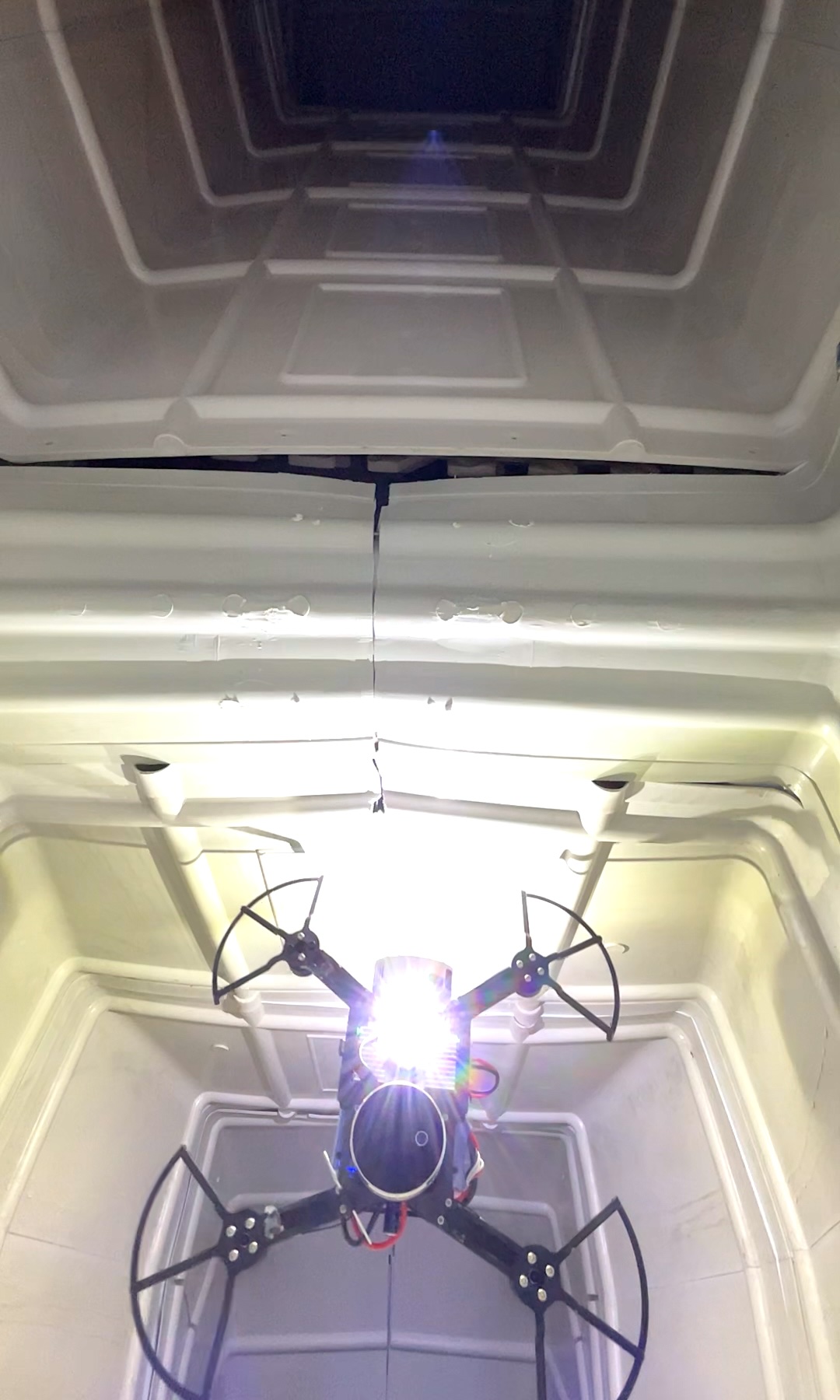}}
\subfigure[\label{fig:vent_view_3} Flying down the second vertical section.]
{\includegraphics[width=0.48\columnwidth]{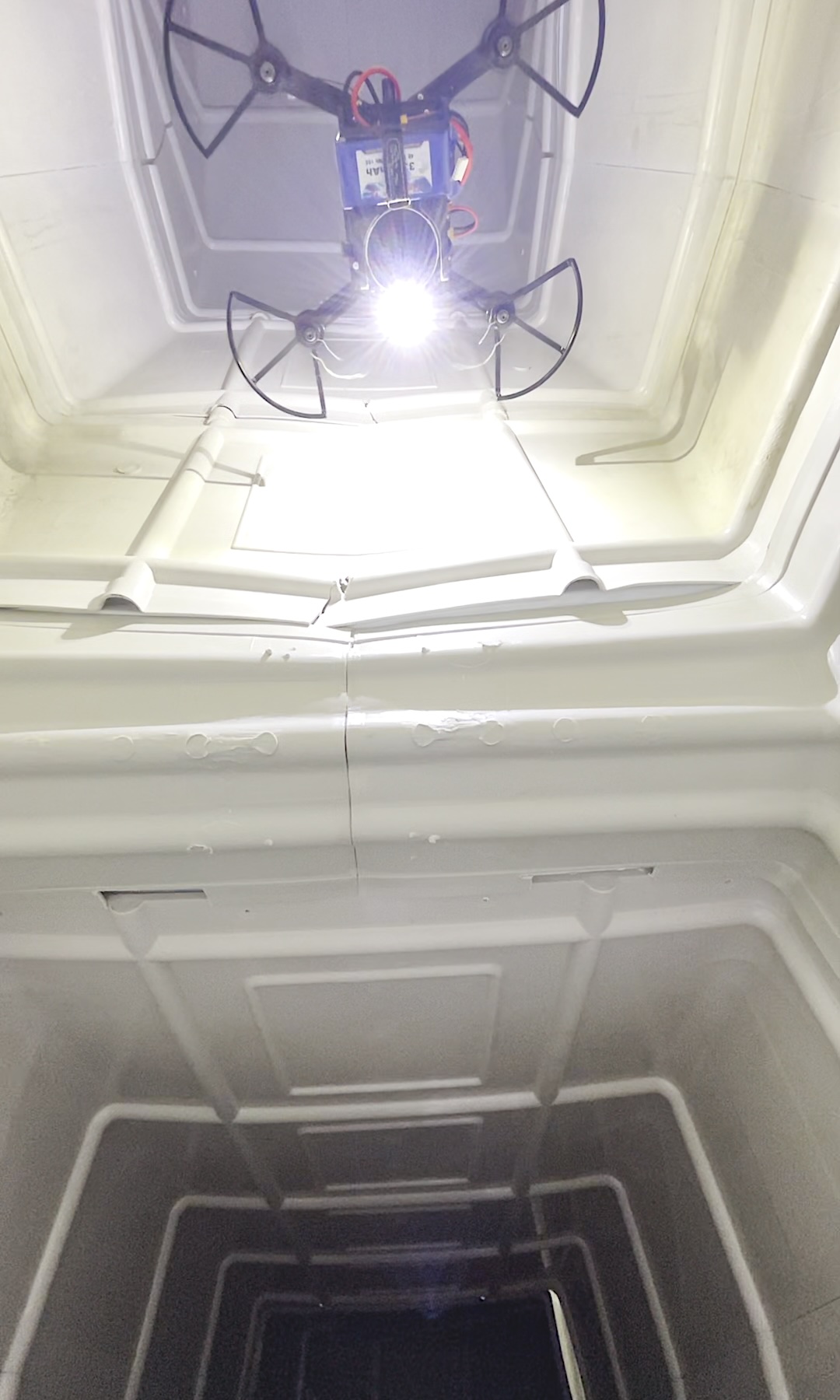}}
\subfigure[\label{fig:vent_view_4} Flying through the last turn.]
{\includegraphics[width=0.48\columnwidth]{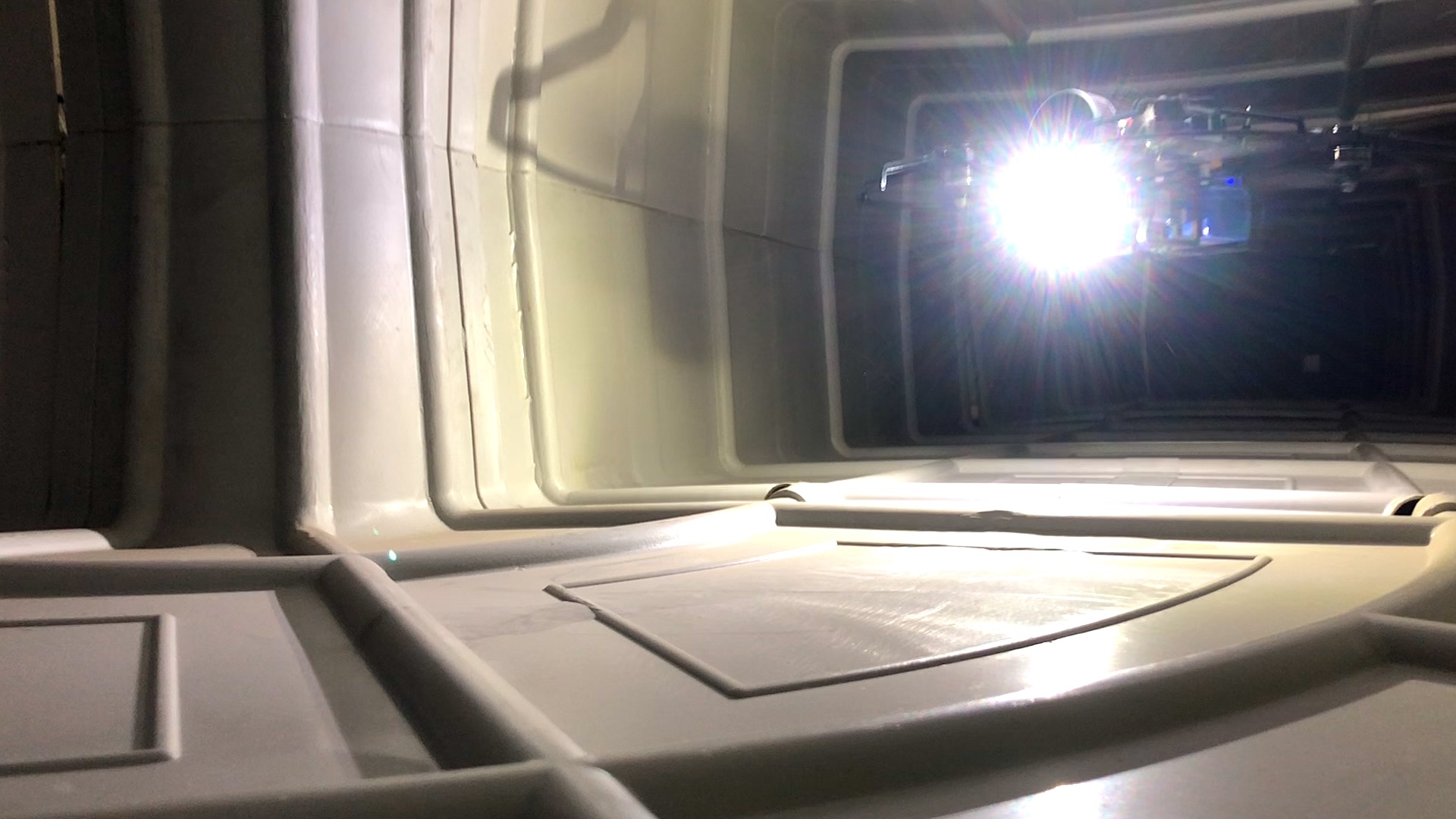}}
\subfigure[\label{fig:vent_view_5} Flying out of the exit.]
{\includegraphics[width=0.48\columnwidth]{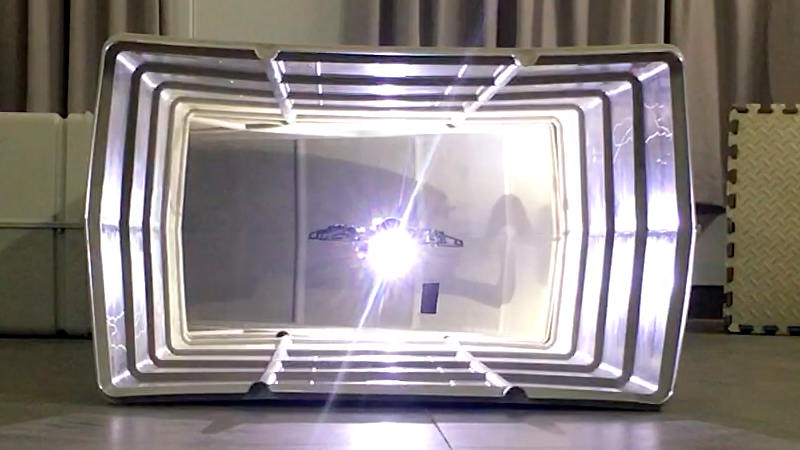}}
\end{center}
\vspace{-0.5cm}
\caption{\label{fig:vent_view}Snapshots and visualization results when the quadrotor flies through a rigid vent pipe shown in Fig.~\ref{fig:vent}. The markers are same as previous figures and the quadrotor positions in the snapshots are labelled in the visualization.}
\vspace{-0.9cm}
\end{figure}

Previous experiments are predominantly conducted in flexible vent pipes. To validate the generality of the system, we also conduct experiments in a rigid vent pipe with rectangular cross-sections, ranging from 0.5 m to 0.8 m, and featuring right angle sharp turns, as shown in Fig.~\ref{fig:vent}. The proposed system takes 66 s to successively traverse this narrow vent pipe. The snapshots and visualization results during the flight are shown in Fig.~\ref{fig:vent_view}. To our knowledge, no previous work has demonstrated the capability to navigate such a narrow vent pipe. This is the first system capable of being adopted in such a 3-D narrow tunnel. This experiment showcases the system's exceptional robustness and capability.

\subsection{Vent Pipe on a Construction Site}
\label{subsec:construction_site_result}

\begin{figure}[t]
\begin{center}
\subfigure[\label{fig:tunnel_ssl_rviz} Visualization.]
{\includegraphics[height=0.42\columnwidth]{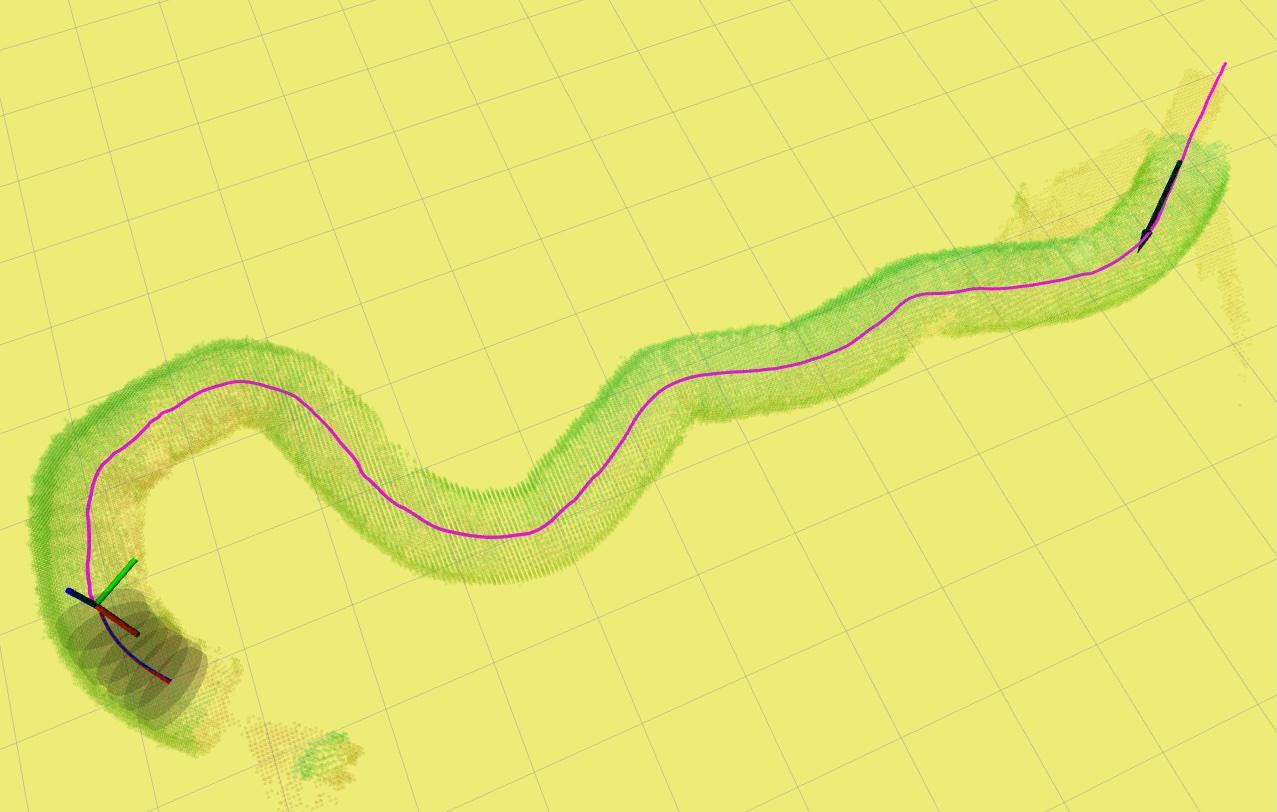}}       
\subfigure[\label{fig:tunnel_ssl_fpv} Onboard camera view during the flight.]
{\includegraphics[height=0.42\columnwidth]{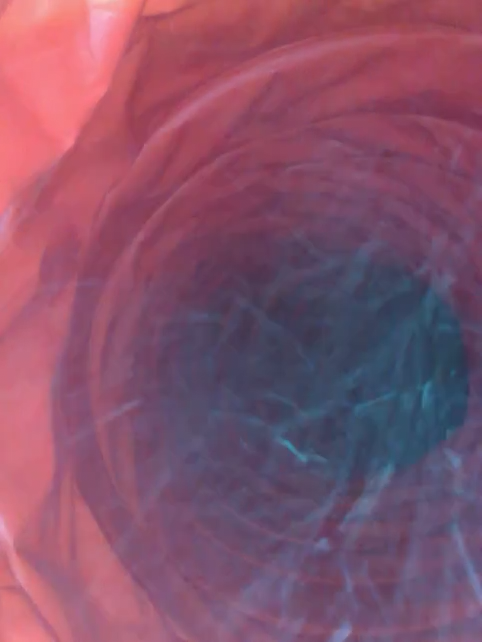}}
\subfigure[\label{fig:tunnel_ssl_view_0} Flying into the entrance.]
{\includegraphics[width=0.95\columnwidth]{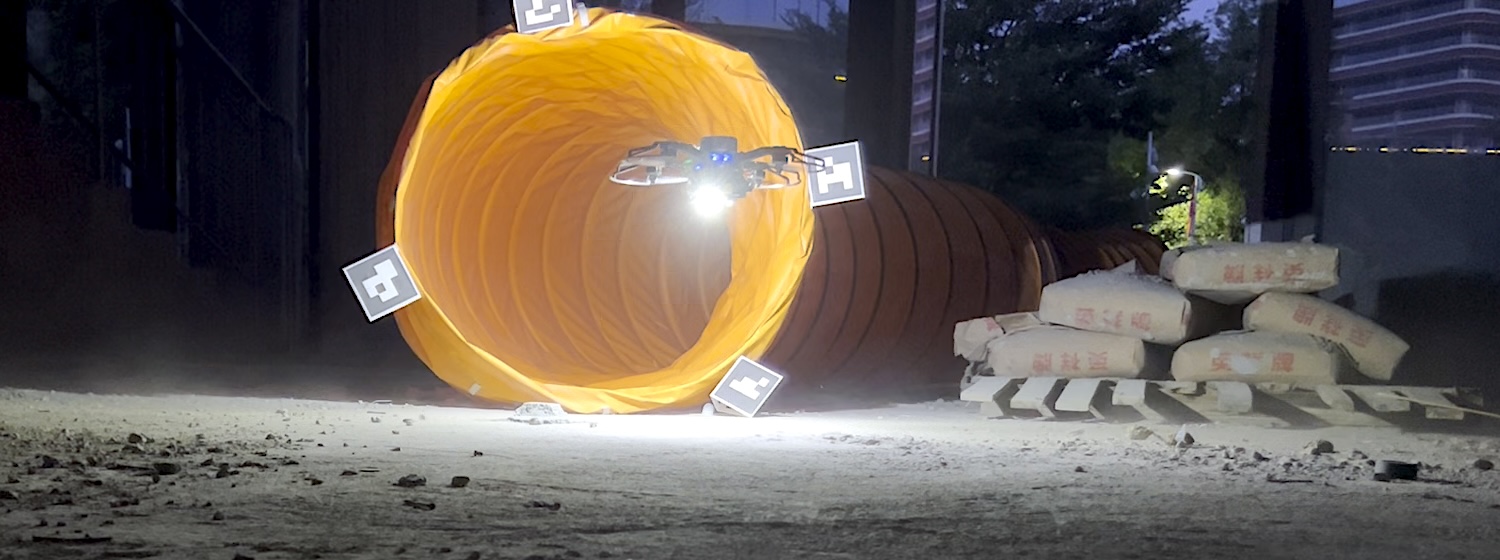}}
\subfigure[\label{fig:tunnel_ssl_view_1} Flying out of the exit.]
{\includegraphics[width=0.95\columnwidth]{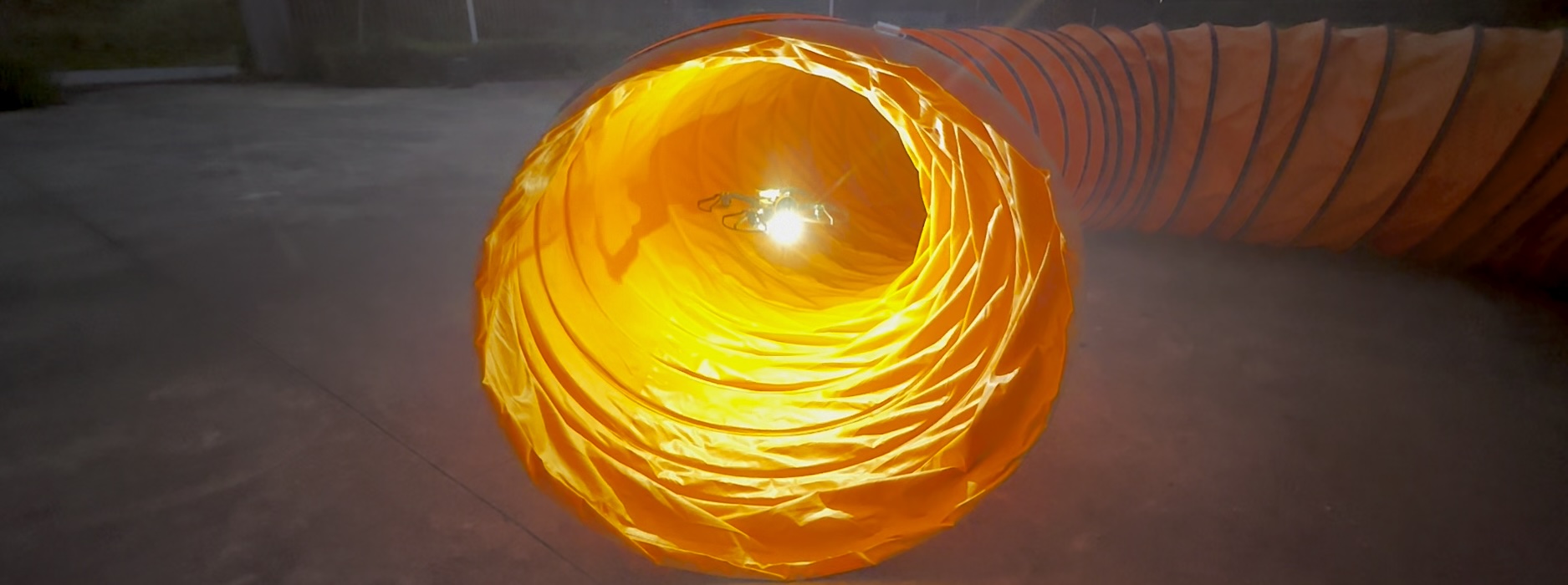}}
\end{center}
\vspace{-0.5cm}
\caption{\label{fig:tunnel_ssl_view}Snapshots and visualization results when the quadrotor flies through a vent pipe on a construction site shown in Fig.~\ref{fig:tunnel_ssl}. The markers are same as previous figures and the quadrotor positions in the snapshots are labelled in the visualization.}
%\vspace{-0.7cm}
\end{figure}

To further validate the system in a real-world setting, we also conduct flight experiments in a vent pipe on an active construction site, transitioning from an indoor area to an outdoor area, as shown in Fig.~\ref{fig:tunnel_ssl}. The vent pipe is 20 m long, with a circular cross-section ranging from 0.7 to 0.8 m in diameter. The visualization result and snapshots during flights are shown in Fig.~\ref{fig:tunnel_ssl_view}. Despite of the strong dust on the site and in the pipe that can be clearly seen in Fig.~\ref{fig:tunnel_ssl_view_0} and~\ref{fig:tunnel_ssl_fpv}, the proposed autonomous quadrotor system is still able to traverse the pipe smoothly from indoor to outdoor. This experiment on the construction site demonstrates the adaptivity and robustness of the system and its potential for industrial applications in the future.

\subsection{Summary on Traversalibilities of Different Systems}
\label{subsec:comparison_results}

\begin{table}[t]
\centering
\caption{\label{tab:system_benchmark}Comparison on Traversalibilities of Different Systems in 2-D and 3-D tunnels}
\begin{tabular}{@{}lccc@{}}
\toprule
System & 2-D tunnel case 2 & 2-D tunnel case 3  & 3-D tunnel case 2 \\
\midrule
Proposed  &  $\circ$ & $\circ$ & $\circ$\\
\cite{wang2022neither} & $\times$ & $\times$ & $-$\\
Manual & $\times$ & $\times$ & $\times$ \\
\bottomrule
\multicolumn{4}{l}{$\circ$: success in all trails. $\times$: fail in all trails. $-$: not supported.}
\end{tabular}
%\vspace{-0.4cm}
\end{table}

As indicated in \cite{wang2022neither}, state-of-the-art generalized motion planners not designed for navigating narrow tunnels, for instance \cite{zhou2019robust}, are not able to navigate the quadrotor through a narrow tunnel. Therefore, comparisons with them are not meaningful. Moreover, it is observed that during manual flights, the quadrotor tends to veer towards the tunnel walls and difficult to recover, even when just centimeters away from the centerline. This suggests that planners not following the centerline of narrow tunnels~\cite{arrizabalaga2022towards} or systems with large control tracking errors~\cite{elmokadem2021method} are not capable of navigating narrow tunnels effectively.  Therefore, the comparisons with those planners or systems also hold little significance.

%As indicated in ~\cite{wang2022neither}, state-of-the-art generalized motion planners are normally applicable for more general outdoor environments, and the narrow tunnel scenarios are not well investigated. For instance, ~\cite{zhou2019robust} focuses on the robustness of flight controller in a .... In regard of tunnel flight systems, the planner in ~\cite{arrizabalaga2022towards} does not follow the centerline of narrow tunnels, and ~\cite{elmokadem2021method} involves large control tracking errors, which make them not capable for narrow tunnel flights. During our manual-pilot experiments, the quadrotor is immediately sucked to the tunnel wall even when it is just several centimeters away from the centerline, and difficult to recover. Thus, the aforementioned extreme challenges in such an special environment make it barely possible for the current planners to navigate the quadrotor through a narrow tunnel, which outstands our contribution.

As a result, we compare the proposed system with others that are intuitively capable of navigating narrow tunnels in more general scenarios. In summary, we compared the proposed system with the system in \cite{wang2022neither} and manual flights by an experienced pilot in 2-D tunnel case 2, case 3 and 3-D tunnel case 2. Five flights are conducted for each configuration, except that additional manual flights are performed in certain cases. As shown in the traversal results in Tab. \ref{tab:system_benchmark}, the proposed system successfully navigates all scenarios, whereas the system in \cite{wang2022neither} and manual flights fail to traverse any of the tunnels. The system in \cite{wang2022neither} always fails in 2-D tunnel case 2 and 3 due to its inability to manage sharp turns and variations in cross-sectional shapes and sizes. Additionally, it lacks support for 3-D tunnels because of inadequate perception capabilities in upward and downward directions. Manual flights also struggle with these tunnels, no matter the pilot is using a conventional controller or a motion controller, as maintaining the drone along the tunnel centerline is extremely challenging. Any slight deviation results in the drone being sucked into the tunnel wall, leaving almost no chance for recovery in such narrow spaces. For the 2-D tunnel case 3, manual flight cannot even make to the first turn in all ten trails. These outcomes highlight the superior performance and capability of the proposed system over the system in \cite{wang2022neither} and experienced human pilots.

\section{Extendability of the System and Discussion}
\label{sec:extendability}

In the experiments and results, the proposed autonomous aerial system demonstrates exceptional robustness in navigating a wide range of narrow tunnels with both circular and rectangular cross-sections. The proposed system can be extended to other tunnel-like environments with various cross-sectional shapes. Additionally, both the complete system framework and individual modules, including perception, planning, and control, can be transferred to multirotor platforms of different sizes and purposes.

We can also outline the following pipeline to adapt the proposed method for different multirotors and expand its application to tunnels with various cross-sectional shapes:
\begin{enumerate}
\item Construct CFD analysis cases according to the multirotor configuration and specification, as well as the possible cross-section shapes of the tunnel to collect the raw ego airflow disturbance data. Then, retrain the MLP disturbance factor model based on the GRNN results derived from the raw data.

\item Configure the perception factor model according to the configuration and parameters of the cameras on the multirotor.

\item Generate images of all possible shapes of the tunnel cross-sections and retrain the classification CNN based on these images. Use the generalized Hough transform to perform cross-section pattern detection for tunnels of general shapes.

\item Deploy the proposed planning module, which includes the updated cross-section shape classification and detection, along with the revised ego airflow disturbance and perception factor models. Then, integrate these with the perception and control modules on the multirotor.
\end{enumerate}

In the current setup, we successfully use a quadrotor with a 40 cm diameter to navigate tunnels with 50 cm diameter cross-sections, achieving an cross-section to quadrotor diameter ratio of 1.25. It is also foreseeable that a quadrotor is able to navigate a tunnel with larger cross-section to quadrotor diameter ratio and lesser ego airflow disturbance. While the diameter ratio is evident for a specific quadrotor and tunnel, analyzing disturbances requires either CFD simulations or real-world tests, which are complex. Cases not meeting these criteria also need further analysis. Developing explicit traversability criteria could be explored in future work.

Additionally, the corner shape can also affect the traversability. In the current setting, the quadrotor is able to navigate tunnels with right angles, even U-turns with radius as small as 0.3 m. Except for the physical size constraints, the major constraint is on the FoV of the perception. While the current module with yaw movements and a simple 3-camera system achieves omni-directional capability, a true omni-directional perception system offers a stronger FoV. This system eliminates the need for yaw movement during flights, resulting in reduced optical flow speed and potentially enhanced perception performance. It is worth mentioning that a physical omni-directional setup may increase the payload for the multirotor system, posing significant challenges for hardware design and requiring more computational resources, which could compromise traversability and flexibility.

\section{Conclusion}
\label{sec:conclusion}

In this paper, we proposed a comprehensive autonomous aerial system designed for navigating narrow tunnels. Extensive flight experiments in various realistic and challenging narrow tunnels validated the robustness of the entire system. Furthermore, a pipeline for deploying the system on other multirotor platforms is provided, facilitating future developments. The proposed system is the first autonomous aerial system capable of smoothly navigating tunnels in arbitrary directions, with sharp corners, and as narrow as 0.5 m in diameter, outperforming experienced human pilots. This work significantly extends the capabilities of MAVs and offers guidelines for future aerial applications in narrow tunnels.

Despite the superior performance of the system, there are still potential improvements in the future:

\begin{enumerate}

\item Currently, the system is limited to narrow tunnels without obstacles. We plan to enhance the planning framework to handle obstacles inside tunnels, thereby improving safety and versatility, in future work.

\item The current system can only navigate tunnels without forks. A more advanced planning framework will be introduced in the future to deal with forks in tunnels. This enhancement can also incorporate exploration planning algorithms, which can further improve the traversal and
mapping efficiency.

\item The proposed system comprises a single MAV. There is potential to scale this system by incorporating multiple MAVs and considering the interactions between individual drones in terms of perception and airflow disturbances, thereby enhancing the system’s efficiency.
\end{enumerate}
These enhancements are aimed at broadening the application scope of the system, including uses in geological prospecting, industrial pipe inspection, search and rescue operations in mine tunnels, etc.. These future developments will leverage the capabilities of MAVs to operate in environments that are typically challenging or hazardous for humans.

\newlength{\bibitemsep}\setlength{\bibitemsep}{.0238\baselineskip}
\newlength{\bibparskip}\setlength{\bibparskip}{0pt}
\let\oldthebibliography\thebibliography
\renewcommand\thebibliography[1]{%
  \oldthebibliography{#1}%
  \setlength{\parskip}{\bibitemsep}%
  \setlength{\itemsep}{\bibparskip}%
}

%\clearpage
\bibliography{TRO2024_Luqi} 

\begin{thebibliography}{10}
\providecommand{\url}[1]{#1}
\csname url@rmstyle\endcsname
\providecommand{\newblock}{\relax}
\providecommand{\bibinfo}[2]{#2}
\providecommand\BIBentrySTDinterwordspacing{\spaceskip=0pt\relax}
\providecommand\BIBentryALTinterwordstretchfactor{4}
\providecommand\BIBentryALTinterwordspacing{\spaceskip=\fontdimen2\font plus
\BIBentryALTinterwordstretchfactor\fontdimen3\font minus
  \fontdimen4\font\relax}
\providecommand\BIBforeignlanguage[2]{{%
\expandafter\ifx\csname l@#1\endcsname\relax
\typeout{** WARNING: IEEEtran.bst: No hyphenation pattern has been}%
\typeout{** loaded for the language `#1'. Using the pattern for}%
\typeout{** the default language instead.}%
\else
\language=\csname l@#1\endcsname
\fi
#2}}

\bibitem{mathe2016vision}
K.~M{\'a}th{\'e}, L.~Bu{\c{s}}oniu, L.~Barab{\'a}s, C.-I. Iuga, L.~Miclea, and
  J.~Braband, ``Vision-based control of a quadrotor for an object inspection
  scenario,'' in \emph{Proc. Int. Conf. Unmanned Aircr. Syst. ({ICUAS})}.\hskip
  1em plus 0.5em minus 0.4em\relax IEEE, 2016, pp. 849--857.

\bibitem{luqi2018collaborative}
L.~Wang, D.~Cheng, F.~Gao, F.~Cai, J.~Guo, M.~Lin, and S.~Shen, ``A
  collaborative aerial-ground robotic system for fast exploration,'' in
  \emph{Proc. Int. Symp. Exp. Robot. ({ISER})}, 2018, pp. 59--71.

\bibitem{manyam2017surveillance}
S.~G. {Manyam}, S.~{Rasmussen}, D.~W. {Casbeer}, K.~{Kalyanam}, and
  S.~{Manickam}, ``Multi-uav routing for persistent intelligence surveillance
  reconnaissance missions,'' in \emph{Proc. Int. Conf. Unmanned Aircr. Syst.
  ({ICUAS})}, 2017, pp. 573--580.

\bibitem{ozaslan2017autonomous}
T.~{\"O}zaslan, G.~Loianno, J.~Keller, C.~J. Taylor, V.~Kumar, J.~M.
  Wozencraft, and T.~Hood, ``Autonomous navigation and mapping for inspection
  of penstocks and tunnels with mavs,'' \emph{{IEEE} Robot. Automat. Lett.
  ({RA-L})}, vol.~2, no.~3, pp. 1740--1747, 2017.

\bibitem{vong2019integral}
C.~H. Vong, K.~Ryan, and H.~Chung, ``Integral backstepping position control for
  quadrotors in tunnel-like confined environments,'' in \emph{Proc. {IEEE} Int.
  Conf. on Robot. Autom. ({ICRA})}.\hskip 1em plus 0.5em minus 0.4em\relax
  IEEE, 2019, pp. 6425--6431.

\bibitem{wang2022neither}
L.~Wang, H.~Xu, Y.~Zhang, and S.~Shen, ``Neither fast nor slow: How to fly
  through narrow tunnels,'' \emph{{IEEE} Robot. Automat. Lett. ({RA-L})},
  vol.~7, no.~2, pp. 5489--5496, 2022.

\bibitem{zhou2019robust}
B.~Zhou, F.~Gao, L.~Wang, C.~Liu, and S.~Shen, ``Robust and efficient quadrotor
  trajectory generation for fast autonomous flight,'' \emph{{IEEE} Robot.
  Automat. Lett. ({RA-L})}, vol.~4, no.~4, pp. 3529--3536, 2019.

\bibitem{betz1937ground}
A.~Betz, ``The ground effect on lifting propellers,'' 1937.

\bibitem{cheeseman1955effect}
I.~Cheeseman and W.~Bennett, ``The effect of ground on a helicopter rotor in
  forward flight,'' Aeroplane and Armament Experimental Establishment, Ministry
  of Supply, U.K., Aeronautical Research Council Reports and Memoranda No.
  3021, Sep. 1955.

\bibitem{eberhart2017modeling}
\BIBentryALTinterwordspacing
G.~M. Eberhart, ``Modeling of ground effect benefits for multi-rotor small
  unmanned aerial systems at hover,'' Master's thesis, Dept. Mech. Eng., Ohio
  Univ., 2017. [Online]. Available:
  \url{http://rave.ohiolink.edu/etdc/view?acc_num=ohiou1502802483367365}
\BIBentrySTDinterwordspacing

\bibitem{conyers2019empirical}
\BIBentryALTinterwordspacing
S.~A. Conyers, ``Empirical evaluation of ground, ceiling, and wall effect for
  small-scale rotorcraft,'' Ph.D. dissertation, Dept. Elect. \& Comp. Eng.,
  Univ. of Denver, 2019. [Online]. Available:
  \url{https://digitalcommons.du.edu/etd/1570}
\BIBentrySTDinterwordspacing

\bibitem{kan2019analysis}
X.~Kan, J.~Thomas, H.~Teng, H.~G. Tanner, V.~Kumar, and K.~Karydis, ``Analysis
  of ground effect for small-scale uavs in forward flight,'' \emph{{IEEE}
  Robot. Automat. Lett. ({RA-L})}, vol.~4, no.~4, pp. 3860--3867, 2019.

\bibitem{britcher2021use}
V.~Britcher and S.~Bergbreiter, ``Use of a mems differential pressure sensor to
  detect ground, ceiling, and walls on small quadrotors,'' \emph{{IEEE} Robot.
  Automat. Lett. ({RA-L})}, vol.~6, no.~3, pp. 4568--4575, 2021.

\bibitem{powers2013influence}
C.~Powers, D.~Mellinger, A.~Kushleyev, B.~Kothmann, and V.~Kumar, ``Influence
  of aerodynamics and proximity effects in quadrotor flight,'' in \emph{Proc.
  Int. Symp. Exp. Robot. ({ISER})}.\hskip 1em plus 0.5em minus 0.4em\relax
  Springer, 2013, pp. 289--302.

\bibitem{danjun2015autonomous}
L.~Danjun, Z.~Yan, S.~Zongying, and L.~Geng, ``Autonomous landing of quadrotor
  based on ground effect modelling,'' in \emph{2015 34th Chinese Control
  Conference (CCC)}.\hskip 1em plus 0.5em minus 0.4em\relax IEEE, 2015, pp.
  5647--5652.

\bibitem{johnson2012helicopter}
W.~Johnson, \emph{Helicopter theory}.\hskip 1em plus 0.5em minus 0.4em\relax
  Courier Corporation, 2012.

\bibitem{nishio2020stable}
T.~Nishio, M.~Zhao, F.~Shi, T.~Anzai, K.~Kawaharazuka, K.~Okada, and M.~Inaba,
  ``Stable control in climbing and descending flight under upper walls using
  ceiling effect model based on aerodynamics,'' in \emph{Proc. {IEEE} Int.
  Conf. on Robot. Autom. ({ICRA})}.\hskip 1em plus 0.5em minus 0.4em\relax
  IEEE, 2020, pp. 172--178.

\bibitem{robinson2014computational}
D.~C. Robinson, H.~Chung, and K.~Ryan, ``Computational investigation of micro
  rotorcraft near-wall hovering aerodynamics,'' in \emph{Proc. Int. Conf.
  Unmanned Aircr. Syst. ({ICUAS})}.\hskip 1em plus 0.5em minus 0.4em\relax
  IEEE, 2014, pp. 1055--1063.

\bibitem{wang2021estimation}
L.~Wang, B.~Zhou, C.~Liu, and S.~Shen, ``Estimation and adaption of indoor ego
  airflow disturbance with application to quadrotor trajectory planning,'' in
  \emph{Proc. {IEEE} Int. Conf. on Robot. Autom. ({ICRA})}.\hskip 1em plus
  0.5em minus 0.4em\relax IEEE, 2021, pp. 384--390.

\bibitem{mueggler2014event}
E.~Mueggler, B.~Huber, and D.~Scaramuzza, ``Event-based, 6-dof pose tracking
  for high-speed maneuvers,'' in \emph{Proc. {IEEE/RSJ} Int. Conf. Intell.
  Robots Syst.({IROS})}.\hskip 1em plus 0.5em minus 0.4em\relax IEEE, 2014, pp.
  2761--2768.

\bibitem{shen2013vision}
S.~Shen, Y.~Mulgaonkar, N.~Michael, and V.~Kumar, ``Vision-based state
  estimation and trajectory control towards high-speed flight with a
  quadrotor.'' in \emph{Proc. Robot. Sci. Syst. ({RSS})}, vol.~1.\hskip 1em
  plus 0.5em minus 0.4em\relax Citeseer, 2013, p.~32.

\bibitem{morrell2018comparison}
B.~Morrell, R.~Thakker, G.~Merewether, R.~Reid, M.~Rigter, T.~Tzanetos, and
  G.~Chamitoff, ``Comparison of trajectory optimization algorithms for
  high-speed quadrotor flight near obstacles,'' \emph{{IEEE} Robot. Automat.
  Lett. ({RA-L})}, vol.~3, no.~4, pp. 4399--4406, 2018.

\bibitem{fridovich2018planning}
D.~Fridovich-Keil, S.~L. Herbert, J.~F. Fisac, S.~Deglurkar, and C.~J. Tomlin,
  ``Planning, fast and slow: A framework for adaptive real-time safe trajectory
  planning,'' in \emph{Proc. {IEEE} Int. Conf. on Robot. Autom.
  ({ICRA})}.\hskip 1em plus 0.5em minus 0.4em\relax IEEE, 2018, pp. 387--394.

\bibitem{quan2021eva}
L.~Quan, Z.~Zhang, X.~Zhong, C.~Xu, and F.~Gao, ``Eva-planner: Environmental
  adaptive quadrotor planning,'' in \emph{Proc. {IEEE} Int. Conf. on Robot.
  Autom. ({ICRA})}.\hskip 1em plus 0.5em minus 0.4em\relax IEEE, 2021, pp.
  398--404.

\bibitem{chen2016online}
J.~Chen, T.~Liu, and S.~Shen, ``Online generation of collision-free
  trajectories for quadrotor flight in unknown cluttered environments,'' in
  \emph{Proc. {IEEE} Int. Conf. on Robot. Autom. ({ICRA})}.\hskip 1em plus
  0.5em minus 0.4em\relax IEEE, 2016, pp. 1476--1483.

\bibitem{loianno2016estimation}
G.~Loianno, C.~Brunner, G.~McGrath, and V.~Kumar, ``Estimation, control, and
  planning for aggressive flight with a small quadrotor with a single camera
  and imu,'' \emph{{IEEE} Robot. Automat. Lett. ({RA-L})}, vol.~2, no.~2, pp.
  404--411, 2016.

\bibitem{gao2020teach}
F.~Gao, L.~Wang, B.~Zhou, X.~Zhou, J.~Pan, and S.~Shen, ``Teach-repeat-replan:
  A complete and robust system for aggressive flight in complex environments,''
  \emph{{IEEE} Trans. Robot.}, vol.~36, no.~5, pp. 1526--1545, 2020.

\bibitem{petrlik2020robust}
M.~Petrl{\'\i}k, T.~B{\'a}{\v{c}}a, D.~He{\v{r}}t, M.~Vrba, T.~Krajn{\'\i}k,
  and M.~Saska, ``A robust uav system for operations in a constrained
  environment,'' \emph{{IEEE} Robot. Automat. Lett. ({RA-L})}, vol.~5, no.~2,
  pp. 2169--2176, 2020.

\bibitem{arrizabalaga2022towards}
J.~Arrizabalaga and M.~Ryll, ``Towards time-optimal tunnel-following for
  quadrotors,'' in \emph{Proc. {IEEE} Int. Conf. on Robot. Autom.
  ({ICRA})}.\hskip 1em plus 0.5em minus 0.4em\relax IEEE, 2022, pp. 4044--4050.

\bibitem{elmokadem2021method}
T.~Elmokadem and A.~V. Savkin, ``A method for autonomous collision-free
  navigation of a quadrotor uav in unknown tunnel-like environments,''
  \emph{Robotica}, 2021, first view. doi: 10.1017/S0263574721000849.

\bibitem{qin2018vins}
T.~Qin, P.~Li, and S.~Shen, ``Vins-mono: A robust and versatile monocular
  visual-inertial state estimator,'' \emph{{IEEE} Trans. Robot.}, vol.~34,
  no.~4, pp. 1004--1020, 2018.

\bibitem{qin2019a}
T.~Qin, J.~Pan, S.~Cao, and S.~Shen, ``A general optimization-based framework
  for local odometry estimation with multiple sensors,'' \emph{arXiv preprint
  arXiv:1901.03638}, 2019.

\bibitem{wang2024vinsmulti}
L.~Wang, Y.~Xu, and S.~Shen, ``Vins-multi: A robust asynchronous
  multi-camera-imu state estimator,'' \emph{arXiv preprint arXiv:2405.14539},
  2024.

\bibitem{han2019fiesta}
L.~Han, F.~Gao, B.~Zhou, and S.~Shen, ``Fiesta: Fast incremental euclidean
  distance fields for online motion planning of aerial robots,'' in \emph{Proc.
  {IEEE/RSJ} Int. Conf. Intell. Robots Syst.({IROS})}.\hskip 1em plus 0.5em
  minus 0.4em\relax IEEE, 2019, pp. 4423--4430.

\bibitem{specht1991general}
D.~Specht, ``A general regression neural network,'' \emph{IEEE Trans. Neural
  Netw.}, vol.~2, no.~6, pp. 568--576, 1991.

\bibitem{quinlan1993elastic}
S.~Quinlan and O.~Khatib, ``Elastic bands: Connecting path planning and
  control,'' in \emph{Proc. {IEEE} Int. Conf. on Robot. Autom. ({ICRA})}.\hskip
  1em plus 0.5em minus 0.4em\relax IEEE, 1993, pp. 802--807.

\bibitem{zhu2015convex}
Z.~Zhu, E.~Schmerling, and M.~Pavone, ``A convex optimization approach to
  smooth trajectories for motion planning with car-like robots,'' in
  \emph{Proc. {IEEE} Conf. Decis. Control ({CDC})}.\hskip 1em plus 0.5em minus
  0.4em\relax IEEE, 2015, pp. 835--842.

\bibitem{jung2004rectangle}
C.~R. Jung and R.~Schramm, ``Rectangle detection based on a windowed hough
  transform,'' in \emph{Proceedings. 17th Brazilian Symposium on Computer
  Graphics and Image Processing}.\hskip 1em plus 0.5em minus 0.4em\relax IEEE,
  2004, pp. 113--120.

\bibitem{dolgov2010path}
D.~Dolgov, S.~Thrun, M.~Montemerlo, and J.~Diebel, ``Path planning for
  autonomous vehicles in unknown semi-structured environments,'' \emph{Int. J.
  Robot. Research ({IJRR})}, vol.~29, no.~5, pp. 485--501, 2010.

\bibitem{mueller2015computationally}
M.~W. Mueller, M.~Hehn, and R.~D'Andrea, ``A computationally efficient motion
  primitive for quadrocopter trajectory generation,'' \emph{{IEEE} Trans.
  Robot.}, vol.~31, no.~6, pp. 1294--1310, 2015.

\end{thebibliography}

\begin{minipage}{1.0\columnwidth}
\begin{wrapfigure}{l}{0.3\columnwidth}
\begin{center}
\includegraphics[width=0.3\columnwidth]{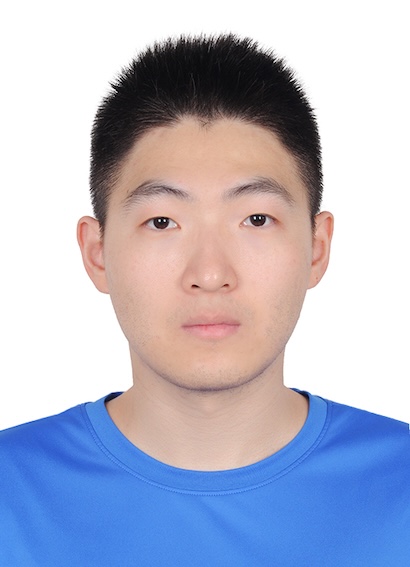}
\end{center}
\end{wrapfigure}
Luqi Wang received a B.Eng. degree in Computer Engineering and Aerospace Engineering and a Ph.D. degree in Electronic and Computer Engineering, both from the Hong Kong University of Science and Technology (HKUST), Hong Kong, China, in 2018 and 2024, respectively. He is currently a Research Associate in the Department of Electronic and Computer Engineering at HKUST. His research interests include control, perception, navigation, and path planning for autonomous robots.\\
\end{minipage}

\begin{minipage}{1.0\columnwidth}
\begin{wrapfigure}{l}{0.3\columnwidth}
\begin{center}
\includegraphics[width=0.3\columnwidth]{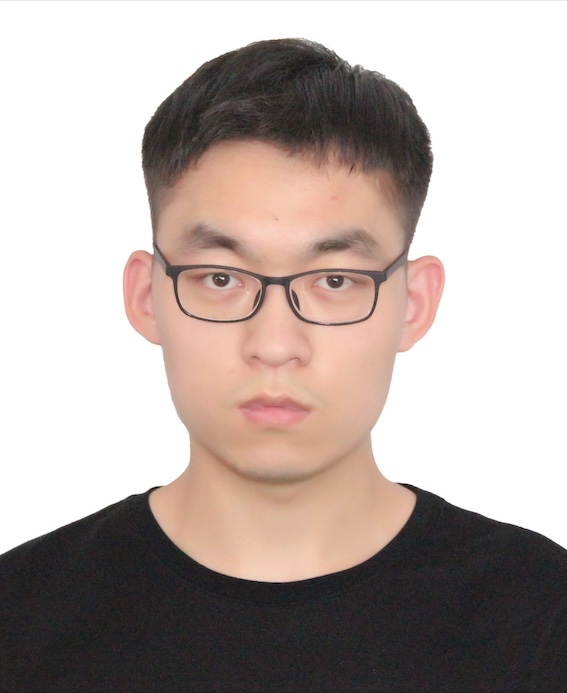}
\end{center}
\end{wrapfigure}
Yan Ning received a B.Eng. degree in Mechanical and Aerospace Engineering and an M.Phil. degree in Electronic and Computer Engineering, both from the Hong Kong University of Science and Technology (HKUST), Hong Kong, China, in 2018 and 2023, respectively. He is currently pursuing a Ph.D. degree in the Department of Electronic and Computer Engineering at HKUST. His research interests include hardware design, model-based and learning-based locomotion control, and planning for legged robots.\\
\end{minipage}

\begin{minipage}{1.0\columnwidth}
\begin{wrapfigure}{l}{0.3\columnwidth}
\begin{center}
\includegraphics[width=0.3\columnwidth]{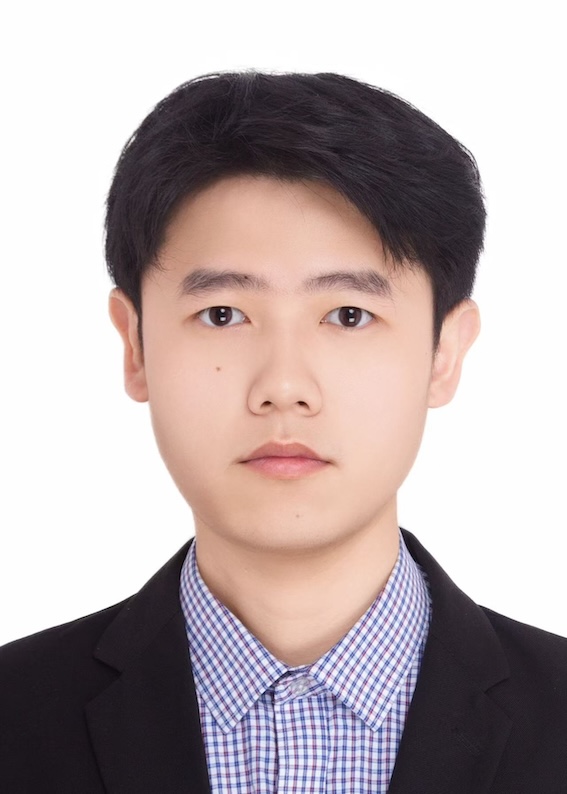}
\end{center}
\end{wrapfigure}
Hongming Chen received a B.Eng. degree in Software Engineering from the University of Electronic Science and Technology of China, Chengdu, China, in 2024. He is currently pursuing an M.Phil. degree in the Department of Intelligent Systems Engineering at Sun Yat-sen University, Shenzhen, China, under the supervision of Prof. Ximin Lyu. His research interests include unmanned aerial vehicles, semantic navigation, and aerial manipulation.\\
\end{minipage}

\begin{minipage}{1.0\columnwidth}
\begin{wrapfigure}{l}{0.3\columnwidth}
\begin{center}
\includegraphics[width=0.3\columnwidth]{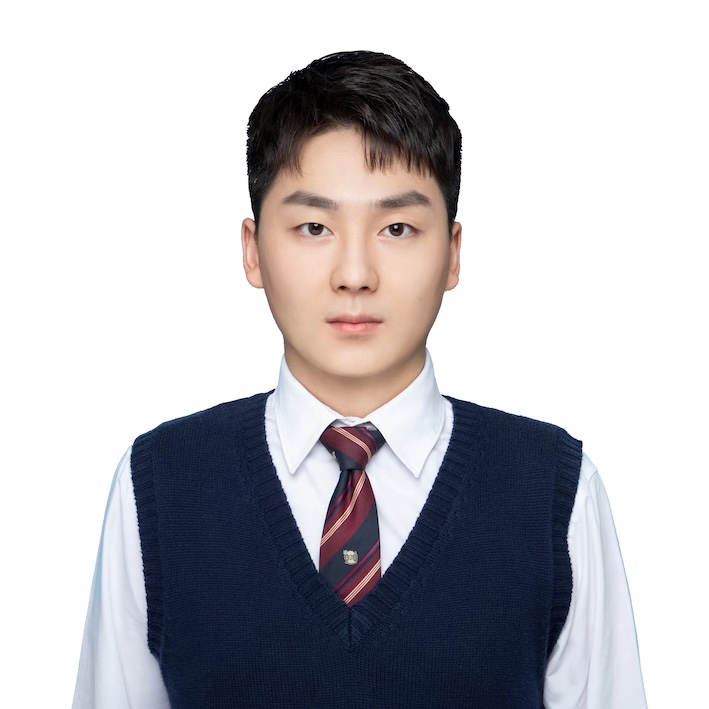}
\end{center}
\end{wrapfigure}
Peize Liu received a B.Eng. degree in Software Engineering from the University of Electronic Science and Technology of China, Chengdu, China, in 2022. He is currently pursuing a Ph.D. degree at the Hong Kong University of Science and Technology, Hong Kong, under the supervision of Prof. Shaojie Shen. His research interests include unmanned aerial vehicles, state estimation in aerial swarms, and swarm systems.\\
\end{minipage}

\begin{minipage}{1.0\columnwidth}
\begin{wrapfigure}{l}{0.3\columnwidth}
\begin{center}
\includegraphics[width=0.3\columnwidth]{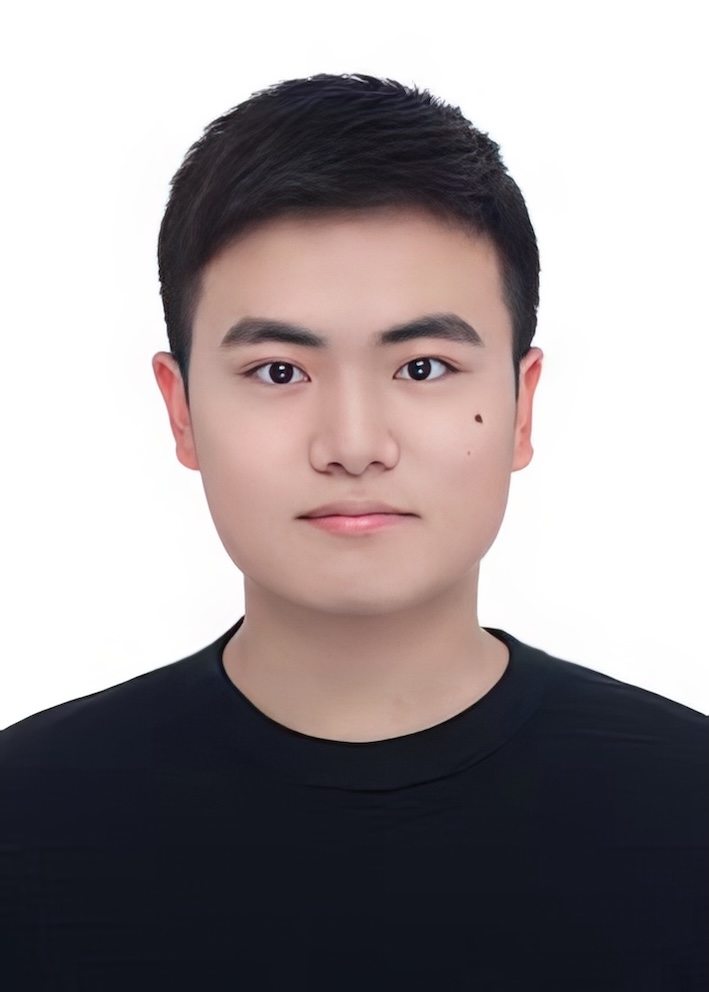}
\end{center}
\end{wrapfigure}
Yang Xu received a B.Eng. degree in Software Engineering from the University of Electronic Science and Technology of China, Chengdu, China, in 2023. He is currently pursuing an M.Phil. degree in the Division of Emerging Interdisciplinary Areas at The Hong Kong University of Science and Technology (HKUST), Hong Kong SAR, China, under the supervision of Prof. Shaojie Shen and Prof. Huan Yin. His research focuses on unmanned aerial vehicles, sensor fusion, and robot learning.\\
\end{minipage}

\begin{minipage}{1.0\columnwidth}
\begin{wrapfigure}{l}{0.3\columnwidth}
\begin{center}
\includegraphics[width=0.3\columnwidth]{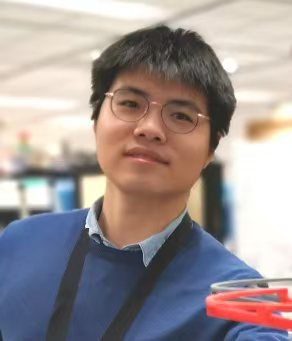}
\end{center}
\end{wrapfigure}
Hao Xu received a B.Sc. degree in Physics from the University of Science and Technology of China in 2016 and a Ph.D. degree in Electronic and Computer Engineering from The Hong Kong University of Science and Technology (HKUST) in 2023. He previously worked as an Algorithm Engineer at Shenzhen DJI Sciences and Technologies Ltd.. His research interests include unmanned aerial vehicles, aerial swarms, state estimation, sensor fusion, localization, and mapping. He is currently an Associate Professor with the Institute of Unmanned Systems at Beihang University.\\
\end{minipage}

\begin{minipage}{1.0\columnwidth}
\begin{wrapfigure}{l}{0.3\columnwidth}
\begin{center}
\includegraphics[width=0.3\columnwidth]{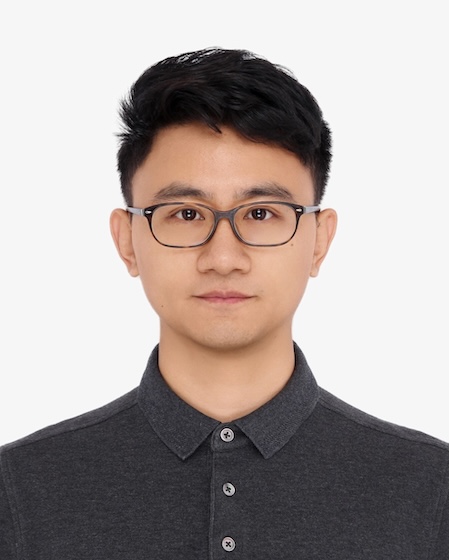}
\end{center}
\end{wrapfigure}
Ximin Lyu received B.Eng. and M.Phil. degrees in Aircraft Manufacturing from the Harbin Institute of Technology, Harbin, China, in 2012 and 2014, respectively. He received a Ph.D. degree in Electronic and Computer Engineering from The Hong Kong University of Science and Technology (HKUST), Hong Kong, China, in 2019. In 2018, he served as a Senior Flight Control Researcher at Shenzhen DJI Sciences and Technologies Ltd., China. Since 2021, he has been an Associate Professor with the Department of Intelligent Systems Engineering at Sun Yat-sen University, Shenzhen, China. His research interests include robotics and controls, with a specific focus on UAV/UGV design, control, and planning.\\
\end{minipage}

\begin{minipage}{1.0\columnwidth}
\begin{wrapfigure}{l}{0.3\columnwidth}
\begin{center}
\includegraphics[width=0.3\columnwidth]{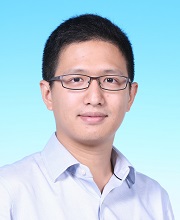}
\end{center}
\end{wrapfigure}
Shaojie Shen received his B.Eng. degree in Electronic Engineering from the Hong Kong University of Science and Technology (HKUST) in 2009. He received an M.S. degree in Robotics and a Ph.D. degree in Electrical and Systems Engineering in 2011 and 2014, respectively, from the University of Pennsylvania. He is currently an Associate Professor in the Department of Electronic and Computer Engineering and the founding director of the HKUST-DJI Joint Innovation Laboratory (HDJI Lab) at HKUST. His research interests are in the areas of robotics and unmanned aerial vehicles, with focus on state estimation, sensor fusion, localization and mapping, and autonomous navigation in complex environments.
He is currently serving as a senior editor for ICRA 2024-2026, and as an associate editor for IJRR 2023-2024. He and his research team received the 2023 IEEE T-RO King-Sun Fu Memorial Best Paper Award and the 2023 IEEE RA-L Best Paper Award, and also achieved the Honorable Mention status for the IEEE T-RO Best Paper Award in 2018 and 2020, and won the Best Student Paper Award in IROS 2018. Additionally, Prof. Shen was recognized as the AI 2000 Most Influential Scholar Award Honorable Mention in 2020 and consecutively from 2021-2024.
\end{minipage}

\end{document}